%% file: Energy-Efficient-FL-TCOM-final.tex
\documentclass[journal, comsoc, 10pt]{IEEEtran}
\usepackage[T1]{fontenc}

\usepackage{amssymb,amsmath,amsfonts,amsthm}
\usepackage{mathrsfs}
\usepackage{cite}
\usepackage{array}
\usepackage{tabularx}
\usepackage[table]{xcolor}
\usepackage{color}
\usepackage{mathtools}
\usepackage{subcaption}
\usepackage{mathtools}
\usepackage{tikz,pgf}
\usepackage{multirow}
\usepackage{paralist}
\usetikzlibrary{calc,positioning,mindmap,trees,decorations.pathreplacing}
\usepackage{graphicx}
\usepackage{bbm}
\usepackage[utf8]{inputenc}
\usepackage{algorithm,algcompatible}
\usepackage{pgfplots}
\usetikzlibrary{plotmarks}

\newtheorem{remark}{Remark}
\newtheorem{definition}{Definition}
\newtheorem{theorem}{Theorem}
\newtheorem{lemma}{Lemma}
\newtheorem{assumption}{Assumption}
\newcommand{\argmax}{\mathop{\rm arg~max}\limits}
\newcommand{\maximize}{\mathop{\rm maximize}\limits}
\newcommand{\minimize}{\mathop{\rm minimize}\limits}
\newcommand{\argmin}{\mathop{\rm arg~min}\limits}

\usepackage[colorinlistoftodos,prependcaption,backgroundcolor=black!5!white,bordercolor=red]{todonotes}
\algblockdefx{FORALLP}{ENDFAP}[1]%
{\textbf{for all }#1 \textbf{do in parallel}}%
{\textbf{end for}}

\IEEEoverridecommandlockouts
\ifCLASSINFOpdf
\else
\fi
\usepackage[cmintegrals]{newtxmath}

\begin{document}
	\title{Energy-Efficient Federated Edge Learning with Streaming Data: A Lyapunov Optimization Approach}
	
	\author{Chung-Hsuan Hu, Zheng~Chen,~and~Erik G. Larsson
	\thanks{The authors are with the Department of Electrical Engineering (ISY), Link\"{o}ping University, Link\"{o}ping, SE-58183 Sweden. E-mail:\{chung-hsuan.hu, zheng.chen, erik.g.larsson\}@liu.se. This work was supported in part by Zenith, ELLIIT, the Swedish Research Council (VR), and the Knut and Alice Wallenberg (KAW) Foundation.
	A part of this work was presented at the 2023 IEEE International Conference on Acoustics, Speech and Signal Processing (ICASSP 2023) \cite{confVersion}.}}
	
	\maketitle

\begin{abstract}
	Federated learning (FL) has received significant attention in recent years for its advantages in efficient training of machine learning models across distributed clients without disclosing user-sensitive data. Specifically, in federated edge learning (FEEL) systems, the time-varying nature of wireless channels introduces inevitable system dynamics in the communication process, thereby affecting training latency and energy consumption. In this work, we further consider a streaming data scenario where new training data samples are randomly generated over time at edge devices. Our goal is to develop a dynamic scheduling and resource allocation algorithm to address the inherent randomness in data arrivals and resource availability under long-term energy constraints. To achieve this, we formulate a stochastic network optimization problem and use the Lyapunov drift-plus-penalty framework to obtain a dynamic resource management design. Our proposed algorithm makes adaptive decisions on device scheduling, computational capacity adjustment, and allocation of bandwidth and transmit power in every round. We provide convergence analysis for the considered setting with heterogeneous data and time-varying objective functions, which supports the rationale behind our proposed scheduling design. The effectiveness of our scheme is verified through simulation results, demonstrating improved learning performance and energy efficiency as compared to baseline schemes.

\end{abstract} 


\section{Introduction}
With the increasing amount of data generated at numerous devices for machine learning (ML) tasks, data-parallel distributed optimization frameworks become extremely useful for learning from decentralized data with on-device processing. Federated learning (FL) is one example of distributed ML that combines on-device training and server-based aggregation \cite{mcmahan2017communication}.
One major issue in FL is the communication bottleneck. Particularly, for FL at the wireless network edge, the unreliable and resource-constrained communication poses additional challenges in the design of resource allocation schemes to achieve optimal learning performance \cite{gafni2022fed}.

Device scheduling in federated edge learning (FEEL) have been explored in \cite{yang2019scheduling, fu2023cl,pandey2022cont,qiao2022con,hu2023sch}, which focus on the effects of data heterogeneity and communication resource limitations. However, energy efficiency and computational resource availability have not been considered in this line of work. 
In every iteration of the training process, each device executes model training and update transmission, leading to the computation-communication tradeoff in terms of the energy consumption. 
Among existing works that jointly consider computation and communication resource constraints, three main performance indicators are learning performance, training latency, and energy consumption. Some works aim at optimizing learning performance subject to energy limitations \cite{sun2021dynamic,xu2021client}, and latency constraints \cite{guo2022dynamicSch,chen2020ajoint,zeng2022joint}. Other works aim at minimizing energy consumption \cite{yang2020energy, li2021opt, alba2021findGrain,battiloro2023lya}, latency \cite{bou2022wireless}, or both \cite{luo2021costeffective,wan2021ca,Nic2022res}, for reaching a certain level of learning performance. Moreover, joint optimization of learning, energy, and/or latency is considered in \cite{zheng2021fed,ji2022client,kim2022novel,jin2022comm,yu2022jointly,peng2023online,zeng2020eng,da2021sim,chen2023enh}.
To achieve high system efficiency, the optimization parameters include device scheduling decisions \cite{li2021opt,kim2022novel,jin2022comm,yu2022jointly}, Central Processing Unit (CPU) frequency \cite{kim2022novel}, transmit power or rate, bandwidth, mini-batch size \cite{Nic2022res,li2021opt},
training data usage \cite{kim2022novel,jie2022heter},
quantization level \cite{bou2022wireless,battiloro2023lya,zh2021design},
and the number of local/global epochs \cite{li2021opt,wan2021ca,luo2021costeffective}.

\begin{table*}[h]
	\caption{Summary of some related works in the literature and their differences.} 
	\centering 
	\begin{tabular}{c|cccc||cccc} 
		\hline
		\textbf{Reference}& \textbf{Wireless}& \textbf{Energy} & \textbf{Static }&\textbf{Data}&\textbf{Device}&\textbf{CPU}&\textbf{Bandwidth}&\textbf{Transmit}\\
		 & \textbf{access} & \textbf{constraint} & \textbf{channel} &\textbf{streams}&\textbf{scheduling}&\textbf{frequency}&\textbf{allocation}&\textbf{power}\\ [0.3ex]
		\hline 
		\cite{energy-aware} &over-the-air & long-term& \checkmark &&\checkmark & & &\\
		\cite{luo2021costeffective}  & -& long-term & - & & \checkmark& & &\\
		\cite{sun2021dynamic}  &over-the-air& long-term & & &\checkmark & & &\\
		\cite{wan2021ca} & digital & iteration-wise & \checkmark& & \checkmark& \checkmark & \checkmark &\\
		\cite{guo2022dynamicSch,zheng2021fed,peng2023online} & digital & long-term & \checkmark& & \checkmark & & &\\
		\cite{ji2022client,xu2021client} & digital & long-term & \checkmark & &  \checkmark & & \checkmark &\\
		\cite{bou2022wireless} & digital & iteration-wise & \checkmark & & &\checkmark&\checkmark&\\
		\cite{chen2020ajoint} & digital & iteration-wise & \checkmark & & \checkmark &&\checkmark&\\
		\cite{zeng2022joint} & digital & iteration-wise & \checkmark & & \checkmark& \checkmark& \checkmark & \\
		\cite{zeng2020eng} & digital & iteration-wise & \checkmark & & \checkmark &&\checkmark&\\
		\cite{battiloro2023lya} & digital & long-term & \checkmark & & \checkmark& \checkmark&&\\
		\cite{yang2020energy,salh2023energy} & digital & iteration-wise & \checkmark& && \checkmark& \checkmark& \checkmark\\
		\cite{alba2021findGrain} & digital & iteration-wise & \checkmark& &\checkmark& \checkmark& \checkmark& \checkmark\\
		\cite{mahmoud2023fl} & digital & iteration-wise & && \checkmark& \checkmark& \checkmark& \checkmark\\
		\cite{confVersion} & digital & long-term & &\checkmark&\checkmark&&&\\
		\textbf{This work} & digital & long-term &&\checkmark&\checkmark&\checkmark&\checkmark&\checkmark\\
		[1ex] 
		\hline 
	\end{tabular}
	\label{tab:literature}
\end{table*}

\subsection{Motivation and Related Work}
\label{sec:moti}
Most of the state-of-the-art methods assume static training data at the edge devices, i.e., the entire dataset is available from the beginning and remains unchanged over the learning process. In real-world applications, training data are usually randomly and continuously generated over time. 
One motivating example is the test driving of autonomous vehicles, where the drivers actively identify upcoming objects on the road and thus provide the training task with streaming labeled data.\footnote{Labeling streaming data is a non-trivial and resource-demanding task. This entails, in addition to increased processing time, a labeling effort that may have to be done manually (as in the given example, human action is required).} Some recent works have considered online FL with time-varying datasets, focusing on the effects of dynamic aggregation weighting \cite{chen2020asynchronous}, workload adjustment \cite{damaskinos2020fleet}, local model prediction \cite{mitra2021ofl}, and adaptation to concept drift \cite{jothimurugesan2023federated}. 
However, joint optimization of computation and communication
resources has not been investigated in these prior works. A few studies have jointly considered latency and learning performance in FL with streaming data \cite{jin2021budget,liu2023dyn}, though under different system models and design parameters.
 
Another aspect that is often overlooked is the computational resource management, as most existing works assume either equal computational energy consumption in every iteration or impose a per-iteration constraint \cite{battiloro2023lya}. When the training process is subject to a long-term energy constraint, dynamic adjustment of CPU frequency can achieve better performance by adapting to the system dynamics as compared to uniform allocation over time \cite{guo2022dynamicSch,bou2022wireless, peng2023online}. 
From the perspective of wireless channels, many existing works assume that the channels remain static during each training iteration, which facilitates the latency analysis and resource allocation decision \cite{yang2020energy,wan2021ca,ji2022client,bou2022wireless,chen2020ajoint}. However, this assumption might be unrealistic, given that the coherent time is usually much shorter than the duration of one training iteration.

We summarize some closely related works that have considered the aforementioned aspects in Tab. \ref{tab:literature}, and highlight their differences in system model and resource management design.

\subsection{Novelty and Contributions}
In this work, we study dynamic resource management in FEEL systems with randomly generated training data, considering time-varying computation capacity, long-term energy constraints and per-iteration latency requirement.\footnote{We focus on supervised learning with labeled data (i.e., the cost of data labeling is not considered a design factor), as our interest mainly lies in efficient allocation of limited computation and communication resources.}
We formulate a stochastic optimization problem and develop a two-stage dynamic scheduling and resource allocation algorithm by using the Lyapunov drift-plus-penalty (DPP) framework. Similar approaches are also adopted in \cite{energy-aware,sun2021dynamic,guo2022dynamicSch,ji2022client,wang2022energy,battiloro2023lya}, but under simpler settings. As compared to the preliminary results presented in the conference version \cite{confVersion}, in this work we further consider the allocation of CPU frequency, bandwidth, and transmit power. 

Another novelty of this work is the consideration of time-varying data importance in the streaming data scenario, which is incorporated in the optimization objective. As compared to the static data scenario, our design can adjust the scheduling decision based on the importance of newly arrived data in terms of the amount and similarity to the distribution of existing data.
The proposed design achieves better learning performance and energy efficiency as compared to alternative schemes, as validated by simulation results.

Moreover, we provide analytical proof that, when the optimal point drifts slowly enough over time, specifically as $O(t^{-1})$
where $t$ refers to the number of iterations, convergence is guaranteed. This analysis is novel and significantly extends existing results on convex optimization with time-varying objectives \cite{sim2020tv,sun2023dist} to the case of FL with heterogeneous data.

\subsection{Paper Organization}
We introduce the system model in Sec. \ref{sec:sys_model} and the long-term stochastic optimization problem formulation in Sec. \ref{sec:problem_form}. The proposed dynamic resource management algorithm is described and elaborated in Sec. \ref{sec:alg}, followed by the convergence analysis in Sec. \ref{sec:conv_ana} and performance evaluation in Sec. \ref{sec:simulation_results}. The conclusions are given in Sec. \ref{sec:conclusion}. In the Appendix, relevant proofs are provided.
	
\section{System Model}
\label{sec:sys_model}
We consider a FEEL system with $K$ devices participating in training a global model $\boldsymbol{\theta}\in\mathbb{R}^d$, with $\mathcal{K}=\{1,...,K\}$ denoting the device set. For each device, we consider that training data are generated randomly over time following some stochastic processes. Let $\mathcal{S}_k(t-1)$ denote the accumulated local dataset at device $k$ until time instant $t-1$, with $\mathcal{S}_k(0)=\emptyset$, and $\mathcal{B}_k(t)$ denote the newly arrived data set at time instant $t$. Then, the entire available local training set at device $k$ is $\mathcal{S}_k(t)=\mathcal{S}_k(t-1)\cup\mathcal{B}_k(t)$. 
The objective is to minimize a time-varying loss function
\begin{equation}
	F\left(\boldsymbol{\theta};\{\mathcal{S}_k(t)\}_{k=1}^K\right)=\sum_{k\in\mathcal{K}}\frac{|\mathcal{S}_k(t)|}{|\cup_{j\in\mathcal{K}}\mathcal{S}_j(t)|} F_k(\boldsymbol{\theta};\mathcal{S}_k(t)),
	\label{eq:globalLoss_t}
\end{equation}
where $F_k(\boldsymbol{\theta};\mathcal{S}_k(t))$ is a local loss function. In the
 special case when all the data arrive at the beginning of the FL process, i.e., $\mathcal{S}_k(t)=\mathcal{S}_k,\forall t$, for some static data set $\mathcal{S}_k$, $\forall k\in\mathcal{K}$, \eqref{eq:globalLoss_t} degenerates to the static counterpart in a standard FL system
\begin{equation}
	F\left(\boldsymbol{\theta};\{\mathcal{S}_k\}_{k=1}^K\right)=\sum_{k\in\mathcal{K}}\frac{|\mathcal{S}_k|}{|\cup_{j\in\mathcal{K}}\mathcal{S}_j|} F_k(\boldsymbol{\theta};\mathcal{S}_k).
	\label{eq:globalLoss_f}
\end{equation}

Federated Averaging is one of the most common and representative FL algorithms, which conducts an iterative process of local training and central aggregation. Specifically, in the $t$-th iteration, $t=1,2,\ldots$,
\begin{enumerate}
	\item The server broadcasts the current global model $\boldsymbol{\theta}(t)$ to the set of scheduled devices $\Pi(t)$, where $|\Pi(t)|=\zeta$.
	\item Each device $k$ runs a fixed number of stochastic gradient descent (SGD) on $F_k(\boldsymbol{\theta};\mathcal{S}_k(t))$, starting from $\boldsymbol{\theta}(t)$, to obtain the model update $\boldsymbol{\triangle}\boldsymbol{\theta}_k(t)$, which is transmitted back to the server.
	\item The server aggregates the received local updates from the devices and obtains a new global model as
	\begin{equation}
		\boldsymbol{\theta}(t+1)=\boldsymbol{\theta}(t)+\sum_{k\in\Pi(t)}w_k(t)\boldsymbol{\triangle}\boldsymbol{\theta}_k(t),
		\label{eq:syncFlAggregation}	
	\end{equation}
	with $\sum_{k\in\Pi(t)}w_k(t)=1$.\footnote{A common choice is $w_k(t)=|\mathcal{S}_k(t)|/\sum_{j\in\Pi(t)}|\mathcal{S}_j(t)|,k\in\Pi(t)$.}
\end{enumerate}
These steps are repeated until the model converges.

In every iteration of the algorithm, energy consumption occurs in two stages: computation and transmission of local model updates, which are affected by the amount of allocated computation and communication resources. The training latency of each device also consists of two parts: model training and transmission delay. In the following subsections, we introduce the energy consumption and latency models. 

\subsection{Energy Consumption Model}
The energy consumption of the $k$-th device in the $t$-th iteration can be written as
\begin{equation}
	E_k(t)=E_k^{\text{cmp}}(t)+E_k^{\text{tr}}(t),
	\label{eq:Ekt}
\end{equation}
where $E_k^{\text{cmp}}(t)$ and $E_k^{\text{tr}}(t)$ are the energy consumed by computation and transmission respectively.

\subsubsection{Energy Consumption for Computation}
We apply dynamic voltage and frequency scaling (DVFS) to adjust the computation speed of a CPU.
Let $f_k(t)$ represent the CPU clock frequency of the $k$-th device in the $t$-th iteration. 
The energy consumption for computing the model update is approximately given by \cite{larsson2011impact}\footnote{The clock frequency $f_k(t)$ and the corresponding energy consumption model $E_k^{\text{cmp}}(t)$ considered here are for both CPU and GPU training scenarios.}
\begin{equation}
	E_k^{\text{cmp}}(t)=\lambda cf_k^2(t),
	\label{eq:Ec}
\end{equation}
where $\lambda$ is a constant and $c$ is the required number of CPU
cycles for executing a fixed number of mini-batch SGD.

\subsubsection{Energy Consumption for Transmission}
To transmit the model updates to the server, the entire bandwidth $B$ is shared among the scheduled devices. We define $\rho_k(t)$ as the fraction of bandwidth assigned to the $k$-th device in the $t$-th iteration with $\sum_{k\in\Pi(t)}\rho_k(t)=1$, and $P_k(t)$ as the allocated transmit power. The transmission rate is, to a first order of approximation, given by Shannon-Hartley formula,
\begin{equation}
	R_k(t)=\rho_k(t)B\log_2 \left(1+\frac{P_k(t)|g_k(t)|^2}{\rho_k(t)BN_0}\right),
	\label{eq:R}
\end{equation}
where $g_k(t)$ is the channel coefficient with $\mathbb{E}[|g_k(t)|^2]=\beta_k$ and $N_0$ is the power spectral density of the noise. 
Assuming that the model updates are compressed to $S$ bits, the required transmission time is 
\begin{equation}
	T^{\text{tr}}_k(t)=\frac{S}{R_k(t)}.
	\label{eq:Tt}
\end{equation}
The energy consumption for the update transmission of each device is then given by
\begin{equation}
	E^{\text{tr}}_k(t)=P_k(t)T^{\text{tr}}_k(t).
	\label{eq:Et}
\end{equation}
In practice, every transmission generates also a constant energy consumption, which is omitted in our problem formulation.

\subsection{Latency Model}
Let $T^{\text{cmp}}_k(t)$ denote the update computation time at device $k$ in iteration $t$, which is given by 
\begin{equation}
	T^{\text{cmp}}_k( t)=\frac{c}{f_k(t)}.
	\label{eq:Tc}
\end{equation}
Recall that $c$ is the required number of CPU cycles for computing the model update.
Based on \eqref{eq:Tt} and \eqref{eq:Tc}, the overall latency of device $k$ to complete the computation and transmission in the $t$-th iteration is
\begin{equation*}
	T_k(t) = T^{\text{cmp}}_k(t)+T^{\text{tr}}_k(t).
\end{equation*}
\begin{figure*}[t!]
	\centering
	\includegraphics[scale=0.3]{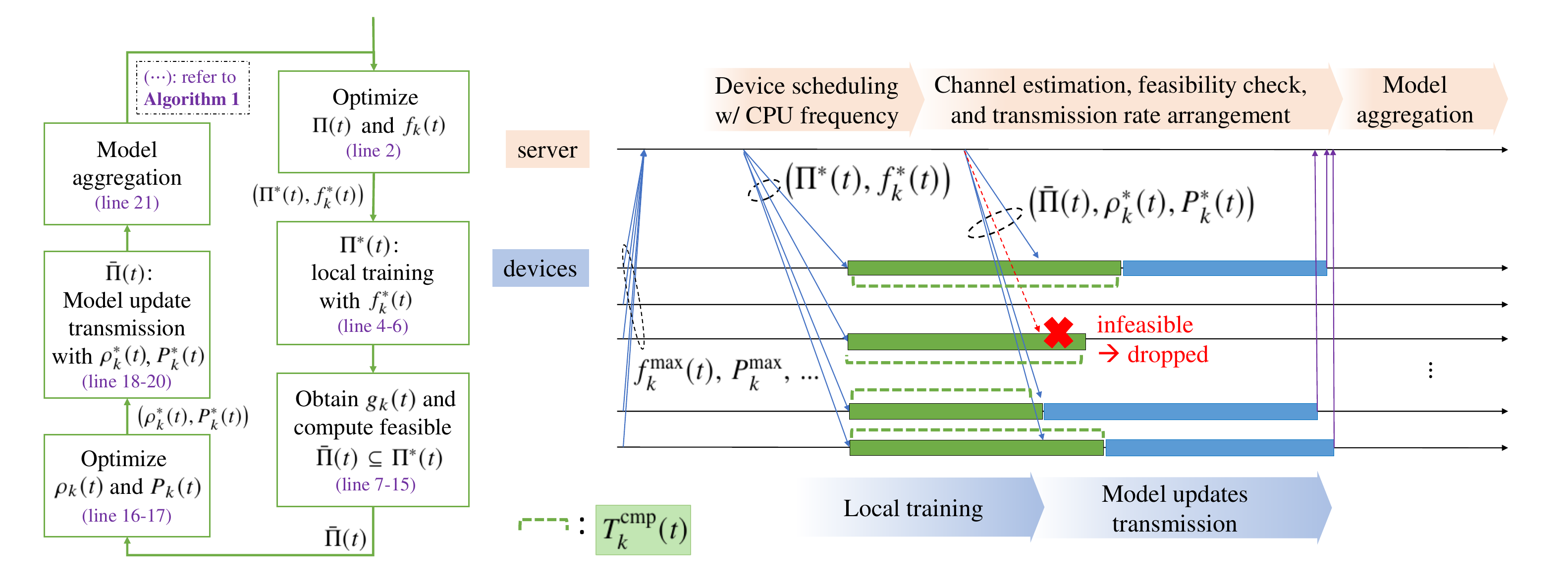}
	\caption{The procedure of solving (P0) and the execution timeline in the $t$-th iteration.  
		The server first computes $\Pi^*(t)$ and $f^*_k(t)$ to initiate the local training process. Before the server conducts model aggregation, a feasible subset $\bar{\Pi}(t)\subseteq\Pi^*(t)$ is determined by checking the transmission time budget $T_{\text{rd}}-T_{k}^{\text{cmp}}(t),\forall k$. Then, $\rho_k^*(t)$ and $P_k^*(t)$, $\forall k\in\bar{\Pi}(t)$, are obtained for the update transmission.}
	\label{fig:flowChart}
\end{figure*}

\section{Joint Computation and Communication Resource Management}
\label{sec:problem_form}
In the scenario of streaming data, the accumulated training data $\mathcal{S}_k(t)$ can be highly heterogeneous over time. This motivates us to consider the time correlation and heterogeneity of data generation in the scheduling and resource allocation design. Meanwhile, device latency is also an important aspect for making scheduling decisions. In the case with stragglers, excluding devices with low computing capability in some iterations can help avoiding long training latency.

First, we define $\mathcal{K}_f(t)\subseteq\mathcal{K}$ as a candidate device set with reasonable estimated latency.\footnote{The design of $\mathcal{K}_f(t)$ will be explained in Section \ref{sec:alg}.}  
Then, for each device $k\in\mathcal{K}_f(t)$, we evaluate its importance based on its accumulated local dataset using the following metric
\begin{equation}
	\label{eq:iptc}
	I_k(t;\mathcal{K}_f(t))=\frac{|\mathcal{K}_f(t)|\cdot|\mathcal{B}_k(t)|}{\sum_{j\in\mathcal{K}_f(t)}|\mathcal{B}_j(t)|}+\boldsymbol{1}\{t>1\}\cdot\frac{||\boldsymbol{x}-\boldsymbol{y}_k||_2^2}{||\boldsymbol{x}||_2^2+||\boldsymbol{y}_k||_2^2}.
\end{equation}
The first term of \eqref{eq:iptc} quantifies the proportion of newly arrived data of device $k$ to the average number of newly arrived data in $\mathcal{K}_f(t)$; the second term quantifies the feature dissimilarity between the data that have been utilized, $\boldsymbol{x}$, and those newly generated, $\boldsymbol{y}_k$, by computing the normalized Euclidean distance between them. For example, with labeled data, such feature vectors can be developed based on the label distribution.
For a data collection $\mathcal{A}$, let $L(\mathcal{A})=\{l_i\}_{i=1}^{m}$ record the number of data with label $i$ and $\bar{L}(\mathcal{A})=(\sum_{i=1}^{m}l_i)/m$ be the average of $L(\mathcal{A})$. 
The feature of $\mathcal{A}$ can be heuristically quantified as
\begin{equation*}
	\delta(\mathcal{A})=\left[L(\mathcal{A})-\bar{L}(\mathcal{A})\right]/\bar{L}(\mathcal{A})
\end{equation*}
from the aspect of label distribution.
Then,  $\boldsymbol{y}_k=\delta(\mathcal{B}_k(t))$ and $\boldsymbol{x}=\delta(\mathcal{X})$ represent the feature of the new data at device $k$ and the previously utilized data in the system respectively, where $\mathcal{X}=\cup_{k\in\mathcal{K}}\mathcal{S}_k(\hat{t}_k)$, and $\hat{t}_k=\max\{\tau|\tau\leq t-1,k\in\Pi(\tau)\}$ is the last time when device $k$ was scheduled.

The processing time of one iteration depends on the latency of each scheduled device, which is affected by the allocated CPU frequency, transmit power, and bandwidth. However, these decisions also affect the energy consumption. For example, allocating higher CPU frequency leads to faster local training, but with increased energy consumption. To balance the energy-latency tradeoff, we formulate a long-term stochastic optimization problem as follows:
\begin{subequations}
	\label{eq:opt}
	\begin{align}
			(\text{P0})~\maximize~~& \limsup\limits_{T\rightarrow \infty}\frac{1}{T} \sum_{t=1}^{T}\mathbb{E}\left[\sum_{k\in \Pi(t)}  I_k\left(t;\mathcal{K}_f(t)\right)\right],\\
			\textrm{subject~to}~~& \limsup\limits_{T\rightarrow \infty}\frac{1}{T} \sum_{t=1}^{T}\mathbb{E}\left[E_k(t)\right]\leq E^{\text{avg}}_{k},\forall k \label{eq:const1}\\
		& P_k(t)\leq P^{\max}_k,\forall k,\forall t\label{eq:const2} \\
		& f_k(t)\leq f^{\max}_k(t),\forall k,\forall t\label{eq:const3} \\
		& T_k(t)\leq T_{\text{rd}},\forall k,\forall t\label{eq:const4}\\
		& \sum_{j\in\Pi(t)}\rho_j(t)=1,\forall t \label{eq:const5}\\
		& \Pi(t)\subseteq\mathcal{K}_f(t)\subseteq \mathcal{K},\forall t.\label{eq:const6}   	
	\end{align}
\end{subequations} 
In every iteration $t$, the decision variables are
\begin{itemize}
	\item  $\Pi(t)$: scheduled device set for the current round of training;
	\item $f_k(t)$: allocated CPU cycle frequency for local training at each scheduled device $k\in\Pi(t)$;
	\item $\rho_k(t)$: fraction of bandwidth allocated for model transmission from device $k$;
	\item $P_k(t)$: transmit power of device $k$.
\end{itemize} 
The expectations are taken over the randomness in data generation, computing capability, and wireless link quality.
Each device has a long-term average energy constraint as in \eqref{eq:const1}; $P^{\max}_k$ denotes the maximum transmit power; $f^{\max}_k(t)$ represents the maximum available computation capacity;\footnote{$f^{\max}_k(t)$ varies over time according to the activity level of device $k$ in every iteration.} the latency constraint \eqref{eq:const4} assures that every global iteration can be completed within a certain time window. 

\section{Dynamic Scheduling and Resource Allocation Algorithm}
\label{sec:alg}
We use the Lyapunov optimization framework to solve (P0) \cite{neely2010stochastic}. 
The time-average constraint in \eqref{eq:const1} is transformed into a queue stability problem by introducing a virtual queue $Q_k(t)$ for each device $k$, which evolves over iterations as
\begin{equation}
	Q_k(t+1)=\max\left[Q_k(t)+s_k(t)E_k(t)-E_k^{\text{avg}},0\right].
	\label{eq:virtualQ}
\end{equation} 
Here, $s_k(t)=1$ when device $k$ is scheduled, i.e., $k\in\Pi(t)$; otherwise, $s_k(t)=0$.
The Lyapunov function and one-step Lyapunov drift with respect to $\boldsymbol{Q}(t)=[Q_1(t),...,Q_K(t)]$ are defined as
$\mathcal{L}(\boldsymbol{Q}(t))=\frac{1}{2}\sum_{k=1}^{K}Q^2_k(t)$ and
\begin{equation}
	\Delta(\boldsymbol{Q}(t))=\mathbb{E}\left[\mathcal{L}(\boldsymbol{Q}(t+1))-\mathcal{L}(\boldsymbol{Q}(t))|\boldsymbol{Q}(t)\right].
	\label{eq:lyaDrift}
\end{equation}
The drift-plus-penalty (DPP) is given as
\begin{equation}
	\Delta(\boldsymbol{Q}(t))-V\mathbb{E}\left[\sum_{k=1}^{K}s_k(t)I_k(t;\mathcal{K}_f(t))\lvert\boldsymbol{Q}(t)\right], \label{eq:dpp}
\end{equation} 
where $V>0$ is a constant parameter that balances the tradeoff between queue stability and objective optimization. 
By the principle of minimizing an upper bound of \eqref{eq:dpp}, (P0) can be transformed into
\begin{align}
	(\text{P1})~\minimize~~&\sum_{k\in\Pi(t)}\left[Q_k(t)E_k(t)-V I_k(t;\mathcal{K}_f(t))\right]\label{obj:minDPP}\\
	\textrm{subject~to}~~&\eqref{eq:const2}\text{--}\eqref{eq:const6}.\nonumber
\end{align}
Details of this transformation are provided in Appendix \ref{sec:pf_P0}.
Compared to the original long-term formulation in (P0), the transformed problem allows us to make sequential decisions by only solving per-iteration optimization problems. 

Note that the instantaneous channel coefficient $g_k(t)$ may vary within one iteration, but the large-scale fading coefficient $\beta_k$ is relatively stable in the entire process. Hence, we use $\beta_k$ to solve the optimization problem at the scheduling phase, while at the aggregation phase, the knowledge of $g_k(t)$ is acquired for optimizing the allocation of communication resources ($\rho_k(t)$ and $P_k(t)$). 
In other words, in every iteration, the server obtains the solution of (P1) and makes the decisions in two stages: 
\begin{inparaenum}[1)]
\item device scheduling and CPU frequency adjustment (Sec. \ref{prob:bf_scheduling}), which occurs before local training;
\item the allocation of bandwidth and transmit power (Sec. \ref{prob:af_training}), which occurs after local training and before update transmission.
\end{inparaenum}
At the second stage, with the knowledge of channel state information, those devices incapable of satisfying the latency requirement \eqref{eq:const4} will be dropped from model aggregation.
\begin{remark}
The two-stage optimization is commonly adopted when the original function is difficult to optimize directly (see, e.g., \cite{sun2021dynamic,mahmoud2023fl}). The solutions can be efficiently computed but not necessarily optimal.
\end{remark}

In the remainder of this section, we provide details on how to obtain the resource allocation decisions by solving (P1), and summarize the algorithm at the end. A high-level flowchart and the notations are presented in Fig. \ref{fig:flowChart} and Tab. \ref{tab:notations}.

\begin{table*}[t!]
	\caption{Summary of parameters in the system, for device $k$ and iteration $t$ if  specified.} 
	\centering 
	\renewcommand{\arraystretch}{1.2}
	\begin{tabular}{cccc} 
		\hline
		\textbf{Notation}& \textbf{Definition} &\textbf{Notation} & \textbf{Definition}\\ [0.3ex]
		\hline 
		$\mathcal{K}/\mathcal{K}_f(t)$ & device set / feasible set for scheduling &  $P_k(t)$/$P_k^{\max}$ & allocated / maximum transmit power\\
		$\mathcal{S}_k(t)$ & training data set & $Q_k(t)$ & virtual queue of energy dissipation\\
		$\mathcal{B}_k(t)$ & newly arrived data & $\rho_k(t)$ & allocated bandwidth fraction \\
		$\boldsymbol{\theta}(t)$ & global model & $\rho_k^*(t)$ & optimized $\rho_k(t)$\\
		$\boldsymbol{\triangle}\boldsymbol{\theta}_k(t)$ & local model update & $c$ & required CPU cycles for local training\\
		$f_k(t)$/$f_k^{\max}(t)$ & allocated / maximum CPU frequency & $g_k(t)$ & channel coefficient\\
		$f_k^*(t)$ & optimized $f_k(t)$ & $\beta_k$ & expected power of $g_k(t)$\\
		$\Pi(t)/\zeta$ & scheduled device set / $|\Pi(t)|$ & $\lambda$ & coefficient of energy consumption\\
		$\Pi^*(t)$/$\bar{\Pi}(t)$ & optimized $\Pi(t)$ / feasible set for aggregation & $T_{\text{rd}}$ & time limit of one iteration\\
		$E_k(t)$ & energy consumption & $N_0$ & noise power spectral density\\
		$E_k^{\text{cmp}}(t)$ & computation energy consumption & $R_k(t)$ & transmission rate \\
		$E_k^{\text{tr}}(t)$ & transmission energy consumption & $I_k(t,\mathcal{K}_f(t))$ & data importance metric\\
		$T_k(t)$ & time spent in the current iteration & $\boldsymbol{x}$/$\boldsymbol{y}_k$ & feature vector of existing/new data\\
		$T_k^{\text{cmp}}(t)$ & time for model update computation & $\gamma$ & time-reserve scaling for transmission\\
		$T_k^{\text{tr}}(t)$ & time for model update transmission&  $\epsilon$ & scaling for frequency optimization\\
		$E_k^{\text{avg}}$ & time-average energy budget & $\tilde{R}_k(t)$ & surrogate transmission rate\\
		$B$ & total bandwidth  & 
		$S$ & model size (number of bits) \\[1ex] 
		\hline 
	\end{tabular}
	\label{tab:notations}
\end{table*}
 
\subsection{Scheduling Phase: Determine $\Pi^*(t)$ and $f_k^*(t)$}
\label{prob:bf_scheduling}
Before local training, the server needs to make decisions on which devices can participate in the current round of training and which CPU frequency should be allocated for computing local model updates.
To handle the constraints \eqref{eq:const2}, \eqref{eq:const4}, and \eqref{eq:const5} concerning unknown channel state information $g_k(t)$ and resource allocation decisions $P_k(t)$ and $\rho_k(t)$, the following measures are taken:
\begin{itemize}
	\item Set $P_k(t)=P_k^{\max}$. Since \eqref{obj:minDPP} increases with $P_k(t)$, minimizing an upper bound on \eqref{obj:minDPP} provides a sub-optimal solution of (P1).\footnote{See  Appendix \ref{sec:pf_increaseP} for the proof.}  
	\item Set $\rho_k(t)=1/\zeta$ such that \eqref{eq:const5} is satisfied, where $\zeta$ is the number of scheduled devices.
	\item Define a surrogate rate function
	\begin{equation}
		\tilde{R}_k(t)=\frac{\gamma B}{\zeta}\log_2\left(1+\frac{P_k^{\max}\beta_k\zeta}{BN_0}\right),
		\label{eq:surrogate_rate}
	\end{equation}
	which will be inserted into \eqref{eq:Tt} to compute device latency $T_k(t)$ and check if the constraint in \eqref{eq:const4} is satisfied.
	This function is obtained by considering $\mathbb{E}[|g_k(t)|^2]=\beta_k$ and applying a scaling factor $\gamma\leq 1$. A lower value of $\gamma$ underestimates the actual achievable rate during the transmission phase, thus reserving more time for update transmission. 
\end{itemize}
Then, (P1) is re-arranged as
\begin{align}
	\underset{\Pi(t),f_k(t),k\in\Pi(t)}{\minimize}~~\sum_{k\in\Pi(t)}&\Big[Q_k(t)\lambda cf_k(t)^2-VI_k(t;\mathcal{K}_f(t))\nonumber \\&+\frac{Q_k(t)P_k^{\max} S\zeta}{\gamma B\log_2\left(1+\frac{P_k^{\max}\beta_k\zeta}{BN_0}\right)}\Big],\nonumber\\
	\textrm{subject~to}~~& \Pi(t)\subseteq\mathcal{K}_f(t)\subseteq \mathcal{K},\nonumber\\
	&f_k(t)\leq f^{\max}_k(t),~\forall k\in\Pi(t),\nonumber\\
	&\frac{c}{f_k(t)}+\frac{S\zeta}{\gamma B\log_2\left(1+\frac{P^{\max}_k\beta_k\zeta}{BN_0}\right)}\leq T_{\text{rd}}.\label{ieq:ltc_const}
\end{align}
Since the objective increases with $f_k(t)$, and \eqref{ieq:ltc_const} gives a lower bound on $f_k(t)$, we obtain the optimal CPU frequency as 
\begin{equation*}
	\tilde{f}_k(t,\gamma)\triangleq c/\left[T_{\text{rd}}-\frac{ S\zeta}{\gamma B\log_2\left(1+\frac{P_k^{\max}\beta_k\zeta}{BN_0}\right)}\right]
\end{equation*}
when $0<\tilde{f}_k(t,\gamma)\leq f_k^{\max}(t)$. An infeasible $\tilde{f}_k(t,\gamma)$ indicates that the latency constraint \eqref{eq:const4} might be violated due to large transmission delay. We define the frequency-optimized set as
\begin{equation}
\tilde{\mathcal{K}_{f}}(t,\gamma)=\{k|0<\tilde{f}_k(t,\gamma)\leq f_k^{\max}(t),k\in\mathcal{K}\}.\label{set:optFreqSet}
\end{equation}
To avoid an overly small candidate set $\tilde{\mathcal{K}_{f}}(t,\gamma)$ for device scheduling, which might lead to large bias, we impose an additional constraint on the size of feasible device set. Specifically,
\begin{equation}
	\label{eq:fsb_set}
	\mathcal{K}_{f}(t)=\begin{cases} \tilde{\mathcal{K}_{f}}(t,\gamma), &|\tilde{\mathcal{K}_{f}}(t,\gamma)|\geq\epsilon\zeta\\
		\{k|\frac{c}{f_k^{\max}(t)}\leq T_\text{rd},k\in\mathcal{K}\}, & \text{otherwise},
	\end{cases}
\end{equation}
with CPU frequencies, $\forall k\in\mathcal{K}_{f}(t)$,
\begin{equation}
	\label{eq:opt_fk}
	f_k^*(t)=\begin{cases} \tilde{f}_k(t,\gamma), &|\tilde{\mathcal{K}_{f}}(t,\gamma)|\geq\epsilon\zeta\\
		f_k^{\max}(t), & \text{otherwise}.
	\end{cases}
\end{equation}
Here, $\epsilon\geq 1$ guarantees a sufficiently large frequency-optimized set $\tilde{\mathcal{K}_{f}}(t,\gamma)$ for device scheduling. When $|\tilde{\mathcal{K}_{f}}(t,\gamma)|$ is too small, as shown in \eqref{eq:fsb_set} and \eqref{eq:opt_fk}, a larger set of devices capable of completing local training in time is selected, with maximum CPU frequencies.
Note that the choice of $\gamma$ affects the feasibility of frequency optimization, as for any $\gamma_1\leq\gamma_2$, $\tilde{\mathcal{K}_{f}}(t,\gamma_1)\subseteq\tilde{\mathcal{K}_{f}}(t,\gamma_2)$.
Therefore, the choice of $\gamma$ involves a trade-off between reserving more time for update transmission and optimizing CPU frequencies.
Finally, we obtain the scheduling decision as
\begin{equation}
	\Pi^*(t)=\underset{\Pi(t)\subseteq\mathcal{K}_f(t)}{\argmin}~~\sum_{k\in\Pi(t)}\xi_k(t),\label{eq:p2_obj}
\end{equation}
where
\begin{equation}
	\xi_k(t)=Q_k(t)\lambda cf_k^*(t)^2
	+\frac{Q_k(t)P_k^{\max} S\zeta}{\gamma B\log_2\left(1+\frac{P_k^{\max}\beta_k\zeta}{BN_0}\right)}-VI_k(t;\mathcal{K}_f(t)).
	\label{opt:factor}
\end{equation}
Finally, the scheduling decision $\Pi^*(t)$ is obtained by collecting $\zeta$ devices with the smallest values of $\xi_k(t)$.

\begin{remark}
	The intuition behind the scheduling policy presented in \eqref{eq:p2_obj} and \eqref{opt:factor} is that those devices with higher data importance value (larger $I_k(t;\mathcal{K}_f(t)$) and under-consumed energy (smaller virtual queue size $Q_k(t)$) will be prioritized. Larger $V$ puts more weight on the data importance value in the scheduling decision.
\end{remark}

\subsection{Aggregation Phase: Determine $\rho_k^*(t)$ and $P_k^*(t)$}
\label{prob:af_training}
After local training, the server needs to determine the allocated bandwidth and transmit power for each participating device. Given the scheduling decision $\Pi^*(t)$ and CPU frequency $f_k^*(t)$, with the channel state information $g_k(t)$, (P1) can be re-arranged as 
\begin{align}
	\text{(P2)}~~
	&\underset{\underset{k\in\Pi^*(t)}{\rho_k(t),P_k(t),}}{\minimize}\sum_{k\in\Pi^*(t)}\frac{Q_k(t)P_k(t)S}{\rho_k(t)B\log_2\left(1+\frac{P_k(t)|g_k(t)|^2}{\rho_k(t)BN_0}\right)},\nonumber\\
	\textrm{s.t.}~~&\sum_{j\in\Pi^*(t)}\rho_j(t)=1,~\forall k\in\Pi^*(t),~P_k(t)\leq P_k^{\max},\nonumber\\
	&\rho_k(t)B\log_2\left(1+\frac{P_k(t)|g_k(t)|^2}{\rho_k(t)BN_0}\right)\geq\frac{Sf_k^*(t)}{f_k^*(t)T_{\text{rd}}-c}.\label{ieq:p1}
\end{align}
Note that (P2) is convex on the feasible domain of $\rho_k(t)$, but not convex on $P_k(t)$.\footnote{See Appendix \ref{sec:pf_conv_p1} for details.}
Recall that the objective is an increasing function of $P_k(t)$. Thus, we form a convex problem from (P2) by minimizing an upper bound on the objective evaluated at $P_k(t)=P_k^{\max}$ and obtain the following problem
\begin{align}
	&\underset{\rho_k(t), k\in\Pi^*(t)}{\minimize}~~\sum_{k\in\Pi^*(t)}\frac{Q_k(t)P_k^{\max}S}{\rho_k(t)B\log_2\left(1+\frac{P_k^{\max}|g_k(t)|^2}{\rho_k(t)BN_0}\right)},\nonumber\\
	&\textrm{subject~to}~~\sum_{j\in\Pi^*(t)}\rho_j(t)=1,~\forall k\in\Pi^*(t)\nonumber\\
	&\rho_k(t)B\log_2\left(1+\frac{P_k^{\max}|g_k(t)|^2}{\rho_k(t)BN_0}\right)\geq \frac{Sf_k^*(t)}{f_k^*(t)T_{\text{rd}}-c}.\label{ieq:latency_rho}
\end{align}

For any device $k\in\Pi^*(t)$, we can obtain a lower bound of $\rho_k(t)$ from \eqref{ieq:latency_rho}, denoted as $\rho_k^{\min}(t)$. Then, there exists no feasible solution if $\sum_{k\in\Pi^*(t)}\rho_k^{\min}(t)>1$.
This lower bound is the minimum required bandwidth to satisfy the latency constraint and it can be explicitly derived as\footnote{See Appendix \ref{sec:pf_rhomin} for details.}
\begin{equation}
	\label{eq:rho_min}
	\rho_k^{\min}(t)=\frac{-C_kP_k^{\max}|g_k(t)|^2}{BN_0\left[W_{-1}(-C_ke^{-C_k})+C_k\right]},
\end{equation}
where
\begin{equation*}
	C_k=\frac{Sf_k^*(t)N_0\ln2}{|g_k(t)|^2P_k^{\max}[f_k^*(t)T_{\text{rd}}-c]},
\end{equation*}
and $W_{-1}(\psi)$ is the Lambert W function that satisfies $W_{-1}(\psi)e^{W_{-1}(\psi)}=\psi$. We define $\bar{\Pi}(t)$ as the feasible device set, initialized as $\Pi^*(t)$.
If $\sum_{k\in\bar{\Pi}(t)}\rho_k^{\min}(t)>1$, we recursively remove one device from $\bar{\Pi}(t)$ with maximum $\rho_k^{\min}(t)$ until $\sum_{k\in\bar{\Pi}(t)}\rho_k^{\min}(t)\leq1$. The idea is to remove the fewest devices for a feasible solution, as more participating devices usually lead to better learning performance.
Then, we obtain a feasible solution $\rho_k^*(t)$ by solving
\begin{align*}
	\text{(P3)}~~\underset{\rho_k(t),k\in\bar{\Pi}(t)}{\minimize}~~&\sum_{k\in\bar{\Pi}(t)}\frac{Q_k(t)P_k^{\max}S}{\rho_k(t)B\log_2\left(1+\frac{P_k^{\max}|g_k(t)|^2}{\rho_k(t)BN_0}\right)},\nonumber\\
	\textrm{subject~to}~&\sum_{j\in\bar{\Pi}(t)}\rho_j(t)=1,~\eqref{ieq:latency_rho},~\forall k\in\bar{\Pi}(t)
\end{align*}
via the barrier method, using the log barrier for \eqref{ieq:latency_rho} followed by Newton's method for minimization \cite{boyd2004convex}.

\begin{algorithm}[t!]
	\caption{Learning-aware dynamic resource management}
	\label{alg:algo1}
	\begin{algorithmic}[1]
		\STATE Obtain $f_k^{\max}(t)$, $P_k^{\max}$, $\mathcal{S}_k(t)$, $\beta_k$, $Q_k(t)$, $E_k^{\text{avg}}$, $\forall k\in\mathcal{K}$, $\gamma$, $c$, $V$, $T_{\text{rd}}$, $S$, $B$, $\lambda$, $N_0$, $\zeta$, $\boldsymbol{\theta}(t)$;
		\STATE Determine scheduling policy $\Pi^*(t)$ by computing \eqref{eq:p2_obj}, with CPU frequency $f_k^*(t)$ computed from \eqref{eq:opt_fk}.
		\STATE Server broadcast $\boldsymbol{\theta}(t)$ to the set of scheduled devices $\Pi^*(t)$.
		\FORALLP{device $k\in\Pi^*(t)$}
		\STATE Compute $\boldsymbol{\triangle}\boldsymbol{\theta}_k(t)$ with CPU frequency $f_k^*(t)$.
		\ENDFAP
		\WHILE{$\{g_k(t)|k\in\Pi^*(t)\}$ is available}
		\STATE Compute $\rho_k^{\min}(t)$ in \eqref{eq:rho_min}, $\forall k\in\Pi^*(t)$
		\STATE $\bar{\Pi}(t)\leftarrow\Pi^*(t)$
		\WHILE{$\sum_{k\in\bar{\Pi}(t)}\rho_k^{\min}(t)>1$}
		\STATE $j=\argmax\{\rho_k^{\min}(t)|k\in\bar{\Pi}(t)\}$
		\STATE $\bar{\Pi}(t)=\bar{\Pi}(t)\setminus j$
		\ENDWHILE
		\STATE \textbf{break}
		\ENDWHILE
		\STATE Determine bandwidth fraction $\rho_k^*(t)$ by solving (P3).
		\STATE Compute transmit power $P_k^*(t)$ from \eqref{eq:opt_Pt}.
		\FORALLP{device $k\in\bar{\Pi}(t)$}
		\STATE Transmit $\boldsymbol{\triangle}\boldsymbol{\theta}_k(t)$ to the server with $\rho_k^*(t)$ and $P_k^*(t)$.
		\ENDFAP
		\STATE Update $\boldsymbol{\theta}(t+1)$ according to \eqref{eq:syncFlAggregation} with $\Pi(t)=\bar{\Pi}(t)$.
		\STATE Update $Q_k(t),\forall k\in\mathcal{K}$ according to \eqref{eq:virtualQ} and \eqref{eq:eng_queue}.
	\end{algorithmic}
\end{algorithm}

Based on the feasible device subset $\bar{\Pi}(t)$, allocated CPU frequency $f_k^*(t)$, and bandwidth faction $\rho_k^*(t)$ obtained from (P3), the only remaining part in (P2) is the transmit power allocation problem, which can be written as
\begin{align}
	\underset{P_k(t),k\in\bar{\Pi}(t)}{\minimize}~~&\sum_{k\in\bar{\Pi}(t)}\frac{Q_k(t)P_k(t)S}{\rho_k^*(t)B\log_2\left(1+\frac{P_k(t)|g_k(t)|^2}{\rho_k^*(t)BN_0}\right)},\nonumber\\
	\textrm{subject~to}~~&\frac{c}{f_k^*(t)}+\frac{S}{\rho_k^*(t)B\log_2\left(1+\frac{P_k(t)|g_k(t)|^2}{\rho_k^*(t)BN_0}\right)}\leq T_{\text{rd}},
	\label{opt_sub3_const}\\	
	&P_k(t)\leq P_k^{\max},~\forall k\in\bar{\Pi}(t).\nonumber
\end{align}	
Since the objective increases with $P_k(t)$ and \eqref{opt_sub3_const} gives a lower bound on $P_k(t)$, we obtain
\begin{equation}
	P_k^*(t)=\frac{\rho_k^*(t)BN_0}{|g_k(t)|^2}\left(2^{D_k(t)}-1\right),\label{eq:opt_Pt}
\end{equation}
where $D_k(t)=S/\left[\left(T_{\text{rd}}-c/f_k^*(t)\right)\rho_k^*(t)B\right]$.

After the model transmission and aggregation, the server updates the energy queues $\{Q_k(t)\}_{k\in\mathcal{K}}$ according to \eqref{eq:virtualQ}, with
\begin{equation}
	\label{eq:eng_queue}
	s_k(t)E_k(t)=\begin{cases}E_k^{\text{cmp}}(t), & k\in\Pi^*(t)\setminus\bar{\Pi}(t)\\
		E_k^{\text{cmp}}(t)+E_k^{\text{tr}}(t), &k\in\bar{\Pi}(t) \\
		0, &\text{otherwise.}\end{cases}
\end{equation}
The updated queue size will affect the scheduling and resource allocation decisions in the next iteration. For instance, a device with a larger queue size will be less likely to be scheduled in the following iterations.

The entire decision-making process for our proposed dynamic resource management scheme is summarized in Algorithm \ref{alg:algo1}.
\section{Convergence Analysis}
\label{sec:conv_ana}
To simplify the notation in the analysis, we let $F_k^t(\boldsymbol{\theta})\triangleq F_k(\boldsymbol{\theta};\mathcal{S}_k(t))$ be the $k$-th local loss function evaluated at the dataset $\mathcal{S}_k(t)$, and $F^t(\boldsymbol{\theta})\triangleq\sum_{k\in\mathcal{K}}\frac{|\mathcal{S}_k(t)|}{|\cup_{j\in\mathcal{K}}\mathcal{S}_j(t)|}F_k^t(\boldsymbol{\theta})$ be the global loss function. Also, we define $F^{t,*}=\min_{\boldsymbol{\theta}} F^t(\boldsymbol{\theta})$ with the corresponding optimizer $\boldsymbol{\theta}^{t,*}$; similarily,  $F_k^{t,*}=\min_{\boldsymbol{\theta}} F_k^t(\boldsymbol{\theta})$ with the optimizer $\boldsymbol{\theta}_k^{t,*}$. To facilitate analysis, we assume single-step SGD, i.e., $\boldsymbol{\triangle}\boldsymbol{\theta}_k(t)=-\alpha\nabla F_k^t(\boldsymbol{\theta}(t))|_{\xi_k(t)\subseteq\mathcal{S}_k(t)},\forall k$, with $\alpha$ as the learning rate and $\xi_k(t)$ as the mini-batch. 
We introduce the following definition and assumptions:\footnote{In the following, $||\cdot||$ refers to $||\cdot||_2$.}
\begin{definition}(\emph{Data heterogeneity level}): 	
	For a device subset $\Pi\subseteq\mathcal{K}$ and $\sum_{k\in\Pi}w_k=1$, we define, for any $t$,
	\vspace{-.2cm}
	\begin{equation*}
		\Gamma_1^t(\Pi)=F^{t,*}-\sum\nolimits_{k\in\Pi}w_kF_k^{t,*},
	\end{equation*}
	\vspace{-.3cm}
	\begin{equation*}
		\Gamma_2^t(\Pi)=F^{t,*}-\sum\nolimits_{k\in\Pi}w_kF_k^t\left(\boldsymbol{\theta}^{t,*}\right).
	\end{equation*}
\end{definition}
\vspace{-.3cm}
\begin{assumption}
	(Smoothness): Each local loss function $F_k^t(\boldsymbol{\theta}),\forall k, \forall t$ is $L$-smooth, i.e., $\forall \boldsymbol{\theta}_1,\boldsymbol{\theta}_2\in\mathbb{R}^d$,
	\vspace{-.2cm}
	\begin{align*}
		||\nabla F_k^t\left(\boldsymbol{\theta}_1\right)&-\nabla F_k^t\left(\boldsymbol{\theta}_2\right)||\leq{L}||\boldsymbol{\theta}_1-\boldsymbol{\theta}_2||,\\
		\noalign{\noindent\text{or equivalently,}}
		F_k^t(\boldsymbol{\theta}_1)-F_k^t(\boldsymbol{\theta}_2)&\leq\nabla F_k^t(\boldsymbol{\theta}_2)^T(\boldsymbol{\theta}_1-\boldsymbol{\theta}_2)+\frac{L}{2}||\boldsymbol{\theta}_1-\boldsymbol{\theta}_2||^2.
	\end{align*}
	\label{asp:smth}	
\end{assumption}
\vspace{-.5cm}
\begin{assumption}
	(Strong convexity): Each local loss function $F_k^t(\boldsymbol{\theta}),\forall k, \forall t$ is $\mu$-strongly-convex, i.e., $\forall \boldsymbol{\theta}_1, \boldsymbol{\theta}_2\in\mathbb{R}^d$,
	\vspace{-.3cm}
	\begin{equation*}
		F_k^t\left(\boldsymbol{\theta}_1\right)-F_k^t\left(\boldsymbol{\theta}_2\right)\geq\nabla F_k^t\left(\boldsymbol{\theta}_2\right)^T\left(\boldsymbol{\theta}_1-\boldsymbol{\theta}_2\right)+\frac{\mu}{2}||\boldsymbol{\theta}_1-\boldsymbol{\theta}_2||^2.
	\end{equation*}
	\label{asp:conv}
\end{assumption}
\vspace{-.7cm}
\begin{assumption}
	(First and second moment constraint): $\forall \boldsymbol{\theta}\in\mathbb{R}^d$, the stochastic gradient $\nabla F_k^t(\boldsymbol{\theta})|_{\mathcal{\xi}\subseteq\mathcal{S}_k(t)},\forall k,\forall t$ satisfies
	\begin{equation*}
		\mathbb{E}_{\xi\subseteq\mathcal{S}_k(t)}\left[\nabla F_k^t\left(\boldsymbol{\theta}\right)|_{\mathcal{\xi}}\right]=\nabla F_k^t\left(\boldsymbol{\theta}\right),
	\end{equation*}
	\vspace{-.7cm}
	\begin{equation*}
		\mathbb{E}_{\xi\subseteq\mathcal{S}_k(t)}\left[||\nabla F_k^t\left(\boldsymbol{\theta}\right)|_{\mathcal{\xi}}||^2\right]\leq C_1||\nabla F^t_k\left(\boldsymbol{\theta}\right)||^2+C_2,
	\end{equation*}
	for some $C_1\geq 1$ and $C_2\geq 0$.
\end{assumption}
\begin{assumption}
	(Bounded optimizer drift): The optimizers of $F^{t-1}$ and $F^{t}$, $\forall t=1,2,...$, satisfy, for some function $M(t)$,
	\begin{equation}
		\label{ieq:boundedDrift}
		||\boldsymbol{\theta}^{t-1,*}-\boldsymbol{\theta}^{t,*}||\leq M(t).
	\end{equation}
\end{assumption}
\vspace{-.3cm}
\begin{assumption}
	(Bounded local optimizers): The local optimizers satisfy $||\boldsymbol{\theta}_k^{t,*}||\leq \Lambda_k,\forall t$, for some $\Lambda_k>0$, $\forall k\in\mathcal{K}$.
	\label{asp:bndLocal}
\end{assumption}
\vspace{-.5cm}
\begin{remark}
	We prove in Appendix \ref{uniBnd:proof} that under \textbf{Assumptions \ref{asp:smth}, \ref{asp:conv}, \ref{asp:bndLocal}}, we have $||\boldsymbol{\theta}^{t,*}||\leq \max_{k\in\mathcal{K}}\Lambda_kL/\mu,\forall t$. Since $||\boldsymbol{\theta}_k^{t,*}||,\forall k$ and $||\boldsymbol{\theta}^{t,*}||$ are uniformly bounded and $\mathcal{K}$ is a finite set, there exist non-negative $\varphi_1$ and $\varphi_2$ such that for any $\Pi\subseteq\mathcal{K}$, $\forall t$,  
	\begin{equation*}
		|\Gamma_1^t(\Pi)|\leq\varphi_1,
		|\Gamma_2^t(\Pi)|\leq\varphi_2.
	\end{equation*}
\end{remark}
\vspace{-.4cm}
\subsection{Convergence Results}
\begin{theorem}
	\label{thm}
	Under \textbf{Assumptions \ref{asp:smth}-\ref{asp:bndLocal}}, with $M(t)=\frac{m}{t+1}$ for some $m>0$, if the  learning rate  satisfies
	\begin{equation*}
		\begin{cases}
			\alpha<\min\left[\frac{\mu}{4(C_1-1)L^2},\frac{1}{L}\right], &\text{ if }C_1\neq 1,\\
			\alpha<1/L, &\text{otherwise,}
		\end{cases}
	\end{equation*}	
	the following result holds for $t_0>\lfloor m\rfloor$,
	\begin{align*}
		\mathbb{E}\left[||\boldsymbol{\theta}(t+1)-\boldsymbol{\theta}^{t,*}||^2\right]&\leq\bar{\kappa}_{t_0,t}^{t-t_0+1}||\boldsymbol{\theta}(t_0)-\boldsymbol{\theta}^{t_0-1,*}||^2\\
		&\quad+3mC_4\sum_{i=t_0}^t\frac{\bar{\kappa}^{t-i}_{t_0+1,t}}{i+1}+\frac{C_3\left(1-\bar{\kappa}_{t_0+1,t}^{t-t_0+1}\right)}{1-\bar{\kappa}_{t_0+1,t}},
	\end{align*}
	where for $t_1,t_2>0,\bar{\kappa}_{t_1,t_2}=\frac{1}{|t_2-t_1|+1}\sum_{t=\min(t_1,t_2)}^{\max(t_1,t_2)}\kappa(t),$
	$$\kappa(t)=C_4(1+2M(t)),$$ $$C_4=1-\mu\alpha+2(C_1-1)L^2\alpha^2<1,$$ $$C_3=\alpha \Omega+\alpha^2\left(C_2+\frac{2(C_1-1)L^2\Omega}{\mu}\right),$$ $\Omega=2(\varphi_1+\varphi_2)$ and $\mathbb{E}[\cdot]$ is total expectation over $\Pi(j)$, $\{\xi_k(j)|k\in\Pi(j)\}$, and $\boldsymbol{\theta}(j)$, $j=t_0,...,t$. The proof is given in  Appendix \ref{thm:proof}.
\end{theorem}
\begin{lemma}
	\label{lm:sr-conv}
	With $0<\kappa<1$, $\sum_{i=1}^{t}\frac{\kappa^{t-i}}{i+1}\rightarrow 0$ when $t\rightarrow\infty$. The proof is given in Appendix \ref{sec:proof_lm3}.
\end{lemma}
\begin{remark}
	$\mathbb{E}\left[||\boldsymbol{\theta}(t+1)-\boldsymbol{\theta}^{t,*}||^2\right]$ converges when ${\kappa(t_0)<1}$ (implying $\bar{\kappa}_{t_0,t},\bar{\kappa}_{t_0+1,t}<1,\forall t\geq t_0$). 
	By selecting $\vartheta$, $0\leq\vartheta<1$, and taking $$t_0>\max\left(\left\lfloor m\right\rfloor,\left\lfloor\frac{2mC_4}{(1-C_4)(1-\vartheta)}\right\rfloor\right),$$ we guarantee that
	$\kappa(t_0)<1-\vartheta(1-C_4)$, such that the sequence converges with a contraction factor  controlled	by $\vartheta$.\footnote{With larger value of $\vartheta$, the contraction starts after a larger number of initial iteration $t_0$.}
Together with \textbf{Lemma \ref{lm:sr-conv}}, we have
	\begin{align*}
		\lim_{t\rightarrow\infty}\mathbb{E}\left[||\boldsymbol{\theta}(t+1)-\boldsymbol{\theta}^{t,*}||^2\right]&\leq\frac{C_3}{1-C_4}\\
		&=\frac{\Omega+\alpha(C_2+\frac{2(C_1-1)L^2\Omega}{\mu})}{\mu-2(C_1-1)L^2\alpha}.
	\end{align*}
Because of the sampling noise, the bound does not vanish even with i.i.d. training data and full device participation ($\Omega=0$ yields $C_2\cdot\frac{\alpha}{\mu-2(C_1-1)L^2\alpha}$). This  is commonly observed in fixed-learning-rate schemes \cite{bottou2018ML}. A smaller $\alpha$ gives a tighter asymptotic bound, but slows the contraction per iteration, as $C_4,\bar{\kappa}_{t_0,t},\bar{\kappa}_{t_0+1,t}$ become larger.
\end{remark}

\begin{remark}
	Suppose that the FL performance is evaluated at iteration $\hat{t}$, with the entire training data denoted by $\mathcal{S}(\hat{t})\triangleq\sum_{k\in\mathcal{K}}\mathcal{S}_k(\hat{t})$. 
	Then, for any $t<\hat{t}$, a higher level of resemblance between the exploited data set $\bar{\mathcal{S}}(t)\triangleq\{\mathcal{S}_k(i)|k\in\Pi(i),i=1,...,t\}$ and $\mathcal{S}(\hat{t})$ intuitively leads to a smaller $||\boldsymbol{\theta}^{t,*}-\boldsymbol{\theta}^{\hat{t},*}||$ and thus potentially a lower or more aggressively decaying $M(t)$. This results in smaller $\bar{\kappa}_{t_0,t}$ and $\bar{\kappa}_{t_0+1,t}$, making the bound contract faster over the iterations and thus accelerating the convergence. This motivates us to prioritize devices with more newly arrived and heterogeneous data, as the size of the undiscovered data set, $|\mathcal{S}(\hat{t})\backslash\bar{\mathcal{S}}(t)|$, is significantly reduced at every iteration.
\end{remark}

\section{Simulation Results}
\label{sec:simulation_results}
We simulate a system of $K=\{40, 25\}$ devices with training data from MNIST and CIFAR-10 \cite{lecun-mnisthandwrittendigit-2010,Krizhevsky09learningmultiple}. The goal is to train a $d$-dimensional convolutional neural network for performing image classification tasks. 
The system parameters are set as follows.
\begin{itemize}
	\item (Training data distribution) $60000$ data samples are distributed evenly to the devices, i.e., $|\cup_{t}\mathcal{B}_k(t)|=60000/K$. For the i.i.d. case, the samples are randomly allocated to all the devices without replacement, while for the non-i.i.d. case, each device contains up to $3$ unique digits. 
	\item (Streaming data generation) Denote by $T_{\text{tot}}$ the entire execution time of the FL system. The data samples arrive in the order of digit, and the first arriving digit is randomly picked at each device. The amount of new arrivals in each iteration is either uniform, or fitted by truncated Poisson or truncated Gaussian distribution with mean value $\mu_k\sim\mathcal{U}(0,T_{\text{tot}})$ and a clipping range $[0,T_{\text{tot}}]$.
	\item CPU frequency limit $f_k^{\max}(t)\sim\mathcal{U}(0.02,1.5)$ GHz.
	\item Maximum transmit power $P_k^{\max}\sim\mathcal{U}(10\text{dBm},30\text{dBm})$.
	\item The devices are randomly placed in a disc of radius $1$ km centered at the server. We model $\beta_k$ as $r_k^{-4}$, where $r_k$ is the distance between device $k$ and the server.
	\item Constant parameters are listed in Table \ref{tab:constParam}.
\end{itemize}
\vspace{-.5cm}
\begin{table}[t!]
	\caption{System parameters} 
	\centering 
	\begin{tabular}{|c|c||c|c|} 
		\hline
		\textbf{Parameter}& \textbf{Value} &\textbf{Parameter} & \textbf{Value} \\ [0.3ex]
		\hline 
		Bandwidth $B$ & $10$MHz & noise power $N_0$ & $10^{-17}$ W \\ power coefficient $\lambda$ & $10^{-25}$ &	model size $S$ &  $32d$\\
		average energy $E_k^{\text{avg}},\forall k$ &$1$ J & $T_{\text{rd}}$(MNIST) &$5$ sec \\
		trade-off scaling $V$ & 50 & $T_{\text{rd}}$(CIFAR-10)& $10$ sec\\[1ex] 
		\hline 
	\end{tabular}
	\label{tab:constParam}
\end{table}

\begin{figure*}[t!]
	\centering
	\begin{subfigure}[c]{0.48\textwidth}
		\input{figure/sameRatioAccuIid-sglCol.tex}
		\caption{MNIST: i.i.d.}	
	\end{subfigure}
	\hfill
	\begin{subfigure}[c]{0.48\textwidth}
	    \vspace{.8cm}
		\input{figure/sameRatioAccuNiid-sglCol.tex}	
		\caption{MNIST: non-i.i.d.}
	\end{subfigure}
	\caption{Test accuracy of the proposed ('prop') and the baseline ('rdm') methods with truncated Gaussian data arrival pattern.}
	\label{fig:testSameRatio}
\end{figure*}
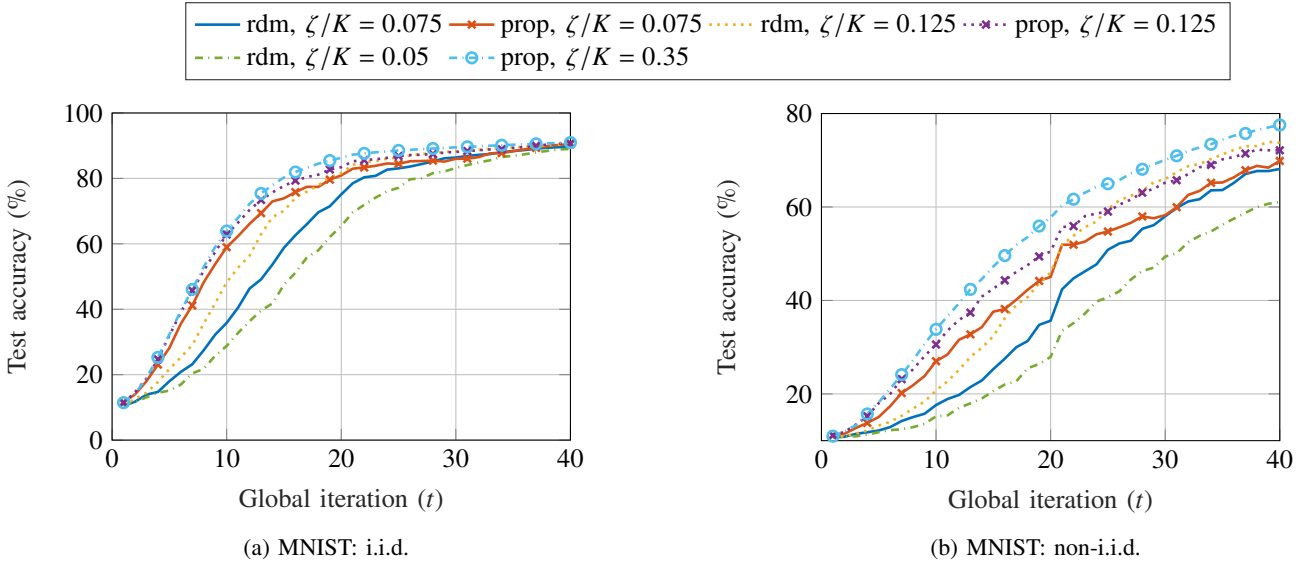
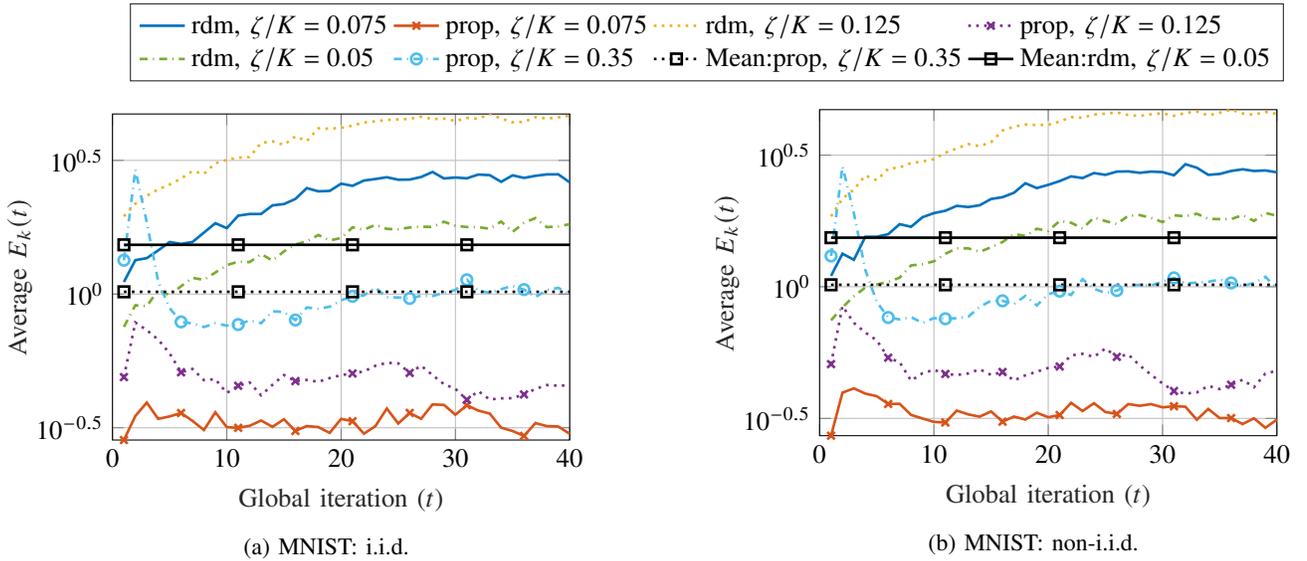
\begin{figure*}[t!]
	\begin{subfigure}[c]{0.48\textwidth}
		\centering
		\input{figure/sameRatioEngIid-sglCol.tex}
		\caption{MNIST: i.i.d.}	
	\end{subfigure}
	\hspace{.45cm}
	\begin{subfigure}[c]{0.48\textwidth}
		\vspace{1.35cm}
		\input{figure/sameRatioEngNiid-sglCol.tex}	
		\caption{MNIST: non-i.i.d.}
	\end{subfigure}
	\caption{Average energy consumption of the proposed ('prop') and the baseline ('rdm') methods with truncated Gaussian data arrival pattern.}
	\label{fig:testSameRatioEng}
\end{figure*}

\subsection{Learning Performance and System Efficiency}
First, we demonstrate the performance gain of our design in both learning and energy efficiency under a per-iteration time constraint $T_{\text{rd}}$. A baseline method for performance comparison is random scheduling. Among all devices that can finish local computation in time, i.e., $f_k^{\max}(t)\geq c/T_{\text{rd}}$, a subset of them will be randomly chosen for participating in the current round. During model transmission, all devices use full transmit power $P_k^{\max}$ and uniform bandwidth allocation with $\rho_k(t)=1/\zeta$. Those who fail to complete transmission in the remaining time $T_{\text{rd}}-T_k^{\text{cmp}}(t)$ will be dropped from model aggregation. 

The comparisons of test accuracy and average energy consumption between our proposed and baseline methods are shown in Figs. \ref{fig:testSameRatio}-\ref{fig:testSameRatioCifar} respectively, for both MNIST (i.i.d. and non-i.i.d.) and CIFAR-10 (non-i.i.d.), with truncated Gaussian arrival pattern.\footnote{The i.i.d. case of CIFAR-10 gives the same remarks as those of the non-i.i.d case and thus is omitted. Moreover, as the results of CIFAR-10 align with those of MNIST, for the remaining experiments we only use MNIST.}
Similarly, Figs. \ref{fig:testSameRatioDist4} and \ref{fig:testSameRatioDist4eng} show the testing loss comparison with  uniform and truncated Poisson arrival patterns.
In general, our proposed design achieves better learning performance than the baseline under the same scheduling ratio. This learning performance gain is more significant with a lower scheduling rate and more imbalanced data arrivals.

In Fig. \ref{fig:testSameRatioCifar}, the learning performance degrades slightly at the later iterations for high scheduling ratios. This is because our algorithm does not solely optimize for learning performance; it also includes an additional constraint on long-term energy consumption in the problem formulation. The random scheduling scheme does not provide any guarantee on the energy consumption level and generally consumes more energy compared to our design.
	There is always a tradeoff between learning performance and energy efficiency. With our design, the learning performance can be improved (at the cost of reduced energy efficiency), for example by tuning $\gamma$ (affecting the feasible device set for scheduling) or $V$ (adjusting the weighting between the importance metric and energy consumption).	
Another possibility could be to adjust the data importance as the model evolves (e.g., adapting the weighting between the two terms in the metric). In particular, the importance metric achieves a fast learning curve in early iterations. However, when the system has acquired knowledge of all labels at later iterations, scheduling those collectively possessing a balanced distribution may be better.\footnote{We have to leave a more detailed study of such an adaptation of the data importance metric for future work.}

More importantly, as shown in Figs. \ref{fig:testSameRatioEng} and \ref{fig:testSameRatioDist4eng}, our method attains to as least $81\%$ reduction in terms of average energy consumption.
The baseline with $\zeta/K=0.05$ and the proposed with $\zeta/K=0.35$ are also provided in both figures to demonstrate the large learning performance gain of our method when the energy consumption of both methods is at a similar level. 

\begin{figure*}[t!]
	\centering
	\begin{subfigure}[c]{0.48\textwidth}
		\input{figure/sameRatioAccuNiidCifar.tex}	
	\end{subfigure}
	\hfill
	\begin{subfigure}[c]{0.48\textwidth}
		\vspace{1cm}
		\centering
		\input{figure/sameRatioEngNiidCifar.tex}	
	\end{subfigure}
	\caption{Test accuracy and average energy consumption of the proposed ('prop') and the baseline ('rdm') methods for CIFAR-10 (non-i.i.d.) with truncated Gaussian data arrival pattern.}
	\label{fig:testSameRatioCifar}
\end{figure*}
\begin{figure*}[t!]
	\hspace{-.5cm}
	\begin{subfigure}[c]{0.48\textwidth}
		\centering
		\input{figure/sameRatioDist4TestIid-sglCol1.tex}
		\caption{i.i.d.}	
	\end{subfigure}
	\hspace{.7cm}
	\begin{subfigure}[c]{0.48\textwidth}
		\centering
		\input{figure/sameRatioDist4TestNiid-sglCol1.tex}	
		\caption{non-i.i.d.}
	\end{subfigure}
	\caption{Testing loss comparison between the proposed ('prop') and the baseline ('rdm') methods with uniformly random ('uni') and truncated Poisson ('ps') data arrival patterns.}
	\label{fig:testSameRatioDist4}
\end{figure*}
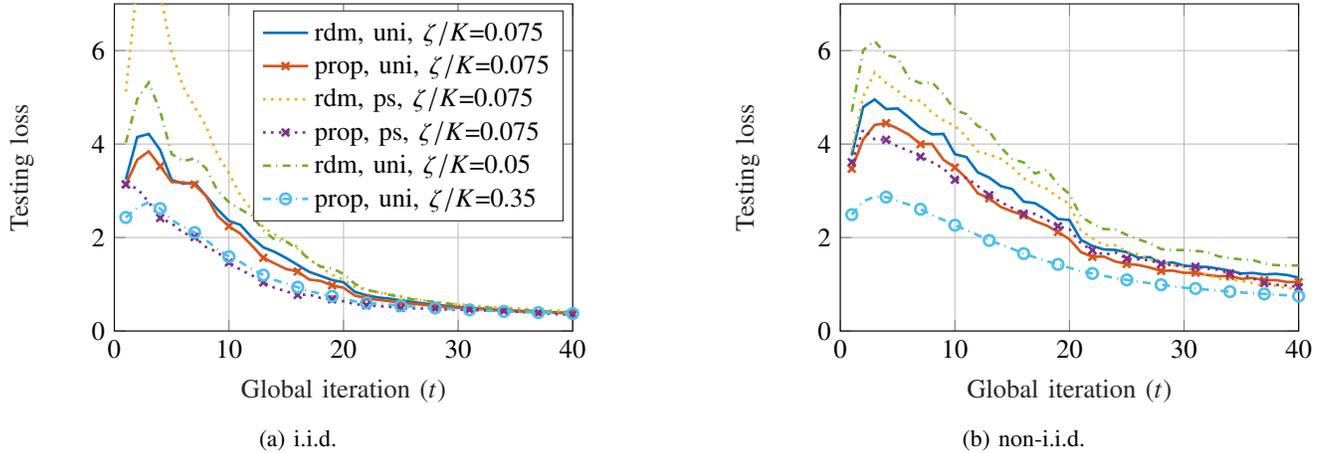
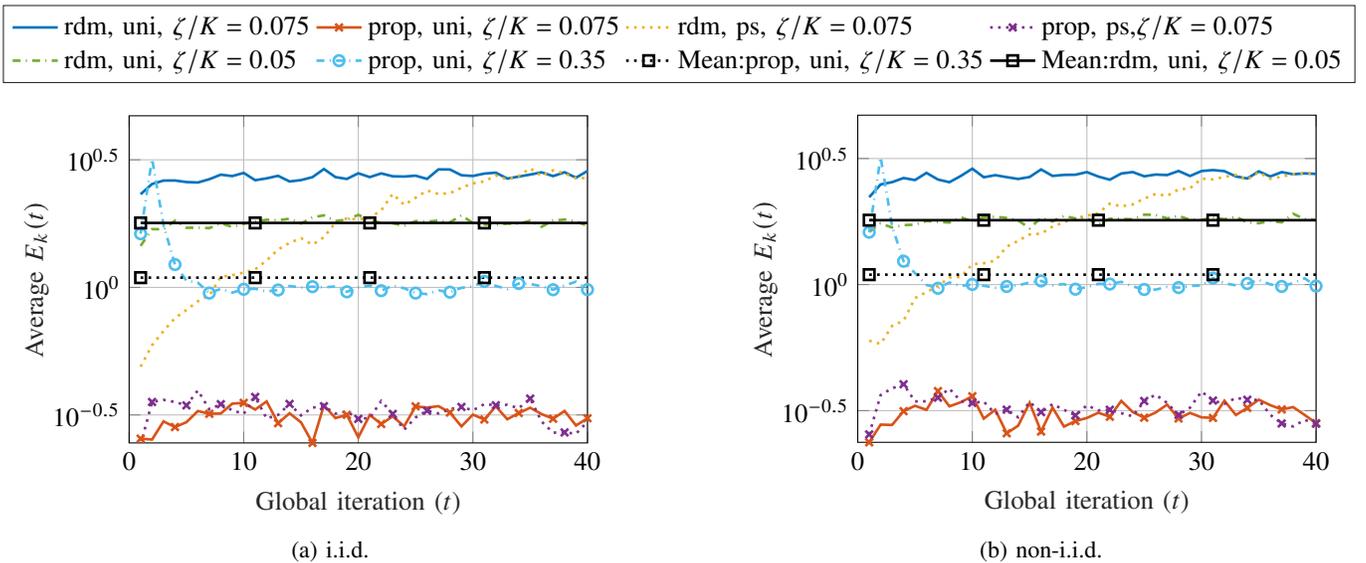
\begin{figure*}[t!]
	\hspace{-.35cm}
	\begin{subfigure}[c]{0.48\textwidth}
		\centering
		\input{figure/sameRatioDist4EngIid-sglCol1.tex}
		\caption{i.i.d.}	
	\end{subfigure}
	\hspace{.5cm}
	\begin{subfigure}[c]{0.48\textwidth}
		\centering
		\vspace{1.45cm}
		\input{figure/sameRatioDist4EngNiid-sglCol1.tex}	
		\caption{non-i.i.d.}
	\end{subfigure}
	\caption{Average energy consumption comparison between the proposed ('prop') and the baseline ('rdm') methods  with uniformly random ('uni') and truncated Poisson ('ps') data arrival patterns.}
	\label{fig:testSameRatioDist4eng}
\end{figure*}

\begin{figure*}[t!]
	\begin{subfigure}[c]{0.48\textwidth}
		\centering
		\input{figure/diffMetricAccu-sglCol.tex}
	\end{subfigure}
	\begin{subfigure}[c]{0.48\textwidth}
		\centering
		\vspace{1cm}
		\input{figure/diffMetricLoss-sglCol.tex}	
	\end{subfigure}
	\caption{Performance comparison between different data importance metrics. Each device has up to $2$ unique digits.}
	\label{fig:metricCmp}
\end{figure*}

\begin{figure*}[t!]
	\begin{subfigure}[c]{0.48\textwidth}
		\centering
		\input{figure/diffGammaAccuIid-sglCol.tex}
		\caption{i.i.d.}	
	\end{subfigure}
	\begin{subfigure}[c]{0.48\textwidth}
		\centering
		\vspace{.9cm}
		\input{figure/diffGammaAccuNiid-sglCol.tex}	
		\caption{non-i.i.d.}
	\end{subfigure}
	\caption{Impact of $\gamma$ on the learning performance, compared with random scheduling. Truncated Gaussian arrival pattern, $T_\text{rd}=1.4$ seconds, $B=3$ MHz, and $\zeta/\mathcal{K}=0.05$.}
	\label{fig:testDiffGamma}
\end{figure*}
\vspace{-.2cm}
\subsection{Effectiveness of the Data Importance Metric}
To validate the appropriateness of our proposed data importance metric in \eqref{eq:iptc}, we show the performance comparison with some alternative metrics, including:\footnote{For this part of simulation, the scaling $V$ is kept sufficiently large to reduce the influence from the energy efficiency perspective.}
\begin{itemize}
	\item $\sum_{k\in\Pi(t)}\log(1+|\mathcal{S}_k(t)|)$, \cite{ji2022client}.
	\item $\sum_{k\in\Pi(t)}|\mathcal{S}_k(t)|$, \cite{xu2021client}.
\end{itemize}
The feature vectors $\boldsymbol{x}$ and $\boldsymbol{y}_k$ in \eqref{eq:iptc} are computed based on the digit-label collection of the training data.
As illustrated in Fig. \ref{fig:metricCmp}, our method gives a clear testing loss improvement over the entire training process and better test accuracy at the later iterations compared to the two alternative designs proposed in \cite{ji2022client,xu2021client}. This highlights the importance of taking into account the statistics of newly arrived data in quantifying data importance, rather than the quantity of accumulated data as considered in the alternative methods.\footnote{Note that due to the difference in optimization framework, only the data importance metrics in \cite{ji2022client,xu2021client} are applied in the simulations.} 
\subsection{Impact of Scaling Factor $\gamma$}
In the model aggregation phase, a device will be dropped if it cannot satisfy the latency constraint. This device dropping effect may cause degraded learning performance and wasted energy, and it can be mitigated by adjusting the scaling factor $\gamma$.  
Recall that $\gamma$ in \eqref{eq:surrogate_rate} affects the calculation of the surrogate rate function, and subsequently affects the optimized computation frequency $f_k^*(t)$. A larger $\gamma$ overestimates the achievable rate during uplink transmission of model updates, and a smaller $\gamma$ underestimates the achievable rate, thus reserving more time for transmission. 
Fig. \ref{fig:testDiffGamma} shows the test accuracy results with different choices of $\gamma$.
With a smaller $\gamma$, our proposed method can have a lower device dropping rate than the baseline, and therefore higher test accuracy is achieved. 
However, adopting an overly small $\gamma$ drastically decreases the size of frequency-optimized set $\tilde{\mathcal{K}_{f}}$ as defined in \eqref{set:optFreqSet}. Then, according to our design in \eqref{eq:fsb_set}, the feasible set will become the same as in the baseline case. 
Moreover, a smaller $\gamma$ also raises the minimum requirement of computation capability. A higher CPU frequency needs to be adopted for each scheduled device and the energy consumption increases.
To summarize, the choice of $\gamma$ should strike a balance between improved learning performance and energy efficiency.
\vspace{-.2cm}
\section{Conclusions}
\label{sec:conclusion}
We propose a dynamic resource management design for FEEL systems with randomly arriving data samples and long-term energy constraints. Our design takes into account the randomness and importance of data arrivals, as well as the time-varying resource availability in device scheduling and resource allocation decisions. Using the Lyapunov DPP framework, our proposed two-stage solution determines, in every global iteration, the device scheduling decision along with CPU frequency adjustment, followed by the allocation of bandwidth and transmit power. The proposed design demonstrates clear advantages in reducing energy consumption while achieving better learning performance compared to alternative methods that ignore the system dynamics.

\appendix
\section{Appendix}
\subsection{Formulation of (P1)}
\label{sec:pf_P0}
By plugging \eqref{eq:virtualQ} into \eqref{eq:lyaDrift}, we have
\begin{align}
	\Delta(\boldsymbol{Q}(t))&=\mathbb{E}\left[\frac{1}{2}\sum_{k=1}^{K}\left[Q^2_k(t+1)-Q^2_k(t)\right]\lvert\boldsymbol{Q}(t)\right]\nonumber\\&\leq D+\mathbb{E}\left[\sum_{k=1}^{K}Q_k(t)\left(s_k(t)E_k(t)-E_k^{\text{avg}}\right)\lvert\boldsymbol{Q}(t)\right],
\end{align}
where $D=\frac{1}{2}\sum_{k=1}^{K}\left[(E_k^{\text{max}})^2+(E_k^{\text{avg}})^2\right]$, and $E_k^{\text{max}}$ is the maximum energy consumption of device $k$ in any iteration.

The drift-plus-penalty, given as
\begin{equation*}
\Delta(\boldsymbol{Q}(t))-V\mathbb{E}\left[\sum_{k=1}^{K}s_k(t)I_k(t;\mathcal{K}_f(t))\lvert\boldsymbol{Q}(t)\right],
\end{equation*}  is upper bounded by the following quantity
\begin{align}
	D+\mathbb{E}\left\{\sum_{k=1}^{K}\left[Q_k(t)s_k(t)E_k(t)-V s_k(t)I_k(t;\mathcal{K}_f(t))\right]\lvert\boldsymbol{Q}(t)\right\}.
\end{align}
By greedily minimizing the term inside the conditional expectation in every time slot, we have \eqref{obj:minDPP} as the per-slot objective.

\subsection{The Increasing Property of \eqref{obj:minDPP}}
\label{sec:pf_increaseP}
Denote by $g(\boldsymbol{P}(t))$ the rewritten form of \eqref{obj:minDPP},
\begin{align*}
	\Upsilon(\boldsymbol{P}(t))=\sum_{k\in\Pi(t)}&\Big[Q_k(t)\lambda cf_k^2(t)-VI_k(t;\mathcal{K}_f(t))\nonumber \\&+\frac{Q_k(t)P_k(t)S}{\rho_k(t)B\log_2\left(1+\frac{P_k(t)|g_k(t)|^2}{\rho_k(t)BN_0}\right)}\Big],
\end{align*}
with $\boldsymbol{P}(t)=\left\{P_k(t)\right\}_{k\in\Pi(t)}$.
Then, $\Upsilon$ increases with $P_k(t)$ since
\begin{equation*}
	\frac{\partial \Upsilon}{\partial P_k(t)}=\frac{y\left[(1+x)\ln(1+x)-x\right]}{\left(1+x\right)\ln(1+x)\log_2(1+x)}\geq 0,
\end{equation*}
where
\begin{equation}
	x=zP_k(t),~y=\frac{Q_k(t)S}{\rho_k(t)B},~ z=\frac{|g_k(t)|^2}{\rho_k(t)BN_0}.\label{eq:xyz}
\end{equation}

\subsection{Convexity Check of (P2)}
\label{sec:pf_conv_p1}
Define $\mathcal{W}(\boldsymbol{\rho}(t),\boldsymbol{P}(t))$ as the objective function of (P2) with $\boldsymbol{\rho}(t)=\left\{\rho_k(t)\right\}_{k\in\Pi^*(t)}$ and $\boldsymbol{P}(t)=\left\{P_k(t)\right\}_{k\in\Pi^*(t)}$. Then, $\frac{\partial^2 \mathcal{W}}{\partial P_k(t)\partial P_j(t)}=0, \forall k\neq j$, and
\begin{align}
	\frac{\partial^2 \mathcal{W}}{\partial P^2_k(t)}&=\frac{yz\left[-\left(2+x\right)\ln(1+x)+2x\right]\ln2}{(1+x)^2(\ln(1+x))^3},\label{eq:2ndDePk}
\end{align}
where $x$, $y$, and $z$ are defined in \eqref{eq:xyz}. Since $x\geq 0$, the term in the brackets of \eqref{eq:2ndDePk} is non-positive, and thus $\frac{\partial^2 \mathcal{W}}{\partial P^2_k(t)}\leq 0$. Hence, the Hessian of $w$ w.r.t. $\boldsymbol{P}(t)$ is not positive semi-definite, and thus $\mathcal{W}$ is not convex on $\boldsymbol{P}(t)$. 

To prove $\mathcal{W}$ is convex on $\boldsymbol{\rho}(t)$, define $\mathcal{W}_k$ as the summand in $\mathcal{W}$, i.e., $\mathcal{W}=\sum_{k\in\Pi^*(t)}\mathcal{W}_k$; $\mathcal{G}(\sigma)\triangleq\frac{Q_k(t)P_k(t)S}{B\sigma}$ and
\begin{equation}
	\mathcal{U}(\sigma)\triangleq \sigma\log_2\left(1+\frac{P_k(t)|g_k(t)|^2}{\sigma BN_0}\right).
\end{equation}
Then, $\mathcal{W}_k=\mathcal{G}(\mathcal{U}(\rho_k(t)))$. Since $\mathcal{G}$ is convex and nonincreasing, and $\mathcal{U}$ is concave, as
\begin{equation}
	\frac{\partial^2 \mathcal{U}}{\partial \sigma^2}=-\frac{\iota^2}{\sigma(\sigma+\iota)^2\ln2}\leq 0,
\end{equation}
with $\iota=P_k(t)|g_k(t)|^2/(BN_0)$ and $\sigma=\rho_k(t)\geq 0$, we conclude that $\mathcal{W}_k$ is convex \cite{boyd2004convex}. Then, $\mathcal{W}$, as a sum of convex functions $\mathcal{W}_k,\forall k\in\Pi^*(t)$, is also convex.
To check the convexity of \eqref{ieq:p1}, we define
\begin{equation*}
	\mathcal{V}_k(\boldsymbol{\rho}(t))=-\rho_k(t)B\log_2\left(1+\frac{P_k(t)|g_k(t)|^2}{\rho_k(t)BN_0}\right).
\end{equation*}
Then, $\mathcal{V}_k(\boldsymbol{\rho}(t))=-B\mathcal{U}(\rho_k(t))$ is convex due to the concavity of $\mathcal{U}$.
Together with the affine constraint $\sum_{j\in\Pi^*(t)}\rho_j(t)=1$, we conclude that (P2) is convex on $\boldsymbol{\rho}(t)$.

\subsection{Proof of \eqref{eq:rho_min}}
\label{sec:pf_rhomin}
The minimum $\rho_k^{\min}(t)$ is obtained when \eqref{ieq:latency_rho} holds with equality. We define
\begin{equation}
	\bar{\rho}_k(t)=BN_0\rho_k(t)/\left(P_k^{\max}|g_k(t)|^2\right).\label{def:rhoBar}
\end{equation}
Then, the equality can be rewritten as
\begin{equation}
	\bar{\rho}_k(t)\ln(1+1/\bar{\rho}_k(t))=C_k.
	\label{eq:rhoBarEq}
\end{equation}
By multiplying \eqref{eq:rhoBarEq} with $1+\frac{1}{\bar{\rho}_k(t)}$,
\begin{equation}
	C_k+\ln(1+1/\bar{\rho}_k(t))=(1+1/\bar{\rho}_k(t))C_k.\label{eq:forExpEval}
\end{equation}
By exponentiating both sides of \eqref{eq:forExpEval},
\begin{equation*}
	e^{C_k}(1+1/\bar{\rho}_k(t))=e^{(1+1/\bar{\rho}_k(t))C_k},
\end{equation*}
and thus
\begin{equation}
	-C_k(1+1/\bar{\rho}_k(t))e^{-(1+1/\bar{\rho}_k(t))C_k}=-C_ke^{-C_k}. \label{eq:preLambert}
\end{equation}
Since $-\frac{1}{e}\leq-C_ke^{-C_k}<0$, we solve \eqref{eq:preLambert} as
\begin{equation}
	\label{eq:W_form}
	\bar{\rho}_k(t)=-C_k/\left[W_{-1}(-C_ke^{-C_k})+C_k\right]
\end{equation}
and obtain $\rho_k(t)$ by inserting \eqref{eq:W_form} into \eqref{def:rhoBar}.

\subsection{Proof of $||\boldsymbol{\theta}^{t,*}||\leq \max_{k\in\mathcal{K}}\Lambda_kL/\mu,\forall t$}
\label{uniBnd:proof}
Note that $\forall\boldsymbol{\theta}_1,\boldsymbol{\theta}_2$,
\begin{equation}
	\nabla F_k^{t}(\boldsymbol{\theta}_1)=\nabla F_k^{t}(\boldsymbol{\theta}_2)+\mathcal{H}^t_k(\boldsymbol{\theta}_1,\boldsymbol{\theta}_2)(\boldsymbol{\theta}_1-\boldsymbol{\theta}_2),
	\label{eq:wHessian}
\end{equation}	
where $\mathcal{H}^t_k(\boldsymbol{\theta}_1,\boldsymbol{\theta}_2)=\int_0^1\nabla^2F_k^{t}(\boldsymbol{\theta}_2+\tau(\boldsymbol{\theta}_1-\boldsymbol{\theta}_2))d\tau$ and\\$\mu\boldsymbol{I}\preceq \mathcal{H}^t_k(\boldsymbol{\theta}_1,\boldsymbol{\theta}_2)\preceq L\boldsymbol{I}$ (Lemma 1.2.2 in \cite{convNesterov}).\footnote{Here, for simplicity, we assume that $F_k^t, \forall k, \forall t$ are twice-differentiable.} Then, based on \eqref{eq:wHessian} and the fact that $\nabla F_k^{t}(\boldsymbol{\theta}_k^{t,*})=0$,
\begin{align}
	\nabla F^{t}(\boldsymbol{\theta}^{t,*})&=\sum_{k\in\mathcal{K}}w_k(t)\nabla F_k^{t}(\boldsymbol{\theta}^{t,*})\nonumber\\
	&=\sum_{k\in\mathcal{K}}w_k(t)\mathcal{H}^t_k(\boldsymbol{\theta}^{t,*},\boldsymbol{\theta}_k^{t,*})\left(\boldsymbol{\theta}^{t,*}-\boldsymbol{\theta}_k^{t,*}\right),\label{eq:wH}
\end{align}
where $w_k(t)=\frac{|\mathcal{S}_k(t)|}{|\cup_{j\in\mathcal{K}}\mathcal{S}_j(t)|}$. Since $\nabla F^{t}(\boldsymbol{\theta}^{t,*})=0$, \eqref{eq:wH} gives
\begin{equation*}
	\boldsymbol{\theta}^{t,*}=\left(\sum_{k\in\mathcal{K}}w_k(t)\mathcal{H}^t_k(\boldsymbol{\theta}^{t,*},\boldsymbol{\theta}_k^{t,*})\right)^{-1}\sum_{k\in\mathcal{K}}w_k(t)\mathcal{H}^t_k(\boldsymbol{\theta}^{t,*},\boldsymbol{\theta}_k^{t,*})\boldsymbol{\theta}_k^{t,*}.
\end{equation*}
Since $||\boldsymbol{\theta}_k^{t,*}||\leq\max_{k\in\mathcal{K}}\Lambda_k$, we have
\begin{align*}
	||\boldsymbol{\theta}^{t,*}||&\leq\frac{1}{\mu}||\sum_{k\in\mathcal{K}}w_k(t)\mathcal{H}^t_k(\boldsymbol{\theta}^{t,*},\boldsymbol{\theta}_k^{t,*})\boldsymbol{\theta}_k^{t,*}||\\
	&\leq\frac{L}{\mu}\sum_{k\in\mathcal{K}}w_k(t)||\boldsymbol{\theta}_k^{t,*}||\leq \max_{k\in\mathcal{K}}\Lambda_kL/\mu.
\end{align*}

\subsection{Proof of \textbf{Theorem \ref{thm}}}
\label{thm:proof}
For the simplicity of exposition, we define 
\begin{equation*}
	\boldsymbol{g}\left(t\right)=\sum_{k\in\Pi(t)}w_k(t)\left[\nabla F_k^t\left(\boldsymbol{\theta}(t)\right)|_{\xi_k(t)\subseteq\mathcal{S}_k(t)}\right],
\end{equation*}  
and its mean $\bar{\boldsymbol{g}}\left(t\right)=\sum_{k\in\Pi(t)}w_k(t)\nabla F_k^t\left(\boldsymbol{\theta}(t)\right)$.
We have  the following two lemmas, which are adaptations of results in \cite{Li2020On} (see Appendices \ref{sec:proof_lm1} and \ref{sec:proof_lm2} for the proofs).
\begin{lemma}
	If $\alpha\leq\frac{1}{L}$, then $\forall \Pi(t)\in\mathcal{K}$,
	\begin{equation*}
		||\boldsymbol{\theta}\left(t\right)-\boldsymbol{\theta}^{t,*}-\alpha\bar{\boldsymbol{g}}\left(t\right)||^2\leq\left(1-\mu\alpha\right)||\boldsymbol{\theta}\left(t\right)-\boldsymbol{\theta}^{t,*}||^2+\alpha \Omega.
	\end{equation*}
	\label{lm:rearrange}
\end{lemma}
\vspace{-.7cm}
\begin{lemma}
	\label{lm:var_g}
	The variance of $\boldsymbol{g}\left(t\right)$ w.r.t. data sampling satisfies
	\begin{align*}
		&\mathbb{E}_{\{\xi_k(t)|k\in\Pi(t)\}}\left[||\boldsymbol{g}\left(t\right)-\bar{\boldsymbol{g}}\left(t\right)||^2\right]\\
		&\leq
		C_2+2(C_1-1)L^2\left(||\boldsymbol{\theta}(t)-\boldsymbol{\theta}^{t,*}||^2+\Omega/\mu\right).
	\end{align*}
\end{lemma}
To establish Theorem 1 we first note that 
\noindent$||\boldsymbol{\theta}\left(t+1\right)-\boldsymbol{\theta}^{t,*}||^2$ can  be rewritten as
\begin{align*}
	&||\boldsymbol{\theta}\left(t+1\right)-\boldsymbol{\theta}^{t,*}||^2=||\boldsymbol{\theta}\left(t\right)-\alpha\boldsymbol{g}\left(t\right)-\boldsymbol{\theta}^{t,*}-\alpha\bar{\boldsymbol{g}}\left(t\right)+\alpha\bar{\boldsymbol{g}}\left(t\right)||^2\nonumber\\
	&=||\boldsymbol{\theta}\left(t\right)-\boldsymbol{\theta}^{t,*}-\alpha\bar{\boldsymbol{g}}\left(t\right)||^2+\alpha^2||\boldsymbol{g}\left(t\right)-\bar{\boldsymbol{g}}\left(t\right)||^2\nonumber\\
	&\quad-2\alpha\left[\left(\boldsymbol{\theta}\left(t\right)-\boldsymbol{\theta}^{t,*}-\alpha\bar{\boldsymbol{g}}\left(t\right)\right)^T\left(\boldsymbol{g}\left(t\right)-\bar{\boldsymbol{g}}\left(t\right)\right)\right].
\end{align*}
Given $\boldsymbol{\theta}(t)$, the following holds based on \textbf{Lemmas \ref{lm:rearrange}} and \textbf{\ref{lm:var_g}},
\begin{align}
	&\mathbb{E}_{\Pi(t),\{\xi_k(t),k\in\Pi(t)\}}\left[||\boldsymbol{\theta}(t+1)-\boldsymbol{\theta}^{t,*}||^2|\boldsymbol{\theta}(t)\right]\nonumber\\
	&=\mathbb{E}_{\Pi(t)}\left[||\boldsymbol{\theta}(t)-\boldsymbol{\theta}^{t,*}-\alpha\bar{\boldsymbol{g}}(t)||^2|\boldsymbol{\theta}(t)\right]\nonumber\\
	&\quad+\alpha^2\mathbb{E}_{\Pi(t)}\left[\mathbb{E}_{\{\xi_k(t),k\in\Pi(t)\}}\left[||\boldsymbol{g}(t)-\bar{\boldsymbol{g}}(t)||^2|\boldsymbol{\theta}(t)\right]|\boldsymbol{\theta}(t)\right]\nonumber\\
	&\leq(1-\mu\alpha)||\boldsymbol{\theta}(t)-\boldsymbol{\theta}^{t,*}||^2+\alpha \Omega\nonumber\\
	&\quad+\alpha^2\left[C_2+2(C_1-1)L^2\left(||\boldsymbol{\theta}(t)-\boldsymbol{\theta}^{t,*}||^2+\Omega/\mu\right)\right]\nonumber\\
	&\leq C_4\left(||\boldsymbol{\theta}(t)-\boldsymbol{\theta}^{t-1,*}||+||\boldsymbol{\theta}^{t-1,*}-\boldsymbol{\theta}^{t,*}||\right)^2+C_3\nonumber\\
	&=C_4\left(||\boldsymbol{\theta}(t)-\boldsymbol{\theta}^{t-1,*}||^2+||\boldsymbol{\theta}^{t-1,*}-\boldsymbol{\theta}^{t,*}||^2\right)+C_3\nonumber\\
	&\quad+2C_4||\boldsymbol{\theta}(t)-\boldsymbol{\theta}^{t-1,*}||\cdot||\boldsymbol{\theta}^{t-1,*}-\boldsymbol{\theta}^{t,*}||.\label{eq:wt_cross}
\end{align}
Note that the cross-term in \eqref{eq:wt_cross} satisfies
\begin{align}
	&||\boldsymbol{\theta}(t)-\boldsymbol{\theta}^{t-1,*}||\cdot||\boldsymbol{\theta}^{t-1,*}-\boldsymbol{\theta}^{t,*}||\leq\nonumber\\
	&\begin{cases}
		||\boldsymbol{\theta}^{t-1,*}-\boldsymbol{\theta}^{t,*}||, & \text{if }||\boldsymbol{\theta}(t)-\boldsymbol{\theta}^{t-1,*}||\leq 1\\
		||\boldsymbol{\theta}(t)-\boldsymbol{\theta}^{t-1,*}||^2\cdot||\boldsymbol{\theta}^{t-1,*}-\boldsymbol{\theta}^{t,*}||, & \text{otherwise,}
	\end{cases}\nonumber\\
	&\leq||\boldsymbol{\theta}^{t-1,*}-\boldsymbol{\theta}^{t,*}||+||\boldsymbol{\theta}(t)-\boldsymbol{\theta}^{t-1,*}||^2\cdot||\boldsymbol{\theta}^{t-1,*}-\boldsymbol{\theta}^{t,*}||.
	\label{ieq:twoCases}
\end{align}
By inserting \eqref{ieq:twoCases} into \eqref{eq:wt_cross}, and based on \eqref{ieq:boundedDrift}, we have
\begin{align}
	&\mathbb{E}_{\Pi(t),\{\xi_k(t),k\in\Pi(t)\}}\left[||\boldsymbol{\theta}(t+1)-\boldsymbol{\theta}^{t,*}||^2|\boldsymbol{\theta}(t)\right]\nonumber\\
	&\leq C_4(1+2||\boldsymbol{\theta}^{t-1,*}-\boldsymbol{\theta}^{t,*}||)||\boldsymbol{\theta}(t)-\boldsymbol{\theta}^{t-1,*}||^2+C_3\nonumber\\
	&\quad+C_4||\boldsymbol{\theta}^{t-1,*}-\boldsymbol{\theta}^{t,*}||^2+2C_4||\boldsymbol{\theta}^{t-1,*}-\boldsymbol{\theta}^{t,*}||\nonumber\\
	&\leq \kappa(t)||\boldsymbol{\theta}(t)-\boldsymbol{\theta}^{t-1,*}||^2+C_3+C_4\left(M^2(t)+2M(t)\right).\label{ieq:sampling3}
\end{align}	
Then, applying telescoping on \eqref{ieq:sampling3} till $t_0\leq t$ gives
\begin{align*}
	&\mathbb{E}\left[||\boldsymbol{\theta}(t+1)-\boldsymbol{\theta}^{t,*}||^2\right]\leq\left[\prod_{i=t_0}^{t}\kappa(i)\right]||\boldsymbol{\theta}(t_0)-\boldsymbol{\theta}^{t_0-1,*}||^2\nonumber\\
	&\quad+\sum_{i=t_0}^t\left[C_4\left(M^2(i)+2M(i)\right)+C_3\right]\prod_{j=1}^{t-i}\kappa(t-j+1).
\end{align*} 
By applying the arithmetic-geometric mean inequality, and using the facts that  $M^2(i)\leq M(i)$ when $i>\lfloor m\rfloor$, and $\bar{\kappa}_{j,k}\leq\bar{\kappa}_{l,k}$ if $l\leq j$, we have
\begin{align*}
	&\mathbb{E}\left[||\boldsymbol{\theta}(t+1)-\boldsymbol{\theta}^{t,*}||^2\right]\leq\bar{\kappa}_{t_0,t}^{t-t_0+1}||\boldsymbol{\theta}(t_0)-\boldsymbol{\theta}^{t_0-1,*}||^2\nonumber\\
	&+3mC_4\sum_{i=t_0}^t\frac{\bar{\kappa}^{t-i}_{t_0+1,t}}{i+1}+C_3\sum_{i=1}^{t-t_0+1}\bar{\kappa}_{t_0+1,t}^{i-1},\text{ when }t_0>\lfloor m\rfloor.
\end{align*}

\subsubsection{Proof of \textbf{Lemma \ref{lm:rearrange}}}
\label{sec:proof_lm1}
\begin{align}
	&||\boldsymbol{\theta}\left(t\right)-\boldsymbol{\theta}^{t,*}-\alpha\bar{\boldsymbol{g}}\left(t\right)||^2\nonumber\\
	&=||\boldsymbol{\theta}\left(t\right)-\boldsymbol{\theta}^{t,*}||^2+\alpha^2||\bar{\boldsymbol{g}}\left(t\right)||^2-2\alpha\bar{\boldsymbol{g}}\left(t\right)^{\text{T}}\left[\boldsymbol{\theta}\left(t\right)-\boldsymbol{\theta}^{t,*}\right]\nonumber\\
	&\leq||\boldsymbol{\theta}\left(t\right)-\boldsymbol{\theta}^{t,*}||^2+\alpha^2\sum_{k\in\Pi(t)}w_k(t)||\nabla F_k^t\left(\boldsymbol{\theta}\left(t\right)\right)||^2\nonumber\\
	&\quad-2\alpha\sum_{k\in\Pi(t)}w_k(t)\left(\boldsymbol{\theta}\left(t\right)-\boldsymbol{\theta}^{t,*}\right)^T\nabla F_k^t\left(\boldsymbol{\theta}\left(t\right)\right)\nonumber\\
	&\leq||\boldsymbol{\theta}\left(t\right)-\boldsymbol{\theta}^{t,*}||^2+2L\alpha^2\sum_{k\in\Pi(t)}w_k(t)\left[F_k^t\left(\boldsymbol{\theta}\left(t\right)\right)-F_k^{t,*}\right]\nonumber\\
	&\quad-2\alpha\sum_{k\in\Pi(t)}w_k(t)\left[F_k^t\left(\boldsymbol{\theta}\left(t\right)\right)-F_k^t\left(\boldsymbol{\theta}^{t,*}\right)+\frac{\mu}{2}||\boldsymbol{\theta}(t)-\boldsymbol{\theta}^{t,*}||^2\right]\nonumber\\
	&\leq\left(1-\mu\alpha\right)||\boldsymbol{\theta}\left(t\right)-\boldsymbol{\theta}^{t,*}||^2+2L\alpha^2\sum_{k\in\Pi(t)}w_k(t)\left[F^{t,*}-F_k^{t,*}\right]\nonumber\\
	&\quad+2L\alpha^2\sum_{k\in\Pi(t)}w_k(t)\left[F_k^t\left(\boldsymbol{\theta}\left(t\right)\right)-F^{t,*}\right]\nonumber\\
	&\quad-2\alpha\sum_{k\in\Pi(t)}w_k(t)\left[F_k^t\left(\boldsymbol{\theta}\left(t\right)\right)-F_k^t\left(\boldsymbol{\theta}^{t,*}\right)\right]\nonumber\\
	&=\left(1-\mu\alpha\right)||\boldsymbol{\theta}\left(t\right)-\boldsymbol{\theta}^{t,*}||^2+2L\alpha^2\sum_{k\in\Pi(t)}w_k(t)\left[F^{t,*}-F_k^{t,*}\right]\nonumber\\
	&\quad-2\alpha\left[1-L\alpha\right]\sum_{k\in\Pi(t)}w_k(t)\left[F_k^t\left(\boldsymbol{\theta}\left(t\right)\right)-F^{t,*}\right]\nonumber\\
	&\quad-2\alpha\sum_{k\in\Pi(t)}w_k(t)\left[F^{t,*}-F_k^t(\boldsymbol{\theta}^{t,*})\right]\nonumber\\
	&\leq\left(1-\mu\alpha\right)||\boldsymbol{\theta}\left(t\right)-\boldsymbol{\theta}^{t,*}||^2+2\alpha\left[1-L\alpha\right]\varphi_1+2L\alpha^2\varphi_1+2\alpha\varphi_2\label{ieq:lm1_1}\\
	&\leq\left(1-\mu\alpha\right)||\boldsymbol{\theta}\left(t\right)-\boldsymbol{\theta}^{t,*}||^2+\alpha \Omega,\nonumber
\end{align}
where \eqref{ieq:lm1_1} is satisfied if $\alpha<1/L$.

\subsubsection{Proof of \textbf{Lemma \ref{lm:var_g}}}
\label{sec:proof_lm2}
\begin{align*}
	&\mathbb{E}_{\{\xi_k(t)|k\in\Pi(t)\}}\left[||\boldsymbol{g}\left(t\right)-\bar{\boldsymbol{g}}\left(t\right)||^2\right]\nonumber\\
	&=\mathbb{E}_{\{...\}}\left[||\sum_{k\in\Pi(t)}w_k(t)\left\{\left[\nabla F_k^t\left(\boldsymbol{\theta}(t)\right)|_{\xi_k(t)}\right]-\nabla F_k^t\left(\boldsymbol{\theta}(t)\right)\right\}||^2\right]\nonumber\\
	&\leq\mathbb{E}_{\{...\}}\left[\sum_{k\in\Pi(t)}w_k(t)||\left[\nabla F_k^t\left(\boldsymbol{\theta}(t)\right)|_{\xi_k(t)}\right]-\nabla F_k^t\left(\boldsymbol{\theta}(t)\right)||^2\right]\nonumber\\
	&\leq\sum_{k\in\Pi(t)}w_k(t)\left[(C_1-1)||\nabla F_k^t\left(\boldsymbol{\theta}(t)\right)||^2+C_2\right]\nonumber\\
	&\leq C_2+(C_1-1)L^2\sum_{k\in\Pi(t)}w_k(t)||\boldsymbol{\theta}(t)-\boldsymbol{\theta}_k^{t,*}||^2\nonumber\\
	&\leq C_2+2(C_1-1)L^2\left(||\boldsymbol{\theta}(t)-\boldsymbol{\theta}^{t,*}||^2+\Omega/\mu\right),
\end{align*}
\begin{align*}
	&\text{since }\sum_{k\in\Pi(t)}w_k(t)||\boldsymbol{\theta}(t)-\boldsymbol{\theta}_k^{t,*}||^2\nonumber\\
	&\leq 2\sum_{k\in\Pi(t)}w_k(t)\left(||\boldsymbol{\theta}(t)-\boldsymbol{\theta}^{t,*}||^2+|| \boldsymbol{\theta}^{t,*}-\boldsymbol{\theta}_k^{t,*}||^2\right)\nonumber\\
	&\leq 2\sum_{k\in\Pi(t)}w_k(t)\left[||\boldsymbol{\theta}(t)-\boldsymbol{\theta}^{t,*}||^2+\frac{2}{\mu}\left(F_k^t\left(\boldsymbol{\theta}^{t,*}\right)-F_k^t\left(\boldsymbol{\theta}_k^{t,*}\right)\right)\right]\nonumber\\
	&\leq2\left(||\boldsymbol{\theta}(t)-\boldsymbol{\theta}^{t,*}||^2+\Omega/\mu\right).
\end{align*}

\subsection{Proof of \textbf{Lemma \ref{lm:sr-conv}}}
\label{sec:proof_lm3}
A direct calculation gives
\begin{align*}
	&\sum_{i=1}^{t}\frac{\kappa^{t-i}}{i+1}=\sum_{i=1}^{\big\lceil\sqrt{t}\big\rceil\triangleq a_t}\frac{\kappa^{t-i}}{i+1}+\sum_{i=a_t+1}^{t}\frac{\kappa^{t-i}}{i+1}\nonumber\\
	&\leq\frac{a_t\kappa^{t-a_t}}{2}+\frac{1}{a_t}\sum_{i=a_t+1}^{t}\kappa^{t-i}=\frac{a_t\kappa^{t-a_t}}{2}+\frac{\kappa^{-1}-\kappa^{t-a_t-1}}{a_t(\kappa^{-1}-1)}\rightarrow 0.
\end{align*}
(See  Theorem 3.20(d) in \cite{rudin1964principles} for the limit $a_t\kappa^{t-a_t}\rightarrow 0$.)
	
	\bibliographystyle{IEEEtran}
	\bibliography{ref.bib}
	\begin{IEEEbiography}[{\includegraphics[width=1in,height=1.25in,clip,keepaspectratio]{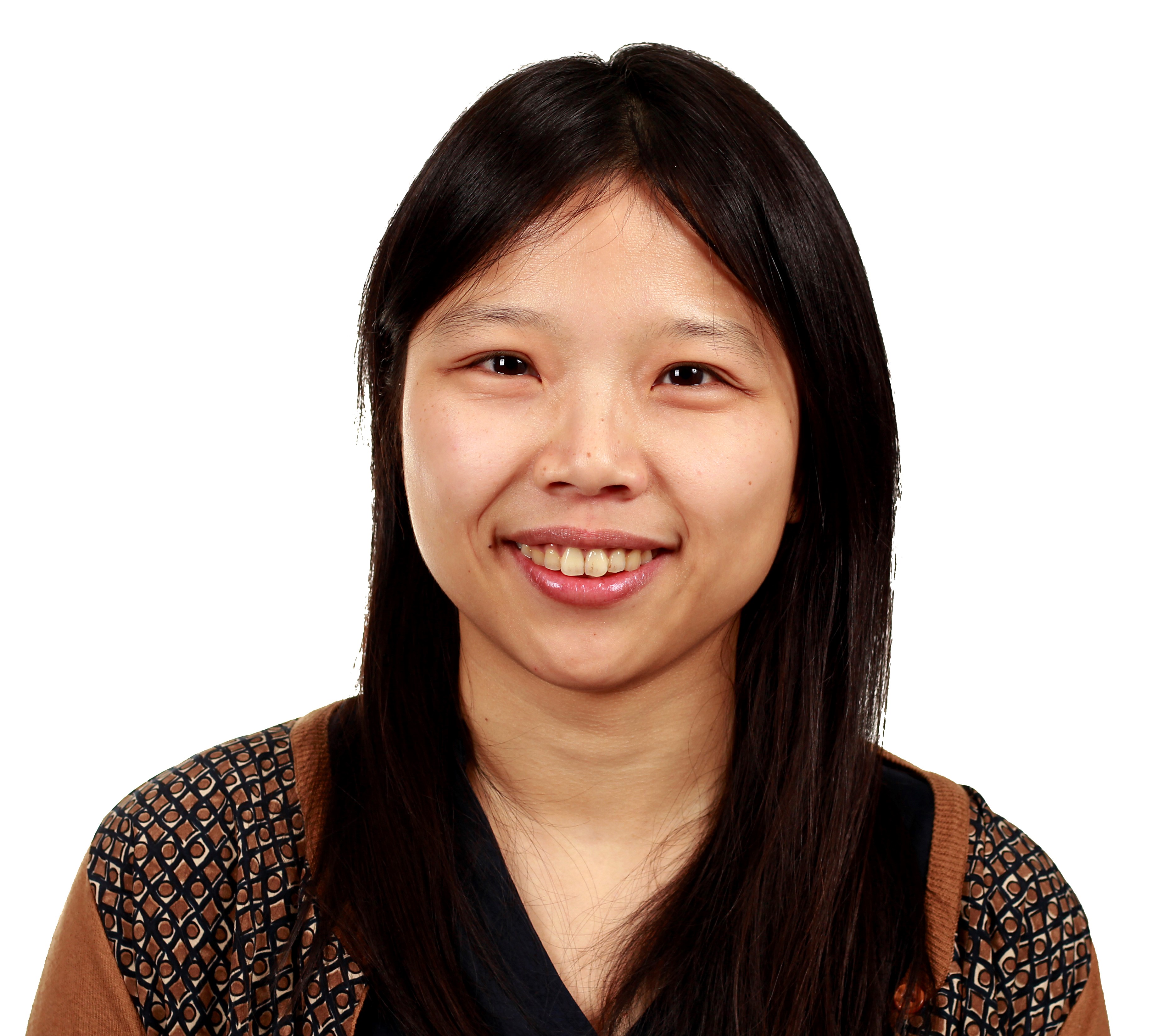}}]
		{\bfseries Chung-Hsuan Hu} received the B.Sc. degree in electronics and electrical engineering, and M.Sc. degree in communications engineering from National Yang Ming Chiao Tung University (NYCU), Taiwan, in 2010 and 2012, respectively. From 2013 to 2020, she worked as a communication systems engineer with MediaTek Inc., Taiwan. Currently, she is pursuing the Ph.D. degree with the Division of Communication Systems, Department of Electrical Engineering, Linköping University, Sweden. Her research interests include distributed learning systems and wireless communications.
	\end{IEEEbiography}
	\begin{IEEEbiography}[{\includegraphics[width=1in,height=1.25in,clip,keepaspectratio]{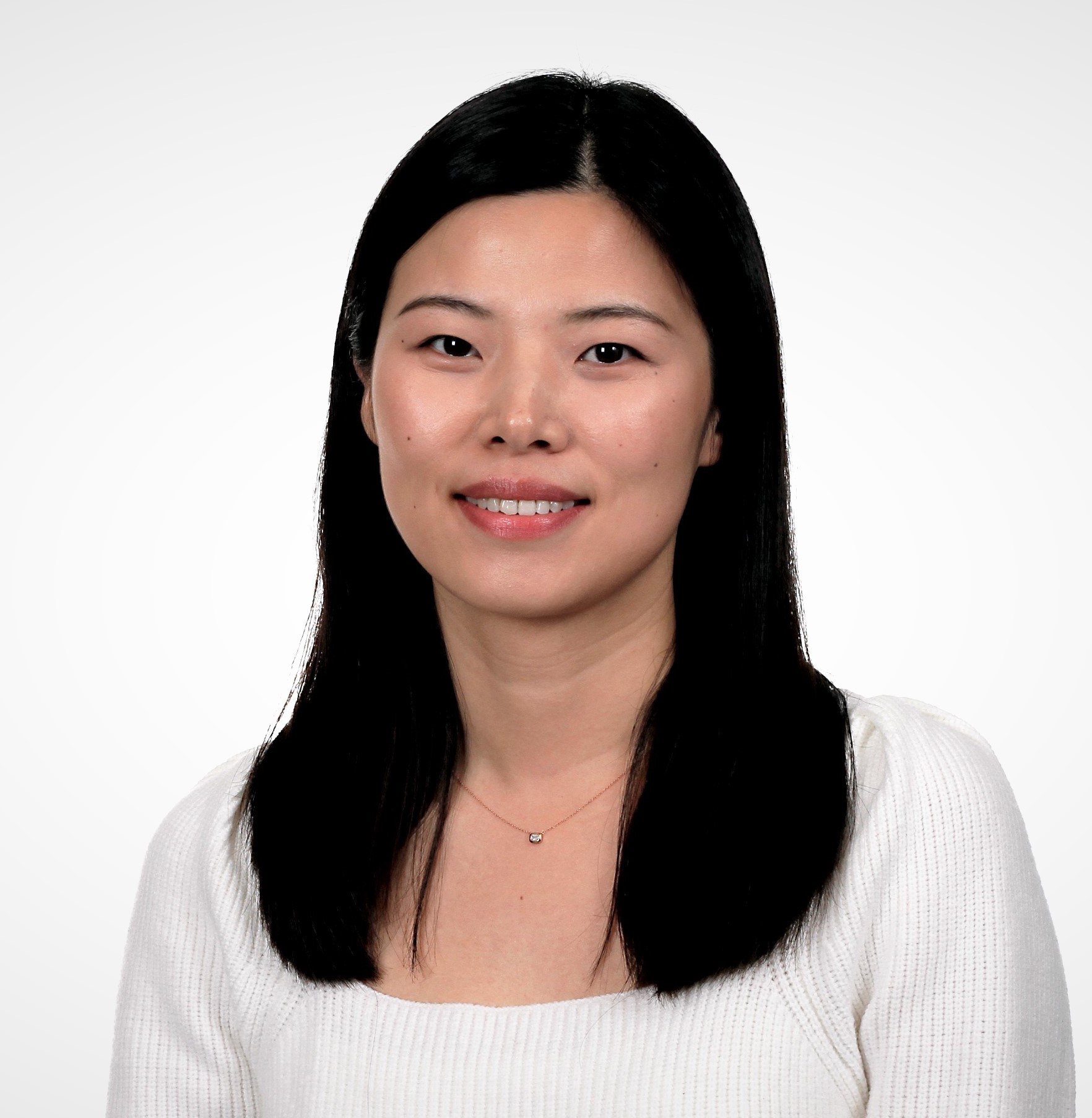}}]
		{\bfseries Zheng Chen} (Member, IEEE) is an Assistant Professor with the Department of Electrical Engineering, Linköping University, Sweden. She received the B.Sc. degree from Huazhong University of Science and Technology (HUST), China, in 2011. Then, she received the M.Sc. and Ph.D. degrees from CentraleSup\'{e}lec, Universit\'{e} Paris-Saclay, France, in 2013 and 2016, respectively. Since January 2017, she has been with Link\"{o}ping University, Sweden. Her main research interests include wireless communications, distributed intelligent systems, and complex networks. 
		She was the recipient of the 2020 IEEE Communications Society Young Author Best Paper Award. She is currently an Associate Editor of the IEEE Transactions on Wireless Communications, IEEE Transactions on Communications, and IEEE Transactions on Green Communications and Networking.
	\end{IEEEbiography}
	\begin{IEEEbiography}[{\includegraphics[width=1in,height=1.25in,clip,keepaspectratio]{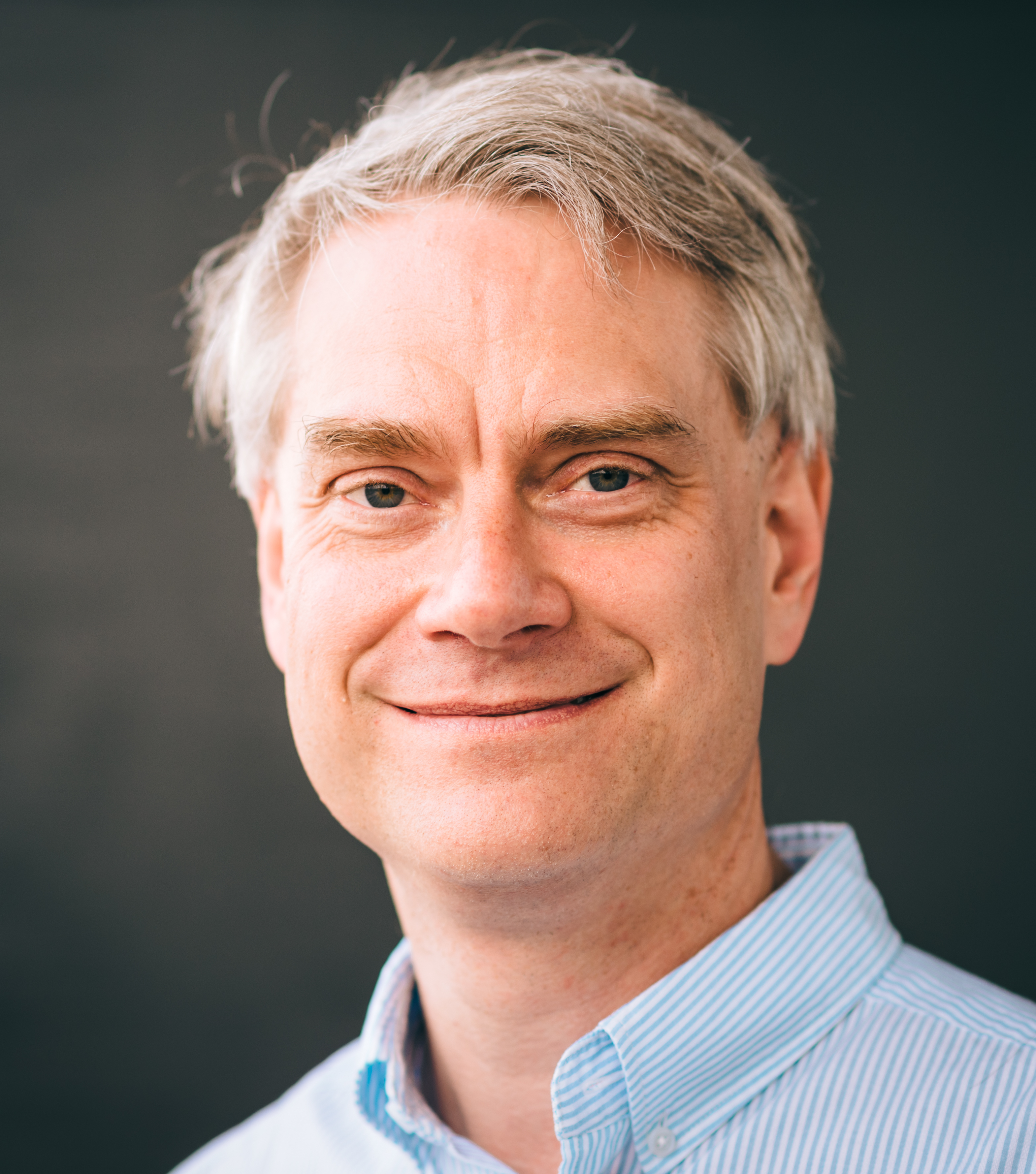}}]
		{\bfseries Erik G. Larsson} (Fellow) received the Ph.D. degree from Uppsala University,
		Uppsala, Sweden, in 2002.  He is currently Professor of Communication
		Systems at Link\"oping University (LiU) in Link\"oping, Sweden. He was
		with the KTH Royal Institute of Technology in Stockholm, Sweden, the
		George Washington University, USA, the University of Florida, USA, and
		Ericsson Research, Sweden.  His main professional interests are within
		wireless communications, signal processing, and network science. He 
		co-authored \emph{Space-Time Block Coding for  Wireless Communications} (Cambridge University Press, 2003) 
		and \emph{Fundamentals of Massive MIMO} (Cambridge University Press, 2016). 	
		
		He served as  chair  of the IEEE Signal Processing Society SPCOM technical committee (2015--2016), 
		chair of  the \emph{IEEE Wireless  Communications Letters} steering committee (2014--2015), 
		member of the  \emph{IEEE Transactions on Wireless Communications}    steering committee (2019-2022),
		General and Technical Chair of the Asilomar SSC conference (2015, 2012), 
		technical co-chair of the IEEE Communication Theory Workshop (2019), 
		and   member of the  IEEE Signal Processing Society Awards Board (2017--2019).
		He was Associate Editor for, among others, the \emph{IEEE Transactions on Communications} (2010-2014), 
		the \emph{IEEE Transactions on Signal Processing} (2006-2010),
		and  the \emph{IEEE Signal  Processing Magazine} (2018-2022).
		
		He received the IEEE Signal Processing Magazine Best Column Award
		twice, in 2012 and 2014, the IEEE ComSoc Stephen O. Rice Prize in
		Communications Theory in 2015, the IEEE ComSoc Leonard G. Abraham
		Prize in 2017, the IEEE ComSoc Best Tutorial Paper Award in 2018, 
		the IEEE ComSoc Fred W. Ellersick Prize in 2019, and the
		2023 IEEE SPS Donald G. Fink Overview Paper Award.
	\end{IEEEbiography}
\end{document}

%% file: figure/sameRatioAccuIid-sglCol.tex
%
%
\definecolor{mycolor1}{rgb}{0.00000,0.44700,0.74100}%
\definecolor{mycolor2}{rgb}{0.85000,0.32500,0.09800}%
\definecolor{mycolor3}{rgb}{0.92900,0.69400,0.12500}%
\definecolor{mycolor4}{rgb}{0.49400,0.18400,0.55600}%
\definecolor{mycolor5}{rgb}{0.46600,0.67400,0.18800}%
\definecolor{mycolor6}{rgb}{0.30100,0.74500,0.93300}%
\begin{tikzpicture}
\begin{axis}[%
width=.64\linewidth,
height=.42\linewidth,
scale only axis,
xmin=0,
xmax=40,
xlabel style={font=\color{white!15!black}},
xlabel={Global iteration ($t$)},
ymin=0,
ymax=100,
ylabel style={font=\color{white!15!black}},
ylabel={Test accuracy ($\%$)},
axis background/.style={fill=white},
xmajorgrids,
ymajorgrids,
legend style={nodes={scale=0.95, transform shape}, at={(1.3,1.1)}, anchor=south, legend cell align=left, align=left, draw=white!15!black},
legend columns=6
]
\addplot [color=mycolor1, line width=1.0pt]
  table[row sep=crcr]{%
1	10.5241666666667\\
2	11.7341666666667\\
3	13.9008333333333\\
4	14.7241666666667\\
5	18\\
6	20.8566666666667\\
7	23.1858333333333\\
8	27.4866666666667\\
9	32.2491666666667\\
10	35.8341666666667\\
11	40.5266666666667\\
12	46.3333333333333\\
13	49.1333333333333\\
14	53.7058333333333\\
15	58.8016666666667\\
16	62.63\\
17	66.0191666666667\\
18	69.6383333333333\\
19	71.4791666666667\\
20	75.1041666666667\\
21	78.5075\\
22	80.3316666666667\\
23	80.7891666666667\\
24	82.6808333333333\\
25	83.06\\
26	83.615\\
27	84.3708333333333\\
28	85.1925\\
29	86.1441666666667\\
30	86.3816666666667\\
31	86.8483333333333\\
32	87.2075\\
33	87.4458333333333\\
34	88.0175\\
35	88.3675\\
36	88.6641666666667\\
37	89.0925\\
38	89.3716666666667\\
39	89.5475\\
40	89.7716666666667\\
};
\addlegendentry{rdm, $\frac{\zeta}{K}=0.075$}

\addplot [color=mycolor2, line width=1.0pt, mark=x, mark repeat=3, mark options={solid, mycolor2}]
  table[row sep=crcr]{%
1	11.6333333333333\\
2	13.9558333333333\\
3	18.0025\\
4	23.0925\\
5	28.1866666666667\\
6	35.5575\\
7	41.1766666666667\\
8	48.2733333333333\\
9	53.8025\\
10	58.9625\\
11	62.3758333333333\\
12	66.1075\\
13	69.3841666666667\\
14	72.9875\\
15	73.9083333333333\\
16	75.8091666666667\\
17	77.4275\\
18	77.4066666666667\\
19	79.6241666666667\\
20	80.9241666666667\\
21	82.9625\\
22	83.345\\
23	83.8625\\
24	84.5566666666667\\
25	84.4408333333333\\
26	85.2416666666667\\
27	85.3108333333333\\
28	85.4291666666667\\
29	85.1383333333333\\
30	86\\
31	86.035\\
32	86.5183333333333\\
33	87.67\\
34	87.7425\\
35	88.3166666666667\\
36	88.8625\\
37	89.31\\
38	89.8908333333333\\
39	90.2333333333333\\
40	90.435\\
};
\addlegendentry{prop, $\frac{\zeta}{K}=0.075$}

\addplot [color=mycolor3, dotted, line width=1.0pt]
  table[row sep=crcr]{%
1	10.6675\\
2	12.5916666666667\\
3	14.4708333333333\\
4	17.7425\\
5	21.7375\\
6	25.3541666666667\\
7	29.0658333333333\\
8	35.6225\\
9	41.8533333333333\\
10	48.2616666666667\\
11	52.7358333333333\\
12	56.3708333333333\\
13	62.96\\
14	68.1725\\
15	70.1008333333333\\
16	74.0666666666667\\
17	75.97\\
18	77.8816666666667\\
19	80.245\\
20	81.6233333333333\\
21	84.1275\\
22	85.0291666666667\\
23	85.3116666666667\\
24	86.0475\\
25	86.4083333333333\\
26	86.98\\
27	87.4825\\
28	87.6591666666667\\
29	88.0466666666667\\
30	88.3383333333333\\
31	88.4241666666667\\
32	88.7508333333333\\
33	89.0241666666667\\
34	89.265\\
35	89.4425\\
36	89.6725\\
37	89.9458333333333\\
38	90.1208333333333\\
39	90.3191666666667\\
40	90.4433333333333\\
};
\addlegendentry{rdm, $\frac{\zeta}{K}=0.125$}

\addplot [color=mycolor4, dotted, line width=1.0pt, mark=x, mark repeat=3, mark options={solid, mycolor4}]
  table[row sep=crcr]{%
1	11.3775\\
2	14.6416666666667\\
3	18.83\\
4	24.7\\
5	31.9658333333333\\
6	39.0275\\
7	45.7233333333333\\
8	51.7566666666667\\
9	57.535\\
10	62.8583333333333\\
11	66.4983333333333\\
12	70.3566666666667\\
13	73.4091666666667\\
14	75.7391666666667\\
15	77.6858333333333\\
16	79.3533333333333\\
17	80.4058333333333\\
18	81.1533333333333\\
19	82.72\\
20	83.4308333333333\\
21	85.2366666666667\\
22	85.5916666666667\\
23	85.8583333333333\\
24	86.285\\
25	86.8458333333333\\
26	87.1316666666667\\
27	87.2425\\
28	87.4591666666667\\
29	87.8958333333333\\
30	88.0883333333333\\
31	88.3508333333333\\
32	88.6183333333333\\
33	88.865\\
34	88.9225\\
35	89.3325\\
36	89.5875\\
37	89.9808333333333\\
38	90.3583333333333\\
39	90.6083333333333\\
40	90.7058333333333\\
};
\addlegendentry{prop, $\frac{\zeta}{K}=0.125$}

\addplot [color=mycolor5, dashdotted, line width=1.0pt]
  table[row sep=crcr]{%
1	10.6516666666667\\
2	11.6491666666667\\
3	13.0633333333333\\
4	14.4816666666667\\
5	15.1325\\
6	16.9666666666667\\
7	20.4225\\
8	21.7783333333333\\
9	25.8275\\
10	28.7858333333333\\
11	32.7991666666667\\
12	36.3416666666667\\
13	39.6175\\
14	41.7825\\
15	47.4958333333333\\
16	51.1866666666667\\
17	55.825\\
18	58.0358333333333\\
19	62.0658333333333\\
20	65.4775\\
21	69.5908333333333\\
22	71.9133333333333\\
23	74.1383333333333\\
24	76.0633333333333\\
25	77.1441666666667\\
26	79.7175\\
27	80.0725\\
28	81.6066666666667\\
29	82.0991666666667\\
30	83.2158333333333\\
31	84.0783333333333\\
32	84.8725\\
33	85.5591666666667\\
34	86.5241666666667\\
35	86.93\\
36	87.3658333333333\\
37	87.9533333333333\\
38	88.4091666666667\\
39	88.8616666666667\\
40	89.0083333333333\\
};
\addlegendentry{rdm, $\frac{\zeta}{K}=0.05$}

\addplot [color=mycolor6, dashdotted, line width=1.0pt, mark=o, mark repeat=3, mark options={solid, mycolor6}]
  table[row sep=crcr]{%
1	11.445\\
2	14.1833333333333\\
3	18.9883333333333\\
4	25.2983333333333\\
5	32.3025\\
6	39.5416666666667\\
7	46.0633333333333\\
8	52.965\\
9	59.04\\
10	63.91\\
11	67.99\\
12	72.22\\
13	75.3983333333333\\
14	77.93\\
15	80.3025\\
16	81.945\\
17	83.235\\
18	84.4958333333333\\
19	85.4466666666667\\
20	86.3033333333333\\
21	87.2666666666667\\
22	87.645\\
23	87.9833333333333\\
24	88.2525\\
25	88.5466666666667\\
26	88.7358333333333\\
27	88.95\\
28	89.1358333333333\\
29	89.3108333333333\\
30	89.485\\
31	89.6566666666667\\
32	89.8208333333334\\
33	89.9941666666667\\
34	90.1283333333333\\
35	90.2958333333333\\
36	90.4358333333333\\
37	90.5633333333333\\
38	90.7158333333333\\
39	90.84\\
40	90.9558333333333\\
};
\addlegendentry{prop, $\frac{\zeta}{K}=0.35$}
\end{axis}
\end{tikzpicture}%

%% file: figure/sameRatioAccuNiid-sglCol.tex
%
%
\definecolor{mycolor1}{rgb}{0.00000,0.44700,0.74100}%
\definecolor{mycolor2}{rgb}{0.85000,0.32500,0.09800}%
\definecolor{mycolor3}{rgb}{0.92900,0.69400,0.12500}%
\definecolor{mycolor4}{rgb}{0.49400,0.18400,0.55600}%
\definecolor{mycolor5}{rgb}{0.46600,0.67400,0.18800}%
\definecolor{mycolor6}{rgb}{0.30100,0.74500,0.93300}%
\begin{tikzpicture}

\begin{axis}[%
width=.64\linewidth,
height=.42\linewidth,
scale only axis,
xmin=0,
xmax=40,
xlabel style={font=\color{white!15!black}},
xlabel={Global iteration ($t$)},
ymin=10,
ymax=80,
ylabel style={font=\color{white!15!black}},
ylabel={Test accuracy ($\%$)},
axis background/.style={fill=white},
xmajorgrids,
ymajorgrids,
]
\addplot [color=mycolor1, line width=1.0pt]
  table[row sep=crcr]{%
1	10.6233333333333\\
2	10.8425\\
3	11.4466666666667\\
4	11.7916666666667\\
5	12.175\\
6	12.9041666666667\\
7	14.2083333333333\\
8	15.0466666666667\\
9	15.7566666666667\\
10	17.615\\
11	18.9066666666667\\
12	19.775\\
13	21.465\\
14	22.8833333333333\\
15	25.3141666666667\\
16	27.5375\\
17	29.9841666666667\\
18	31.3275\\
19	34.7775\\
20	35.6708333333333\\
21	42.3841666666667\\
22	44.7333333333333\\
23	46.2383333333333\\
24	47.7133333333333\\
25	50.85\\
26	52.1933333333333\\
27	52.7716666666667\\
28	55.3266666666667\\
29	56.1166666666667\\
30	58.0016666666667\\
31	59.8008333333333\\
32	61.1791666666667\\
33	61.6691666666667\\
34	63.5575\\
35	63.6258333333333\\
36	65.1733333333333\\
37	67.0575\\
38	67.6933333333333\\
39	67.6891666666667\\
40	68.1333333333333\\
};

\addplot [color=mycolor2, line width=1.0pt, mark=x, mark repeat=3, mark options={solid, mycolor2}]
  table[row sep=crcr]{%
1	11.1208333333333\\
2	11.4591666666667\\
3	12.7083333333333\\
4	13.8\\
5	15.0675\\
6	17.385\\
7	20.185\\
8	21.8658333333333\\
9	23.8275\\
10	27.0016666666667\\
11	28.4283333333333\\
12	31.5783333333333\\
13	32.7433333333333\\
14	34.295\\
15	37.5933333333333\\
16	38.1666666666667\\
17	40.1566666666667\\
18	42.2975\\
19	44.1808333333333\\
20	45.04\\
21	51.9291666666667\\
22	51.9225\\
23	52.595\\
24	54.1616666666667\\
25	54.75\\
26	55.6591666666667\\
27	56.6075\\
28	57.9725\\
29	57.5891666666667\\
30	58.255\\
31	59.945\\
32	62.5766666666667\\
33	63.4333333333333\\
34	65.175\\
35	65.2416666666667\\
36	66.3533333333333\\
37	67.8658333333333\\
38	68.785\\
39	68.4225\\
40	69.8583333333333\\
};

\addplot [color=mycolor3, dotted, line width=1.0pt]
  table[row sep=crcr]{%
1	10.6491666666667\\
2	10.9575\\
3	11.515\\
4	12.1508333333333\\
5	13.2375\\
6	14.0366666666667\\
7	15.3108333333333\\
8	16.8758333333333\\
9	18.375\\
10	20.7241666666667\\
11	22.7008333333333\\
12	25.125\\
13	27.9166666666667\\
14	29.8841666666667\\
15	32.3425\\
16	36.2016666666667\\
17	38.9791666666667\\
18	40.7358333333333\\
19	43.6425\\
20	45.9875\\
21	51.9025\\
22	53.8475\\
23	55.6958333333333\\
24	56.8783333333333\\
25	59.5708333333333\\
26	61.4566666666667\\
27	62.3808333333333\\
28	63.4558333333333\\
29	65.2966666666667\\
30	65.9725\\
31	67.4516666666667\\
32	68.735\\
33	69.1908333333333\\
34	70.265\\
35	71.2633333333333\\
36	72.2183333333333\\
37	73.005\\
38	73.0333333333333\\
39	73.9275\\
40	73.7391666666667\\
};

\addplot [color=mycolor4, dotted, line width=1.0pt, mark=x, mark repeat=3, mark options={solid, mycolor4}]
  table[row sep=crcr]{%
1	11.1716666666667\\
2	11.8625\\
3	13.425\\
4	15.2108333333333\\
5	18.0275\\
6	19.9958333333333\\
7	23.1566666666667\\
8	25.2766666666667\\
9	28.0958333333333\\
10	30.5758333333333\\
11	33.54\\
12	35.8216666666667\\
13	37.3941666666667\\
14	40.8183333333333\\
15	42.5175\\
16	44.2991666666667\\
17	46.1066666666667\\
18	47.5791666666667\\
19	49.4216666666667\\
20	50.4825\\
21	55.7066666666667\\
22	55.9033333333333\\
23	58.035\\
24	58.6058333333333\\
25	59.0183333333333\\
26	60.5083333333333\\
27	61.5816666666667\\
28	63.0733333333333\\
29	63.9\\
30	65.1925\\
31	65.7183333333333\\
32	67.0308333333333\\
33	68.495\\
34	69.0033333333333\\
35	70.1958333333333\\
36	70.7041666666667\\
37	71.4433333333333\\
38	72.2566666666667\\
39	72.1358333333333\\
40	72.1258333333333\\
};

\addplot [color=mycolor5, dashdotted, line width=1.0pt]
  table[row sep=crcr]{%
1	10.94\\
2	10.9933333333333\\
3	10.9216666666667\\
4	11.3025\\
5	11.8083333333333\\
6	12.2483333333333\\
7	12.4391666666667\\
8	12.8933333333333\\
9	13.5908333333333\\
10	15.2183333333333\\
11	15.42\\
12	17.0058333333333\\
13	17.9841666666667\\
14	19.0433333333333\\
15	20.5966666666667\\
16	22.005\\
17	22.6916666666667\\
18	25.505\\
19	26.4033333333333\\
20	27.9308333333333\\
21	33.5266666666667\\
22	35.1525\\
23	37.2\\
24	39.7591666666667\\
25	40.6875\\
26	41.9591666666667\\
27	44.4683333333333\\
28	46.3\\
29	47.0241666666667\\
30	49.3908333333333\\
31	50.0833333333333\\
32	52.4483333333333\\
33	53.8241666666667\\
34	54.7625\\
35	56.1966666666667\\
36	57.5083333333333\\
37	58.7466666666667\\
38	59.9666666666667\\
39	60.6925\\
40	61.0933333333333\\
};

\addplot [color=mycolor6, dashdotted, line width=1.0pt, mark=o, mark repeat=3, mark options={solid, mycolor6}]
  table[row sep=crcr]{%
1	10.9541666666667\\
2	11.9841666666667\\
3	13.28\\
4	15.7108333333333\\
5	18.2\\
6	21.1183333333333\\
7	24.1291666666667\\
8	27.24\\
9	30.6966666666667\\
10	33.8191666666667\\
11	36.6291666666667\\
12	39.3666666666667\\
13	42.3733333333333\\
14	44.5816666666667\\
15	47.26\\
16	49.6025\\
17	51.9108333333333\\
18	53.5691666666667\\
19	55.9208333333333\\
20	57.7558333333333\\
21	60.3325\\
22	61.66\\
23	63.0633333333333\\
24	64.2733333333333\\
25	64.97\\
26	65.7475\\
27	67.1958333333333\\
28	68.0675\\
29	69.1416666666667\\
30	70.0841666666667\\
31	70.9441666666667\\
32	71.9891666666667\\
33	72.7416666666667\\
34	73.4225\\
35	74.4366666666667\\
36	75.2441666666667\\
37	75.7316666666667\\
38	76.4675\\
39	77.0283333333333\\
40	77.5466666666667\\
};

\end{axis}
\end{tikzpicture}%

%% file: figure/sameRatioEngIid-sglCol.tex
%
%
\definecolor{mycolor1}{rgb}{0.00000,0.44700,0.74100}%
\definecolor{mycolor2}{rgb}{0.85000,0.32500,0.09800}%
\definecolor{mycolor3}{rgb}{0.92900,0.69400,0.12500}%
\definecolor{mycolor4}{rgb}{0.49400,0.18400,0.55600}%
\definecolor{mycolor5}{rgb}{0.46600,0.67400,0.18800}%
\definecolor{mycolor6}{rgb}{0.30100,0.74500,0.93300}%
\begin{tikzpicture}
\begin{semilogyaxis}[%
width=.8\linewidth,
height=.62\linewidth,
xmin=0,
xmax=40,
xlabel style={font=\color{white!15!black}},
xlabel={Global iteration ($t$)},
ymin=0.284535690808333,
ymax=4.71707576869167,
ylabel style={font=\color{white!15!black}},
ylabel={Average $E_k(t)$},
axis background/.style={fill=white},
xmajorgrids,
ymajorgrids,
legend style={at={(1.4,1.1)}, anchor=south, legend cell align=left, align=left, draw=white!15!black},
legend columns=4
]
\addplot [color=mycolor1, line width=1.0pt]
  table[row sep=crcr]{%
1	1.10997310360833\\
2	1.33958355981667\\
3	1.36289455758333\\
4	1.456723534875\\
5	1.56575842031667\\
6	1.54040565691667\\
7	1.56642608163333\\
8	1.69790781788333\\
9	1.841958468725\\
10	1.764026283125\\
11	1.96504510741667\\
12	1.993562575325\\
13	1.99362321769167\\
14	2.14751248263333\\
15	2.17114898166667\\
16	2.26916861071667\\
17	2.49170680094167\\
18	2.42301608614167\\
19	2.431380257675\\
20	2.58791073539167\\
21	2.54077197204167\\
22	2.65918122089167\\
23	2.694074384575\\
24	2.73574096158333\\
25	2.67807468104167\\
26	2.680743437425\\
27	2.74161771239167\\
28	2.862505610575\\
29	2.7070549752\\
30	2.729625766\\
31	2.710460271525\\
32	2.799955305325\\
33	2.7851943863\\
34	2.629585093725\\
35	2.78085483410833\\
36	2.72018173640833\\
37	2.76953572398333\\
38	2.80336448009167\\
39	2.80457147383333\\
40	2.6181076179\\
};
\addlegendentry{rdm, $\zeta/K=0.075$}

\addplot [color=mycolor2, line width=1.0pt, mark=x, mark repeat=5, mark options={solid, mycolor2}]
  table[row sep=crcr]{%
1	0.284535690808333\\
2	0.349913114241667\\
3	0.392827659158333\\
4	0.341343115725\\
5	0.3486834609\\
6	0.359031369983333\\
7	0.335583013141667\\
8	0.310049581708333\\
9	0.361284270966667\\
10	0.317824173925\\
11	0.316089126658333\\
12	0.321603089783333\\
13	0.336543114991667\\
14	0.318821876791667\\
15	0.340190892125\\
16	0.307646350975\\
17	0.320911225583333\\
18	0.318138664425\\
19	0.302602126041667\\
20	0.341356028275\\
21	0.334637196133333\\
22	0.300534122375\\
23	0.3207003305\\
24	0.373168177\\
25	0.328040473316667\\
26	0.359365184616667\\
27	0.341638096433333\\
28	0.386643563116667\\
29	0.385084178475\\
30	0.353697702391667\\
31	0.384051015566667\\
32	0.366592002391667\\
33	0.357348298366667\\
34	0.316400263741667\\
35	0.307463244016667\\
36	0.29480695765\\
37	0.329237663966667\\
38	0.320691716725\\
39	0.319625685625\\
40	0.300081026691667\\
};
\addlegendentry{prop, $\zeta/K=0.075$}

\addplot [color=mycolor3, dotted, line width=1.0pt]
  table[row sep=crcr]{%
1	1.9546041623\\
2	2.17807523196667\\
3	2.33073746568333\\
4	2.46401572740833\\
5	2.56932939580833\\
6	2.70521413345\\
7	2.88426087968333\\
8	2.81636523745\\
9	3.08499421576667\\
10	3.17828066824167\\
11	3.236915569975\\
12	3.25482686773333\\
13	3.677597436975\\
14	3.72589712655\\
15	3.616038899475\\
16	3.889443872375\\
17	3.73481607681667\\
18	4.168737971425\\
19	4.154637912725\\
20	4.186937554475\\
21	4.26663648023333\\
22	4.38393966445833\\
23	4.42865682848333\\
24	4.48328488765\\
25	4.50314327740833\\
26	4.520682814925\\
27	4.61112011745833\\
28	4.51372330743333\\
29	4.53242052021667\\
30	4.443473827375\\
31	4.5566804758\\
32	4.52621770273333\\
33	4.71707576869167\\
34	4.536375110675\\
35	4.3906808828\\
36	4.408398959325\\
37	4.59034438974167\\
38	4.53359074449167\\
39	4.58101063875\\
40	4.64824202465833\\
};
\addlegendentry{rdm, $\zeta/K=0.125$}

\addplot [color=mycolor4, dotted, line width=1.0pt, mark=x, mark repeat=5, mark options={solid, mycolor4}]
  table[row sep=crcr]{%
1	0.489177235541667\\
2	0.782828848008333\\
3	0.733040333325\\
4	0.675647289408333\\
5	0.60000787025\\
6	0.509863394716667\\
7	0.524908731783333\\
8	0.476568182483333\\
9	0.480490586175\\
10	0.426745677291667\\
11	0.453984293683333\\
12	0.467100417816667\\
13	0.418401700641667\\
14	0.457859550941667\\
15	0.497873801366667\\
16	0.472469684108333\\
17	0.477410585516667\\
18	0.469605817408333\\
19	0.48496741815\\
20	0.497707403441667\\
21	0.504764553941667\\
22	0.51377221195\\
23	0.53931396385\\
24	0.553378831033333\\
25	0.550598753258333\\
26	0.507122221058333\\
27	0.539602285591667\\
28	0.497564908583333\\
29	0.460163992575\\
30	0.442069056983333\\
31	0.402321113416667\\
32	0.43372006145\\
33	0.403435033616667\\
34	0.407305527541667\\
35	0.40802728665\\
36	0.421295633183333\\
37	0.441671192708333\\
38	0.45673467695\\
39	0.45418853585\\
40	0.455797302733333\\
};
\addlegendentry{prop, $\zeta/K=0.125$}

\addplot [color=mycolor5, dashdotted, line width=1.0pt]
  table[row sep=crcr]{%
1	0.754068054983333\\
2	0.908154186241667\\
3	0.90094711545\\
4	1.03536637905\\
5	0.99115276685\\
6	1.06998839514167\\
7	1.15002562883333\\
8	1.13040665488333\\
9	1.203696225425\\
10	1.28291748703333\\
11	1.32225032546667\\
12	1.3170590567\\
13	1.41199910845\\
14	1.33374724083333\\
15	1.4443482357\\
16	1.5337813582\\
17	1.5749216377\\
18	1.66331856465\\
19	1.58720487363333\\
20	1.62094420790833\\
21	1.777477170125\\
22	1.74062080544167\\
23	1.81411315331667\\
24	1.7537671822\\
25	1.74697493900833\\
26	1.775023896675\\
27	1.78643035488333\\
28	1.783601791225\\
29	1.86481379494167\\
30	1.79952694713333\\
31	1.78170875798333\\
32	1.78035602825\\
33	1.754706036175\\
34	1.86833661004167\\
35	1.7083502041\\
36	1.8463671145\\
37	1.925262611025\\
38	1.77930491820833\\
39	1.79205695305\\
40	1.82523977045833\\
};
\addlegendentry{rdm, $\zeta/K=0.05$}

\addplot [color=mycolor6, dashdotted, line width=1.0pt, mark=o, mark repeat=5, mark options={solid, mycolor6}]
  table[row sep=crcr]{%
1	1.33826204024167\\
2	2.91828415359167\\
3	1.855640537775\\
4	1.11512502150833\\
5	0.886361163558333\\
6	0.787201308366667\\
7	0.776708008591667\\
8	0.75325199985\\
9	0.777198720208333\\
10	0.762347946925\\
11	0.769202139283333\\
12	0.797106769458333\\
13	0.778063601133333\\
14	0.862583198166667\\
15	0.856692124716667\\
16	0.799965559033333\\
17	0.8908367542\\
18	0.880833469725\\
19	0.896619455133333\\
20	0.94764776775\\
21	0.980703498808333\\
22	0.989047496691667\\
23	1.05804407054167\\
24	0.973407958616667\\
25	0.971272001208333\\
26	0.962986010383333\\
27	0.985382158816667\\
28	0.982904886908333\\
29	1.01188389011667\\
30	1.03608353083333\\
31	1.130988361675\\
32	1.03004406273333\\
33	1.010122898825\\
34	1.04314541483333\\
35	1.07408923078333\\
36	1.03942818233333\\
37	0.974892734033333\\
38	1.03467461401667\\
39	1.0553878329\\
40	0.997943288241667\\
};
\addlegendentry{prop, $\zeta/K=0.35$}

\addplot [color=black, dotted, line width=1.0pt, mark=square, mark repeat=10, mark options={solid, rotate=270, black}]
  table[row sep=crcr]{%
1	1.01980909656292\\
2	1.01980909656292\\
3	1.01980909656292\\
4	1.01980909656292\\
5	1.01980909656292\\
6	1.01980909656292\\
7	1.01980909656292\\
8	1.01980909656292\\
9	1.01980909656292\\
10	1.01980909656292\\
11	1.01980909656292\\
12	1.01980909656292\\
13	1.01980909656292\\
14	1.01980909656292\\
15	1.01980909656292\\
16	1.01980909656292\\
17	1.01980909656292\\
18	1.01980909656292\\
19	1.01980909656292\\
20	1.01980909656292\\
21	1.01980909656292\\
22	1.01980909656292\\
23	1.01980909656292\\
24	1.01980909656292\\
25	1.01980909656292\\
26	1.01980909656292\\
27	1.01980909656292\\
28	1.01980909656292\\
29	1.01980909656292\\
30	1.01980909656292\\
31	1.01980909656292\\
32	1.01980909656292\\
33	1.01980909656292\\
34	1.01980909656292\\
35	1.01980909656292\\
36	1.01980909656292\\
37	1.01980909656292\\
38	1.01980909656292\\
39	1.01980909656292\\
40	1.01980909656292\\
};
\addlegendentry{Mean:prop, $\zeta/K=0.35$}

\addplot [color=black, line width=1.0pt, mark=square, mark repeat=10, mark options={solid, rotate=270, black}]
  table[row sep=crcr]{%
1	1.52850841354688\\
2	1.52850841354688\\
3	1.52850841354688\\
4	1.52850841354688\\
5	1.52850841354688\\
6	1.52850841354688\\
7	1.52850841354688\\
8	1.52850841354688\\
9	1.52850841354688\\
10	1.52850841354688\\
11	1.52850841354688\\
12	1.52850841354688\\
13	1.52850841354688\\
14	1.52850841354688\\
15	1.52850841354688\\
16	1.52850841354688\\
17	1.52850841354688\\
18	1.52850841354688\\
19	1.52850841354688\\
20	1.52850841354688\\
21	1.52850841354688\\
22	1.52850841354688\\
23	1.52850841354688\\
24	1.52850841354688\\
25	1.52850841354688\\
26	1.52850841354688\\
27	1.52850841354688\\
28	1.52850841354688\\
29	1.52850841354688\\
30	1.52850841354688\\
31	1.52850841354688\\
32	1.52850841354688\\
33	1.52850841354688\\
34	1.52850841354688\\
35	1.52850841354688\\
36	1.52850841354688\\
37	1.52850841354688\\
38	1.52850841354688\\
39	1.52850841354688\\
40	1.52850841354688\\
};
\addlegendentry{Mean:rdm, $\zeta/K=0.05$}
\end{semilogyaxis}
\end{tikzpicture}%

%% file: figure/sameRatioEngNiid-sglCol.tex
%
%
\definecolor{mycolor1}{rgb}{0.00000,0.44700,0.74100}%
\definecolor{mycolor2}{rgb}{0.85000,0.32500,0.09800}%
\definecolor{mycolor3}{rgb}{0.92900,0.69400,0.12500}%
\definecolor{mycolor4}{rgb}{0.49400,0.18400,0.55600}%
\definecolor{mycolor5}{rgb}{0.46600,0.67400,0.18800}%
\definecolor{mycolor6}{rgb}{0.30100,0.74500,0.93300}%
\begin{tikzpicture}
\begin{semilogyaxis}[%
width=.8\linewidth,
height=.62\linewidth,
xmin=0,
xmax=40,
xlabel style={font=\color{white!15!black}},
xlabel={Global iteration ($t$)},
ymin=0.271889760666667,
ymax=4.70585687310833,
ylabel style={font=\color{white!15!black}},
ylabel={Average $E_k(t)$},
axis background/.style={fill=white},
xmajorgrids,
ymajorgrids,
yminorgrids,
legend style={at={(-1.5,0)}, anchor=north, legend cell align=left, align=left, draw=white!15!black},
]
\addplot [color=mycolor1, line width=1.0pt, forget plot]
  table[row sep=crcr]{%
1	1.09980032416667\\
2	1.33523931628333\\
3	1.266427865275\\
4	1.54843833159167\\
5	1.54813641915833\\
6	1.58335217105\\
7	1.728905490575\\
8	1.68934281580833\\
9	1.83719394479167\\
10	1.90428932671667\\
11	1.94411225679167\\
12	2.02921591218333\\
13	2.007536013875\\
14	2.03955557535\\
15	2.15426591905\\
16	2.188441877625\\
17	2.27444033974167\\
18	2.47285038938333\\
19	2.370103906475\\
20	2.43775361741667\\
21	2.519530611525\\
22	2.62530197231667\\
23	2.58712888260833\\
24	2.70295525945833\\
25	2.66072700563333\\
26	2.73747603326667\\
27	2.742417121775\\
28	2.708163982375\\
29	2.73720360974167\\
30	2.7251946475\\
31	2.65046153634167\\
32	2.92037866574167\\
33	2.83164205490833\\
34	2.66068164121667\\
35	2.68183762794167\\
36	2.75263419354167\\
37	2.79305357921667\\
38	2.75258656831667\\
39	2.764832304575\\
40	2.72059382165833\\
};
\addplot [color=mycolor2, line width=1.0pt, mark=x, mark repeat=5, mark options={solid, mycolor2}, forget plot]
  table[row sep=crcr]{%
1	0.271889760666667\\
2	0.396123084175\\
3	0.411257229258333\\
4	0.392768858158333\\
5	0.38410632705\\
6	0.35882681805\\
7	0.35942196195\\
8	0.326030706291667\\
9	0.315947087483333\\
10	0.306769257275\\
11	0.305094924325\\
12	0.338590250816667\\
13	0.32758735195\\
14	0.321009640241667\\
15	0.336150422975\\
16	0.3074156594\\
17	0.321316631658333\\
18	0.313103681716667\\
19	0.332241070033333\\
20	0.319488714758333\\
21	0.32557814445\\
22	0.362425515441667\\
23	0.337216639216667\\
24	0.3611914399\\
25	0.334423245283333\\
26	0.328811218308333\\
27	0.358446882108333\\
28	0.353108893533333\\
29	0.358379831783333\\
30	0.348342474091667\\
31	0.3517346816\\
32	0.350229255991667\\
33	0.312532036316667\\
34	0.343550358833333\\
35	0.317048622916667\\
36	0.317217696566667\\
37	0.300957196366667\\
38	0.31647178975\\
39	0.291244987508333\\
40	0.312944238608333\\
};
\addplot [color=mycolor3, dotted, line width=1.0pt, forget plot]
  table[row sep=crcr]{%
1	1.85138434514167\\
2	2.16136319965833\\
3	2.340226050825\\
4	2.64633850181667\\
5	2.53886484619167\\
6	2.80839660143333\\
7	2.83624949950833\\
8	2.95242255318333\\
9	2.989560018675\\
10	3.05279251734167\\
11	3.22889157773333\\
12	3.35242547439167\\
13	3.48127266075833\\
14	3.41884585841667\\
15	3.54154196445833\\
16	3.91800482575\\
17	4.05464079224167\\
18	4.13276035271667\\
19	4.11258977775833\\
20	4.11973385766667\\
21	4.24278136249167\\
22	4.40663338241667\\
23	4.38348541255\\
24	4.51100866796667\\
25	4.54168910140833\\
26	4.59336771893333\\
27	4.49150162453333\\
28	4.45460526866667\\
29	4.51963546444167\\
30	4.55848471391667\\
31	4.458033285875\\
32	4.574109602225\\
33	4.62672316854167\\
34	4.47616483155833\\
35	4.58063939178333\\
36	4.70585687310833\\
37	4.55333989448333\\
38	4.501959540375\\
39	4.6099910645\\
40	4.53259096364167\\
};
\addplot [color=mycolor4, dotted, line width=1.0pt, mark=x, mark repeat=5, mark options={solid, mycolor4}, forget plot]
  table[row sep=crcr]{%
1	0.508430259516667\\
2	0.844339068466667\\
3	0.72931770745\\
4	0.670095331433333\\
5	0.625684020391667\\
6	0.537817380483333\\
7	0.517132079408333\\
8	0.4397515402\\
9	0.46943858935\\
10	0.483741883725\\
11	0.46641616645\\
12	0.462280800033333\\
13	0.468427076558333\\
14	0.486135095791667\\
15	0.462374114858333\\
16	0.4744723848\\
17	0.440429525416667\\
18	0.461040826433333\\
19	0.475787811783333\\
20	0.491746204025\\
21	0.497908312825\\
22	0.548341111008333\\
23	0.556370551108333\\
24	0.541611625366667\\
25	0.583049761633333\\
26	0.541528336533333\\
27	0.553346486\\
28	0.502643839075\\
29	0.463304117983333\\
30	0.429159777083333\\
31	0.401855214991667\\
32	0.392773900683333\\
33	0.396977427966667\\
34	0.420250519433333\\
35	0.408661069275\\
36	0.424412760166667\\
37	0.443240664741667\\
38	0.415214540783333\\
39	0.468309124266667\\
40	0.479417619566667\\
};
\addplot [color=mycolor5, dashdotted, line width=1.0pt, forget plot]
  table[row sep=crcr]{%
1	0.745869614383333\\
2	0.835098709075\\
3	0.920228178266667\\
4	0.993370187383333\\
5	1.01966053115833\\
6	1.04380567365833\\
7	1.08242052988333\\
8	1.2052586925\\
9	1.21747664848333\\
10	1.249705931025\\
11	1.33369697794167\\
12	1.40001215545833\\
13	1.3988403289\\
14	1.35751464258333\\
15	1.41221017340833\\
16	1.47851075301667\\
17	1.586260750075\\
18	1.55990236284167\\
19	1.66041982725833\\
20	1.62986365620833\\
21	1.77801458505\\
22	1.75120058750833\\
23	1.664106762375\\
24	1.83112120565\\
25	1.776125504475\\
26	1.73035161391667\\
27	1.86188133843333\\
28	1.87139646048333\\
29	1.82613318809167\\
30	1.761001272075\\
31	1.86727560766667\\
32	1.853176033875\\
33	1.89382088865833\\
34	1.81017227159167\\
35	1.8070323167\\
36	1.806746644575\\
37	1.8708202812\\
38	1.778490328025\\
39	1.90729242121667\\
40	1.86053025644167\\
};
\addplot [color=mycolor6, dashdotted, line width=1.0pt, mark=o, mark repeat=5, mark options={solid, mycolor6}, forget plot]
  table[row sep=crcr]{%
1	1.30866871794167\\
2	2.86725426133333\\
3	1.90169063824167\\
4	1.15657313004167\\
5	0.876133313858333\\
6	0.764305617233333\\
7	0.751893275591667\\
8	0.766569459983333\\
9	0.729046482258333\\
10	0.760290567033333\\
11	0.756077466241667\\
12	0.759318012725\\
13	0.768982229266667\\
14	0.838898707266667\\
15	0.884933849233333\\
16	0.8843251313\\
17	0.858982990075\\
18	0.92358582155\\
19	0.849496896925\\
20	0.9373285484\\
21	0.962152461041667\\
22	0.930334408533333\\
23	1.06775575526667\\
24	0.982690966925\\
25	0.947682588616667\\
26	0.967594418775\\
27	1.01241967329167\\
28	1.01588697841667\\
29	1.01058256954167\\
30	1.045959559475\\
31	1.08087332804167\\
32	1.03037382351667\\
33	1.03603086066667\\
34	1.044503463\\
35	1.06474200679167\\
36	1.03400862443333\\
37	1.00573760415833\\
38	1.03326395765\\
39	1.09144687776667\\
40	0.98189389415\\
};
\addplot [color=black, dotted, line width=1.0pt, mark=square, mark repeat=10, mark options={solid, black}]
  table[row sep=crcr]{%
1	1.01725722341396\\
2	1.01725722341396\\
3	1.01725722341396\\
4	1.01725722341396\\
5	1.01725722341396\\
6	1.01725722341396\\
7	1.01725722341396\\
8	1.01725722341396\\
9	1.01725722341396\\
10	1.01725722341396\\
11	1.01725722341396\\
12	1.01725722341396\\
13	1.01725722341396\\
14	1.01725722341396\\
15	1.01725722341396\\
16	1.01725722341396\\
17	1.01725722341396\\
18	1.01725722341396\\
19	1.01725722341396\\
20	1.01725722341396\\
21	1.01725722341396\\
22	1.01725722341396\\
23	1.01725722341396\\
24	1.01725722341396\\
25	1.01725722341396\\
26	1.01725722341396\\
27	1.01725722341396\\
28	1.01725722341396\\
29	1.01725722341396\\
30	1.01725722341396\\
31	1.01725722341396\\
32	1.01725722341396\\
33	1.01725722341396\\
34	1.01725722341396\\
35	1.01725722341396\\
36	1.01725722341396\\
37	1.01725722341396\\
38	1.01725722341396\\
39	1.01725722341396\\
40	1.01725722341396\\
};

\addplot [color=black, line width=1.0pt, mark=square, mark repeat=10, mark options={solid, black}]
  table[row sep=crcr]{%
1	1.53592039728792\\
2	1.53592039728792\\
3	1.53592039728792\\
4	1.53592039728792\\
5	1.53592039728792\\
6	1.53592039728792\\
7	1.53592039728792\\
8	1.53592039728792\\
9	1.53592039728792\\
10	1.53592039728792\\
11	1.53592039728792\\
12	1.53592039728792\\
13	1.53592039728792\\
14	1.53592039728792\\
15	1.53592039728792\\
16	1.53592039728792\\
17	1.53592039728792\\
18	1.53592039728792\\
19	1.53592039728792\\
20	1.53592039728792\\
21	1.53592039728792\\
22	1.53592039728792\\
23	1.53592039728792\\
24	1.53592039728792\\
25	1.53592039728792\\
26	1.53592039728792\\
27	1.53592039728792\\
28	1.53592039728792\\
29	1.53592039728792\\
30	1.53592039728792\\
31	1.53592039728792\\
32	1.53592039728792\\
33	1.53592039728792\\
34	1.53592039728792\\
35	1.53592039728792\\
36	1.53592039728792\\
37	1.53592039728792\\
38	1.53592039728792\\
39	1.53592039728792\\
40	1.53592039728792\\
};
\end{semilogyaxis}
\end{tikzpicture}%

%% file: figure/sameRatioDist4TestIid-sglCol1.tex
%
%
\definecolor{mycolor1}{rgb}{0.00000,0.44700,0.74100}%
\definecolor{mycolor2}{rgb}{0.85000,0.32500,0.09800}%
\definecolor{mycolor3}{rgb}{0.92900,0.69400,0.12500}%
\definecolor{mycolor4}{rgb}{0.49400,0.18400,0.55600}%
\definecolor{mycolor5}{rgb}{0.46600,0.67400,0.18800}%
\definecolor{mycolor6}{rgb}{0.30100,0.74500,0.93300}%
\begin{tikzpicture}

\begin{axis}[%
width=.61\linewidth,
height=.42\linewidth,
scale only axis,
xmin=0,
xmax=40,
xlabel style={font=\color{white!15!black}},
xlabel={Global iteration ($t$)},
ymin=0,
ymax=7,
ylabel style={font=\color{white!15!black}},
ylabel={Testing loss},
axis background/.style={fill=white},
xmajorgrids,
ymajorgrids,
legend style={nodes={scale=0.9, transform shape}, legend cell align=left, align=left, draw=white!15!black}
]
\addplot [color=mycolor1, line width=1.0pt]
  table[row sep=crcr]{%
1	3.243325\\
2	4.15065\\
3	4.21616666666667\\
4	3.87119166666667\\
5	3.23173333333333\\
6	3.15385\\
7	3.17479166666667\\
8	2.89258333333333\\
9	2.60481666666667\\
10	2.35645833333333\\
11	2.27110833333333\\
12	2.00108333333333\\
13	1.7885\\
14	1.69073333333333\\
15	1.56050833333333\\
16	1.41263333333333\\
17	1.265225\\
18	1.17919166666667\\
19	1.09048333333333\\
20	1.04249166666667\\
21	0.835883333333333\\
22	0.764533333333333\\
23	0.717733333333333\\
24	0.692891666666667\\
25	0.6542\\
26	0.622091666666667\\
27	0.587658333333333\\
28	0.563441666666667\\
29	0.536616666666667\\
30	0.515375\\
31	0.4975\\
32	0.479941666666667\\
33	0.461525\\
34	0.446791666666667\\
35	0.43435\\
36	0.423933333333333\\
37	0.411066666666667\\
38	0.400708333333333\\
39	0.3924\\
40	0.387391666666667\\
};
\addlegendentry{rdm, uni, $\zeta/K$=0.075}

\addplot [color=mycolor2, line width=1.0pt, mark=x, mark repeat=3, mark options={solid, mycolor2}]
  table[row sep=crcr]{%
1	3.1346\\
2	3.66496666666667\\
3	3.843575\\
4	3.52646666666667\\
5	3.1871\\
6	3.18060833333333\\
7	3.13386666666667\\
8	2.88993333333333\\
9	2.46166666666667\\
10	2.24288333333333\\
11	2.080475\\
12	1.82794166666667\\
13	1.56615833333333\\
14	1.44419166666667\\
15	1.32539166666667\\
16	1.27225\\
17	1.10140833333333\\
18	1.07101666666667\\
19	0.976875\\
20	0.925208333333333\\
21	0.7567\\
22	0.701958333333333\\
23	0.666083333333333\\
24	0.640358333333333\\
25	0.602575\\
26	0.584416666666667\\
27	0.561216666666667\\
28	0.537716666666667\\
29	0.516691666666667\\
30	0.497408333333333\\
31	0.479591666666667\\
32	0.466091666666667\\
33	0.447925\\
34	0.434625\\
35	0.424008333333333\\
36	0.413316666666667\\
37	0.401525\\
38	0.393366666666667\\
39	0.384658333333333\\
40	0.3801\\
};
\addlegendentry{prop, uni, $\zeta/K$=0.075}

\addplot [color=mycolor3, dotted, line width=1.0pt]
  table[row sep=crcr]{%
1	5.12172	\\
2	7.44038	\\
3	7.77428	\\
4	7.22408	\\
5	5.98194	\\
6	5.22506	\\
7	4.79814	\\
8	4.38738	\\
9	3.77152	\\
10	3.37394	\\
11	2.90776	\\
12	2.57744	\\
13	2.21898	\\
14	2.06522	\\
15	1.9177	\\
16	1.79586	\\
17	1.53876	\\
18	1.36814	\\
19	1.20104	\\
20	1.09198	\\
21	0.94066	\\
22	0.88218	\\
23	0.83228	\\
24	0.78936	\\
25	0.74074	\\
26	0.69096	\\
27	0.6539	\\
28	0.6209	\\
29	0.59202	\\
30	0.57204	\\
31	0.55286	\\
32	0.52722	\\
33	0.51152	\\
34	0.49612	\\
35	0.48638	\\
36	0.47124	\\
37	0.45632	\\
38	0.44396	\\
39	0.43088	\\
40	0.4217	\\
};
\addlegendentry{rdm, ps, $\zeta/K$=0.075}

\addplot [color=mycolor4, dotted, line width=1.0pt, mark=x, mark repeat=3, mark options={solid, mycolor4}]
  table[row sep=crcr]{%
1	3.13876	\\
2	3.04864	\\
3	2.75672	\\
4	2.41788	\\
5	2.30548	\\
6	2.10248	\\
7	2.00934	\\
8	1.79568	\\
9	1.68792	\\
10	1.46648	\\
11	1.31238	\\
12	1.19188	\\
13	1.03886	\\
14	0.92142	\\
15	0.84726	\\
16	0.77222	\\
17	0.77956	\\
18	0.7065	\\
19	0.67776	\\
20	0.63474	\\
21	0.57484	\\
22	0.54858	\\
23	0.53262	\\
24	0.51952	\\
25	0.49192	\\
26	0.48726	\\
27	0.4807	\\
28	0.4759	\\
29	0.46152	\\
30	0.45498	\\
31	0.4553	\\
32	0.43892	\\
33	0.43686	\\
34	0.43098	\\
35	0.41258	\\
36	0.40108	\\
37	0.37996	\\
38	0.3701	\\
39	0.35504	\\
40	0.34728	\\
};
\addlegendentry{prop, ps, $\zeta/K$=0.075}

\addplot [color=mycolor5, dashdotted, line width=1.0pt]
  table[row sep=crcr]{%
1	4.030575\\
2	4.95959166666667\\
3	5.31588333333333\\
4	4.7044\\
5	3.78535\\
6	3.633725\\
7	3.6929\\
8	3.48146666666667\\
9	3.002475\\
10	2.75493333333333\\
11	2.60614166666667\\
12	2.46451666666667\\
13	2.15439166666667\\
14	1.95310833333333\\
15	1.91090833333333\\
16	1.75255\\
17	1.53060833333333\\
18	1.37713333333333\\
19	1.340675\\
20	1.22005833333333\\
21	1.00679166666667\\
22	0.879466666666667\\
23	0.830616666666667\\
24	0.775991666666667\\
25	0.723875\\
26	0.680058333333333\\
27	0.652283333333333\\
28	0.6176\\
29	0.589558333333333\\
30	0.559025\\
31	0.538491666666667\\
32	0.5146\\
33	0.491191666666667\\
34	0.471558333333333\\
35	0.4569\\
36	0.443333333333333\\
37	0.428533333333333\\
38	0.415216666666667\\
39	0.406358333333333\\
40	0.402341666666667\\
};
\addlegendentry{rdm, uni, $\zeta/K$=0.05}

\addplot [color=mycolor6, dashdotted, line width=1.0pt, mark=o, mark repeat=3, mark options={solid, mycolor6}]
  table[row sep=crcr]{%
1	2.42920833333333\\
2	2.63750833333333\\
3	2.7589\\
4	2.62040833333333\\
5	2.37439166666667\\
6	2.23148333333333\\
7	2.1073\\
8	1.94386666666667\\
9	1.75071666666667\\
10	1.58885833333333\\
11	1.47721666666667\\
12	1.32508333333333\\
13	1.19663333333333\\
14	1.089475\\
15	1.01305833333333\\
16	0.937108333333333\\
17	0.85345\\
18	0.796508333333333\\
19	0.742266666666667\\
20	0.698208333333333\\
21	0.6331\\
22	0.60355\\
23	0.582375\\
24	0.563925\\
25	0.541666666666667\\
26	0.522808333333333\\
27	0.506858333333333\\
28	0.492241666666667\\
29	0.4771\\
30	0.463316666666667\\
31	0.451583333333333\\
32	0.440675\\
33	0.430133333333333\\
34	0.419483333333333\\
35	0.410883333333333\\
36	0.402266666666667\\
37	0.393633333333333\\
38	0.386341666666667\\
39	0.379\\
40	0.373333333333333\\
};
\addlegendentry{prop, uni, $\zeta/K$=0.35}

\end{axis}
\end{tikzpicture}%

%% file: figure/sameRatioDist4TestNiid-sglCol1.tex
%
%
\definecolor{mycolor1}{rgb}{0.00000,0.44700,0.74100}%
\definecolor{mycolor2}{rgb}{0.85000,0.32500,0.09800}%
\definecolor{mycolor3}{rgb}{0.92900,0.69400,0.12500}%
\definecolor{mycolor4}{rgb}{0.49400,0.18400,0.55600}%
\definecolor{mycolor5}{rgb}{0.46600,0.67400,0.18800}%
\definecolor{mycolor6}{rgb}{0.30100,0.74500,0.93300}%
\begin{tikzpicture}

\begin{axis}[%
width=.61\linewidth,
height=.41\linewidth,
scale only axis,
xmin=0,
xmax=40,
xlabel style={font=\color{white!15!black}},
xlabel={Global iteration ($t$)},
ymin=0,
ymax=7,
ylabel style={font=\color{white!15!black}},
ylabel={Testing loss},
axis background/.style={fill=white},
xmajorgrids,
ymajorgrids,
legend style={legend cell align=left, align=left, draw=white!15!black}
]
\addplot [color=mycolor1, line width=1.0pt]
  table[row sep=crcr]{%
1	3.755725\\
2	4.79419166666667\\
3	4.953775\\
4	4.74888333333333\\
5	4.76121666666667\\
6	4.56906666666667\\
7	4.34793333333333\\
8	4.20798333333333\\
9	4.21731666666667\\
10	3.78344166666667\\
11	3.72279166666667\\
12	3.39399166666667\\
13	3.28271666666667\\
14	3.10115\\
15	3.036475\\
16	2.768875\\
17	2.73458333333333\\
18	2.58425\\
19	2.39699166666667\\
20	2.373675\\
21	1.9509\\
22	1.8195\\
23	1.7454\\
24	1.7338\\
25	1.68644166666667\\
26	1.56641666666667\\
27	1.58029166666667\\
28	1.43858333333333\\
29	1.4671\\
30	1.39935833333333\\
31	1.377025\\
32	1.37063333333333\\
33	1.33496666666667\\
34	1.28790833333333\\
35	1.22993333333333\\
36	1.24216666666667\\
37	1.20970833333333\\
38	1.22275833333333\\
39	1.19734166666667\\
40	1.14160833333333\\
};

\addplot [color=mycolor2, line width=1.0pt, mark=x, mark repeat=3, mark options={solid, mycolor2}]
  table[row sep=crcr]{%
1	3.47150833333333\\
2	4.09701666666667\\
3	4.40991666666667\\
4	4.44606666666667\\
5	4.33268333333333\\
6	4.19733333333333\\
7	3.99643333333333\\
8	4.00369166666667\\
9	3.66243333333333\\
10	3.50029166666667\\
11	3.27058333333333\\
12	2.94554166666667\\
13	2.83784166666667\\
14	2.65535\\
15	2.55755\\
16	2.47520833333333\\
17	2.34776666666667\\
18	2.25860833333333\\
19	2.11944166666667\\
20	1.963575\\
21	1.68809166666667\\
22	1.58744166666667\\
23	1.59313333333333\\
24	1.4859\\
25	1.4347\\
26	1.41243333333333\\
27	1.3582\\
28	1.2868\\
29	1.29343333333333\\
30	1.243925\\
31	1.24893333333333\\
32	1.19891666666667\\
33	1.1709\\
34	1.17721666666667\\
35	1.11623333333333\\
36	1.13515833333333\\
37	1.08608333333333\\
38	1.0889\\
39	1.04830833333333\\
40	1.04725\\
};

\addplot [color=mycolor3, dotted, line width=1.0pt]
  table[row sep=crcr]{%
1	3.87172	\\
2	5.01776	\\
3	5.53498	\\
4	5.3087	\\
5	5.14554	\\
6	4.92964	\\
7	4.93046	\\
8	4.66982	\\
9	4.50824	\\
10	4.3881	\\
11	4.15632	\\
12	3.84036	\\
13	3.77478	\\
14	3.68234	\\
15	3.50914	\\
16	3.26586	\\
17	3.14322	\\
18	3.0478	\\
19	2.86218	\\
20	2.72376	\\
21	2.15068	\\
22	1.9842	\\
23	1.83452	\\
24	1.83484	\\
25	1.69476	\\
26	1.63812	\\
27	1.53636	\\
28	1.47112	\\
29	1.41194	\\
30	1.37768	\\
31	1.27862	\\
32	1.2017	\\
33	1.1222	\\
34	1.09758	\\
35	1.07416	\\
36	1.02086	\\
37	0.95672	\\
38	0.96804	\\
39	0.92122	\\
40	0.9079	\\
};

\addplot [color=mycolor4, dotted, line width=1.0pt, mark=x, mark repeat=3, mark options={solid, mycolor4}]
  table[row sep=crcr]{%
1	3.61194	\\
2	4.30906	\\
3	4.10276	\\
4	4.09056	\\
5	3.94884	\\
6	3.85722	\\
7	3.7306	\\
8	3.61168	\\
9	3.47542	\\
10	3.2351	\\
11	3.27834	\\
12	3.10262	\\
13	2.90682	\\
14	2.7586	\\
15	2.63276	\\
16	2.50922	\\
17	2.50612	\\
18	2.38264	\\
19	2.2419	\\
20	2.17624	\\
21	1.87406	\\
22	1.72924	\\
23	1.62984	\\
24	1.68184	\\
25	1.546	\\
26	1.5354	\\
27	1.50892	\\
28	1.43678	\\
29	1.40546	\\
30	1.38564	\\
31	1.37562	\\
32	1.33514	\\
33	1.31014	\\
34	1.23354	\\
35	1.21854	\\
36	1.1255	\\
37	1.02658	\\
38	0.98882	\\
39	0.97756	\\
40	0.93946	\\
};

\addplot [color=mycolor5, dashdotted, line width=1.0pt]
  table[row sep=crcr]{%
1	4.69281666666667\\
2	6.00875\\
3	6.2066\\
4	5.90850833333333\\
5	5.84445\\
6	5.4072\\
7	5.30104166666667\\
8	5.32315833333333\\
9	5.06358333333333\\
10	4.725925\\
11	4.64171666666667\\
12	4.38439166666667\\
13	4.2695\\
14	3.92953333333333\\
15	3.682125\\
16	3.60670833333333\\
17	3.36825\\
18	3.44970833333333\\
19	3.14115\\
20	2.92928333333333\\
21	2.45291666666667\\
22	2.29085833333333\\
23	2.18176666666667\\
24	2.1718\\
25	2.0569\\
26	1.96245\\
27	1.87829166666667\\
28	1.752325\\
29	1.73168333333333\\
30	1.73255\\
31	1.71408333333333\\
32	1.69176666666667\\
33	1.59721666666667\\
34	1.55996666666667\\
35	1.53045833333333\\
36	1.52025833333333\\
37	1.44011666666667\\
38	1.40998333333333\\
39	1.399075\\
40	1.40265833333333\\
};

\addplot [color=mycolor6, dashdotted, line width=1.0pt, mark=o, mark repeat=3, mark options={solid, mycolor6}]
  table[row sep=crcr]{%
1	2.48845833333333\\
2	2.77401666666667\\
3	2.873375\\
4	2.86484166666667\\
5	2.7846\\
6	2.70125833333333\\
7	2.60470833333333\\
8	2.48345833333333\\
9	2.376075\\
10	2.263675\\
11	2.15975833333333\\
12	2.05018333333333\\
13	1.93775833333333\\
14	1.85051666666667\\
15	1.75889166666667\\
16	1.65664166666667\\
17	1.57498333333333\\
18	1.49400833333333\\
19	1.4255\\
20	1.35323333333333\\
21	1.276075\\
22	1.23139166666667\\
23	1.180825\\
24	1.13155833333333\\
25	1.09391666666667\\
26	1.05815\\
27	1.023725\\
28	0.989116666666667\\
29	0.953958333333333\\
30	0.926675\\
31	0.909466666666667\\
32	0.884591666666667\\
33	0.866266666666667\\
34	0.842225\\
35	0.823966666666667\\
36	0.809116666666667\\
37	0.792666666666667\\
38	0.775991666666667\\
39	0.763583333333333\\
40	0.745825\\
};

\end{axis}
\end{tikzpicture}%

%% file: figure/sameRatioDist4EngIid-sglCol1.tex
%
%
\definecolor{mycolor1}{rgb}{0.00000,0.44700,0.74100}%
\definecolor{mycolor2}{rgb}{0.85000,0.32500,0.09800}%
\definecolor{mycolor3}{rgb}{0.92900,0.69400,0.12500}%
\definecolor{mycolor4}{rgb}{0.49400,0.18400,0.55600}%
\definecolor{mycolor5}{rgb}{0.46600,0.67400,0.18800}%
\definecolor{mycolor6}{rgb}{0.30100,0.74500,0.93300}%
\begin{tikzpicture}

\begin{axis}[%
width=.61\linewidth,
height=.42\linewidth,
scale only axis,
xmin=0,
xmax=40,
xlabel style={font=\color{white!15!black}},
xlabel={Global iteration ($t$)},
ymode=log,
ymin=0.245666839058333,
ymax=4.703869610625,
yminorticks=true,
ylabel style={font=\color{white!15!black}},
ylabel={Average $E_k(t)$},
axis background/.style={fill=white},
xmajorgrids,
ymajorgrids,
yminorgrids,
legend style={at={(1.47,1.1)}, anchor=south, legend cell align=left, align=left, draw=white!15!black},
legend columns=4
]
\addplot [color=mycolor1, line width=1.0pt]
  table[row sep=crcr]{%
1	2.31746024416667\\
2	2.5418540323\\
3	2.62157476933333\\
4	2.62320788001667\\
5	2.58700593698333\\
6	2.57662075494167\\
7	2.65403331950833\\
8	2.76538404368333\\
9	2.72965319248333\\
10	2.80957849854167\\
11	2.63009845625\\
12	2.674418398225\\
13	2.73664662118333\\
14	2.60051145185833\\
15	2.63290399509167\\
16	2.7039916307\\
17	2.90965927588333\\
18	2.70535860215833\\
19	2.66023369235833\\
20	2.80031064686667\\
21	2.70461556666667\\
22	2.7968504013\\
23	2.72513530114167\\
24	2.7185562297\\
25	2.73993406830833\\
26	2.66137062829167\\
27	2.90043401893333\\
28	2.89713698176667\\
29	2.75066405081667\\
30	2.73289947793333\\
31	2.79251813259167\\
32	2.81525062193333\\
33	2.66945479471667\\
34	2.71769556859167\\
35	2.7678248537\\
36	2.82354161326667\\
37	2.72912425143333\\
38	2.82628786178333\\
39	2.68970930608333\\
40	2.8594102129\\
};
\addlegendentry{rdm, uni, $\zeta/K=0.075$}
\addplot [color=mycolor2, line width=1.0pt, mark=x, mark repeat=3, mark options={solid, mycolor2}]
  table[row sep=crcr]{%
1	0.255267809741667\\
2	0.253324843183333\\
3	0.298325696933333\\
4	0.283544548941667\\
5	0.2962632256\\
6	0.327683809425\\
7	0.319563913683333\\
8	0.320471632\\
9	0.35023501765\\
10	0.352723745341667\\
11	0.331988912708333\\
12	0.357278097425\\
13	0.293654732441667\\
14	0.321077983366667\\
15	0.295817554033333\\
16	0.245666839058333\\
17	0.3452179787\\
18	0.307484699466667\\
19	0.3165129008\\
20	0.258303609066667\\
21	0.316125661958333\\
22	0.29148880375\\
23	0.312348917308333\\
24	0.285471228125\\
25	0.341737218975\\
26	0.33797833835\\
27	0.342351304675\\
28	0.323217458116667\\
29	0.285158108233333\\
30	0.317283547225\\
31	0.303474132383333\\
32	0.341922540808333\\
33	0.304173877483333\\
34	0.321843646575\\
35	0.33751687615\\
36	0.318704313533333\\
37	0.3056702213\\
38	0.327909117508333\\
39	0.28748381285\\
40	0.307073411725\\
};
\addlegendentry{prop, uni, $\zeta/K=0.075$}
\addplot [color=mycolor3, dotted, line width=1.0pt]
  table[row sep=crcr]{%
1	0.48895892	\\
2	0.593917044	\\
3	0.674722525	\\
4	0.758124179	\\
5	0.81565678	\\
6	0.882980436	\\
7	0.928912923	\\
8	1.100942174	\\
9	1.116229598	\\
10	1.14297701	\\
11	1.177440825	\\
12	1.2746055	\\
13	1.407524506	\\
14	1.494857509	\\
15	1.558216626	\\
16	1.470268342	\\
17	1.57302214	\\
18	1.794796101	\\
19	1.863508497	\\
20	1.845820024	\\
21	1.877812	\\
22	2.002516754	\\
23	2.325362888	\\
24	2.10448727	\\
25	2.244790701	\\
26	2.397402777	\\
27	2.322777944	\\
28	2.339527845	\\
29	2.457778799	\\
30	2.555174776	\\
31	2.602033769	\\
32	2.745263474	\\
33	2.646658798	\\
34	2.734066373	\\
35	2.928015866	\\
36	2.730299	\\
37	2.875472822	\\
38	2.797601469	\\
39	2.675088177	\\
40	2.687229814	\\
};
\addlegendentry{rdm, ps, $\zeta/K=0.075$}

\addplot [color=mycolor4, dotted, line width=1.0pt, mark=x, mark repeat=3, mark options={solid, mycolor4}]
  table[row sep=crcr]{%
1	0.241666433	\\
2	0.355478054	\\
3	0.362394315	\\
4	0.355416701	\\
5	0.34480253	\\
6	0.394878027	\\
7	0.332379846	\\
8	0.348616642	\\
9	0.329624932	\\
10	0.319433874	\\
11	0.372069159	\\
12	0.306378493	\\
13	0.316389508	\\
14	0.349377402	\\
15	0.314911036	\\
16	0.337877901	\\
17	0.342243171	\\
18	0.319964875	\\
19	0.326296459	\\
20	0.305242469	\\
21	0.307784882	\\
22	0.358650141	\\
23	0.319160037	\\
24	0.276928157	\\
25	0.309101462	\\
26	0.328621665	\\
27	0.320677313	\\
28	0.333178466	\\
29	0.339961482	\\
30	0.329728763	\\
31	0.348730666	\\
32	0.345158608	\\
33	0.346488587	\\
34	0.336165709	\\
35	0.366530199	\\
36	0.297824875	\\
37	0.269607034	\\
38	0.270048653	\\
39	0.267069055	\\
40	0.29740575	\\
};
\addlegendentry{prop, ps,$\zeta/K=0.075$}

\addplot [color=mycolor5, dashdotted, line width=1.0pt]
  table[row sep=crcr]{%
1	1.45417339916667\\
2	1.69205811640833\\
3	1.68888796273333\\
4	1.82587198015833\\
5	1.71184056803333\\
6	1.71396445965\\
7	1.70714110666667\\
8	1.7845546371\\
9	1.72622864013333\\
10	1.77371520669167\\
11	1.812131133775\\
12	1.84125725690833\\
13	1.8328513043\\
14	1.85782103494167\\
15	1.77933441825\\
16	1.88131728755\\
17	1.91583327535\\
18	1.81988929829167\\
19	1.84227014384167\\
20	1.9242212224\\
21	1.83628622188333\\
22	1.78852631673333\\
23	1.78363857173333\\
24	1.74261334480833\\
25	1.81884220431667\\
26	1.807839353725\\
27	1.82799131173333\\
28	1.79905193039167\\
29	1.91525493838333\\
30	1.788019635725\\
31	1.76186044843333\\
32	1.761887414825\\
33	1.77675269425\\
34	1.77558071009167\\
35	1.823912592075\\
36	1.72824717035833\\
37	1.82679332085833\\
38	1.84428289474167\\
39	1.80976766001667\\
40	1.74308032248333\\
};
\addlegendentry{rdm, uni, $\zeta/K=0.05$}

\addplot [color=mycolor6, dashdotted, line width=1.0pt, mark=o, mark repeat=3, mark options={solid, mycolor6}]
  table[row sep=crcr]{%
1	1.62269277264167\\
2	3.18999962673333\\
3	1.7559734118\\
4	1.228382054225\\
5	1.062247358325\\
6	1.0430790204\\
7	0.94968213455\\
8	0.987259500166667\\
9	0.953851450616667\\
10	0.982906871991667\\
11	0.987782240341667\\
12	0.963043644691667\\
13	0.976333756058333\\
14	1.01238864163333\\
15	0.999230324841667\\
16	1.00620063870833\\
17	1.000793652125\\
18	1.008228462425\\
19	0.960741473616667\\
20	0.99890628755\\
21	1.015269651925\\
22	0.970491057091667\\
23	1.01559035090833\\
24	0.992246582475\\
25	0.949752114225\\
26	0.941173369433333\\
27	0.972894030925\\
28	0.957355222641667\\
29	0.996137062658333\\
30	1.008773201875\\
31	1.055638750175\\
32	1.01329581091667\\
33	0.9852414335\\
34	1.03537796803333\\
35	1.03133281231667\\
36	1.00978345479167\\
37	0.978737020916667\\
38	1.01641034404167\\
39	1.05326924540833\\
40	0.979686102225\\
};
\addlegendentry{prop, uni, $\zeta/K=0.35$}

\addplot [color=black, dotted, line width=1.0pt, mark=square, mark repeat=10, mark options={solid, black}]
  table[row sep=crcr]{%
1	1.09170447274813\\
2	1.09170447274813\\
3	1.09170447274813\\
4	1.09170447274813\\
5	1.09170447274813\\
6	1.09170447274813\\
7	1.09170447274813\\
8	1.09170447274813\\
9	1.09170447274813\\
10	1.09170447274813\\
11	1.09170447274813\\
12	1.09170447274813\\
13	1.09170447274813\\
14	1.09170447274813\\
15	1.09170447274813\\
16	1.09170447274813\\
17	1.09170447274813\\
18	1.09170447274813\\
19	1.09170447274813\\
20	1.09170447274813\\
21	1.09170447274813\\
22	1.09170447274813\\
23	1.09170447274813\\
24	1.09170447274813\\
25	1.09170447274813\\
26	1.09170447274813\\
27	1.09170447274813\\
28	1.09170447274813\\
29	1.09170447274813\\
30	1.09170447274813\\
31	1.09170447274813\\
32	1.09170447274813\\
33	1.09170447274813\\
34	1.09170447274813\\
35	1.09170447274813\\
36	1.09170447274813\\
37	1.09170447274813\\
38	1.09170447274813\\
39	1.09170447274813\\
40	1.09170447274813\\
};
\addlegendentry{Mean:prop, uni, $\zeta/K=0.35$}

\addplot [color=black, line width=1.0pt, mark=square, mark repeat=10, mark options={solid, black}]
  table[row sep=crcr]{%
1	1.78863978774792\\
2	1.78863978774792\\
3	1.78863978774792\\
4	1.78863978774792\\
5	1.78863978774792\\
6	1.78863978774792\\
7	1.78863978774792\\
8	1.78863978774792\\
9	1.78863978774792\\
10	1.78863978774792\\
11	1.78863978774792\\
12	1.78863978774792\\
13	1.78863978774792\\
14	1.78863978774792\\
15	1.78863978774792\\
16	1.78863978774792\\
17	1.78863978774792\\
18	1.78863978774792\\
19	1.78863978774792\\
20	1.78863978774792\\
21	1.78863978774792\\
22	1.78863978774792\\
23	1.78863978774792\\
24	1.78863978774792\\
25	1.78863978774792\\
26	1.78863978774792\\
27	1.78863978774792\\
28	1.78863978774792\\
29	1.78863978774792\\
30	1.78863978774792\\
31	1.78863978774792\\
32	1.78863978774792\\
33	1.78863978774792\\
34	1.78863978774792\\
35	1.78863978774792\\
36	1.78863978774792\\
37	1.78863978774792\\
38	1.78863978774792\\
39	1.78863978774792\\
40	1.78863978774792\\
};
\addlegendentry{Mean:rdm, uni, $\zeta/K=0.05$}

\end{axis}
\end{tikzpicture}%

%% file: figure/sameRatioDist4EngNiid-sglCol1.tex
%
%
\definecolor{mycolor1}{rgb}{0.00000,0.44700,0.74100}%
\definecolor{mycolor2}{rgb}{0.85000,0.32500,0.09800}%
\definecolor{mycolor3}{rgb}{0.92900,0.69400,0.12500}%
\definecolor{mycolor4}{rgb}{0.49400,0.18400,0.55600}%
\definecolor{mycolor5}{rgb}{0.46600,0.67400,0.18800}%
\definecolor{mycolor6}{rgb}{0.30100,0.74500,0.93300}%
\begin{tikzpicture}

\begin{axis}[%
width=.61\linewidth,
height=.42\linewidth,
scale only axis,
xmin=0,
xmax=40,
xlabel style={font=\color{white!15!black}},
xlabel={Global iteration ($t$)},
ymode=log,
ymin=0.236793151683333,
ymax=4.69916302355833,
yminorticks=true,
ylabel style={font=\color{white!15!black}},
ylabel={Average $E_k(t)$},
axis background/.style={fill=white},
xmajorgrids,
ymajorgrids,
yminorgrids,
legend style={at={(0.97,0.03)}, anchor=south east, legend cell align=left, align=left, draw=white!15!black}
]
\addplot [color=mycolor1, line width=1.0pt, forget plot]
  table[row sep=crcr]{%
1	2.22065943516667\\
2	2.4958438277\\
3	2.54893072275833\\
4	2.64640903693333\\
5	2.59813116893333\\
6	2.77405731419167\\
7	2.61006061931667\\
8	2.5472827209\\
9	2.69763318790833\\
10	2.88269756425\\
11	2.66463514734167\\
12	2.716404023975\\
13	2.66792503951667\\
14	2.62435476366667\\
15	2.671033570975\\
16	2.86281946956667\\
17	2.69794465436667\\
18	2.723208123525\\
19	2.7272010195\\
20	2.79382973755\\
21	2.68182171071667\\
22	2.60593254555\\
23	2.757996482775\\
24	2.79446152719167\\
25	2.68833149159167\\
26	2.73197541039167\\
27	2.71923721381667\\
28	2.8223604828\\
29	2.6956480951\\
30	2.82056457808333\\
31	2.84578537285\\
32	2.81836689890833\\
33	2.685219281025\\
34	2.635142858\\
35	2.81300122961667\\
36	2.68361964461667\\
37	2.79316780834167\\
38	2.739427071475\\
39	2.76041667265833\\
40	2.74829911504167\\
};
\addplot [color=mycolor2, line width=1.0pt, mark=x, mark repeat=3, mark options={solid, mycolor2}, forget plot]
  table[row sep=crcr]{%
1	0.236793151683333\\
2	0.278335011425\\
3	0.277414839116667\\
4	0.314731051516667\\
5	0.330496764066667\\
6	0.318556057866667\\
7	0.378099826025\\
8	0.328874821766667\\
9	0.341531466658333\\
10	0.360614450633333\\
11	0.293992734033333\\
12	0.318800520208333\\
13	0.257172966608333\\
14	0.275568501358333\\
15	0.339845163833333\\
16	0.261287101825\\
17	0.325167937216667\\
18	0.273336719141667\\
19	0.287476224233333\\
20	0.295746983641667\\
21	0.310257771208333\\
22	0.29907348785\\
23	0.345693199641667\\
24	0.312062626558333\\
25	0.296468147475\\
26	0.310918159741667\\
27	0.334100513175\\
28	0.294345723225\\
29	0.309759524858333\\
30	0.297452597083333\\
31	0.2959107269\\
32	0.346193934525\\
33	0.303924906233333\\
34	0.324304144516667\\
35	0.349353981483333\\
36	0.332573169891667\\
37	0.31970395565\\
38	0.326636040916667\\
39	0.304045169525\\
40	0.282160996825\\
};
\addplot [color=mycolor3, dotted, line width=1.0pt, forget plot]
  table[row sep=crcr]{%
1	0.598975352	\\
2	0.581580295	\\
3	0.686154656	\\
4	0.718781157	\\
5	0.884480332	\\
6	0.933263407	\\
7	0.987898595	\\
8	1.066239166	\\
9	1.072091057	\\
10	1.204929475	\\
11	1.215083117	\\
12	1.244109023	\\
13	1.413626724	\\
14	1.492171611	\\
15	1.580879828	\\
16	1.592453819	\\
17	1.697943337	\\
18	1.768632499	\\
19	1.833166562	\\
20	1.876807497	\\
21	1.919917096	\\
22	2.095580102	\\
23	2.042304629	\\
24	2.079887218	\\
25	2.19370754	\\
26	2.288370139	\\
27	2.248126428	\\
28	2.386403853	\\
29	2.408899867	\\
30	2.631889082	\\
31	2.612951906	\\
32	2.673353362	\\
33	2.721871816	\\
34	2.609345638	\\
35	2.788534993	\\
36	2.62894476	\\
37	2.677933342	\\
38	2.742283798	\\
39	2.766446868	\\
40	2.765633318	\\
};
\addplot [color=mycolor4, dotted, line width=1.0pt, mark=x, mark repeat=3, mark options={solid, mycolor4}, forget plot]
  table[row sep=crcr]{%
1	0.254414138	\\
2	0.364865559	\\
3	0.387747133	\\
4	0.402344645	\\
5	0.344026903	\\
6	0.344469153	\\
7	0.356076573	\\
8	0.386795027	\\
9	0.344551665	\\
10	0.339475721	\\
11	0.343617727	\\
12	0.30713087	\\
13	0.319041884	\\
14	0.297518331	\\
15	0.342245573	\\
16	0.312022205	\\
17	0.335567948	\\
18	0.31856418	\\
19	0.30275504	\\
20	0.331937926	\\
21	0.317053571	\\
22	0.318742608	\\
23	0.308984465	\\
24	0.30029828	\\
25	0.345344908	\\
26	0.365969417	\\
27	0.339161708	\\
28	0.304068746	\\
29	0.324950579	\\
30	0.376702964	\\
31	0.346739836	\\
32	0.336223388	\\
33	0.355447118	\\
34	0.349180077	\\
35	0.348562109	\\
36	0.304452208	\\
37	0.281326296	\\
38	0.273712337	\\
39	0.290570362	\\
40	0.280612342	\\
};
\addplot [color=mycolor5, dashdotted, line width=1.0pt, forget plot]
  table[row sep=crcr]{%
1	1.616515772775\\
2	1.7464211688\\
3	1.68472602614167\\
4	1.72157034723333\\
5	1.728427566925\\
6	1.78761759800833\\
7	1.77316201280833\\
8	1.76325946596667\\
9	1.84960998255833\\
10	1.7807208155\\
11	1.84895521399167\\
12	1.8579608516\\
13	1.83790601025833\\
14	1.79743641938333\\
15	1.67222432004167\\
16	1.841419177175\\
17	1.780501725625\\
18	1.871369004425\\
19	1.82918832754167\\
20	1.83203625221667\\
21	1.73146806495\\
22	1.82512722840833\\
23	1.821852736625\\
24	1.81457195114167\\
25	1.89357271335833\\
26	1.89098016009167\\
27	1.83016679655\\
28	1.87259096614167\\
29	1.80606991626667\\
30	1.80396058041667\\
31	1.845433640675\\
32	1.81688029575\\
33	1.859446318025\\
34	1.76479886554167\\
35	1.75125457425833\\
36	1.788164115\\
37	1.76546593975\\
38	1.90949040253333\\
39	1.81946531444167\\
40	1.800817138775\\
};
\addplot [color=mycolor6, dashdotted, line width=1.0pt, mark=o, mark repeat=3, mark options={solid, mycolor6}, forget plot]
  table[row sep=crcr]{%
1	1.61380445590833\\
2	3.20368695991667\\
3	1.70375873090833\\
4	1.23778578085\\
5	1.1105011535\\
6	0.997303967483333\\
7	0.966863175466667\\
8	1.019083985075\\
9	0.98764776165\\
10	1.001361960475\\
11	0.988906517316667\\
12	0.9721289129\\
13	0.982846805383333\\
14	0.993572775625\\
15	1.01888208419167\\
16	1.03363466755\\
17	0.994715106333333\\
18	1.00426820064167\\
19	0.958730846841667\\
20	0.977336481566667\\
21	1.00599923693333\\
22	1.00487183285833\\
23	1.02546404344167\\
24	0.983506880775\\
25	0.9560734425\\
26	0.955412064191667\\
27	0.97812544115\\
28	0.972970739308333\\
29	0.986865389841667\\
30	0.992903561741667\\
31	1.06077559083333\\
32	1.012998535075\\
33	0.993181390633333\\
34	1.00957924720833\\
35	1.04273264165833\\
36	1.002270091825\\
37	0.9829969932\\
38	1.01324952446667\\
39	1.05927282944167\\
40	0.986578838791667\\
};
\addplot [color=black, dotted, line width=1.0pt, mark=square, mark repeat=10, mark options={solid, black}]
  table[row sep=crcr]{%
1	1.09481621613646\\
2	1.09481621613646\\
3	1.09481621613646\\
4	1.09481621613646\\
5	1.09481621613646\\
6	1.09481621613646\\
7	1.09481621613646\\
8	1.09481621613646\\
9	1.09481621613646\\
10	1.09481621613646\\
11	1.09481621613646\\
12	1.09481621613646\\
13	1.09481621613646\\
14	1.09481621613646\\
15	1.09481621613646\\
16	1.09481621613646\\
17	1.09481621613646\\
18	1.09481621613646\\
19	1.09481621613646\\
20	1.09481621613646\\
21	1.09481621613646\\
22	1.09481621613646\\
23	1.09481621613646\\
24	1.09481621613646\\
25	1.09481621613646\\
26	1.09481621613646\\
27	1.09481621613646\\
28	1.09481621613646\\
29	1.09481621613646\\
30	1.09481621613646\\
31	1.09481621613646\\
32	1.09481621613646\\
33	1.09481621613646\\
34	1.09481621613646\\
35	1.09481621613646\\
36	1.09481621613646\\
37	1.09481621613646\\
38	1.09481621613646\\
39	1.09481621613646\\
40	1.09481621613646\\
};

\addplot [color=black, line width=1.0pt, mark=square, mark repeat=10, mark options={solid, black}]
  table[row sep=crcr]{%
1	1.80081514444188\\
2	1.80081514444188\\
3	1.80081514444188\\
4	1.80081514444188\\
5	1.80081514444188\\
6	1.80081514444188\\
7	1.80081514444188\\
8	1.80081514444188\\
9	1.80081514444188\\
10	1.80081514444188\\
11	1.80081514444188\\
12	1.80081514444188\\
13	1.80081514444188\\
14	1.80081514444188\\
15	1.80081514444188\\
16	1.80081514444188\\
17	1.80081514444188\\
18	1.80081514444188\\
19	1.80081514444188\\
20	1.80081514444188\\
21	1.80081514444188\\
22	1.80081514444188\\
23	1.80081514444188\\
24	1.80081514444188\\
25	1.80081514444188\\
26	1.80081514444188\\
27	1.80081514444188\\
28	1.80081514444188\\
29	1.80081514444188\\
30	1.80081514444188\\
31	1.80081514444188\\
32	1.80081514444188\\
33	1.80081514444188\\
34	1.80081514444188\\
35	1.80081514444188\\
36	1.80081514444188\\
37	1.80081514444188\\
38	1.80081514444188\\
39	1.80081514444188\\
40	1.80081514444188\\
};

\end{axis}
\end{tikzpicture}%

%% file: figure/diffMetricAccu-sglCol.tex
%
%
\definecolor{mycolor1}{rgb}{0.00000,0.44700,0.74100}%
\definecolor{mycolor2}{rgb}{0.85000,0.32500,0.09800}%
\definecolor{mycolor3}{rgb}{0.92900,0.69400,0.12500}%
\definecolor{mycolor4}{rgb}{0.49400,0.18400,0.55600}%
\definecolor{mycolor5}{rgb}{0.46600,0.67400,0.18800}%
\definecolor{mycolor6}{rgb}{0.30100,0.74500,0.93300}%
\begin{tikzpicture}

\begin{axis}[%
width=.61\linewidth,
height=.4\linewidth,
scale only axis,
xmin=0,
xmax=40,
xlabel style={font=\color{white!15!black}},
xlabel={Global iteration ($t$)},
ymin=10,
ymax=60,
ylabel style={font=\color{white!15!black}},
ylabel={Test accuracy ($\%$)},
axis background/.style={fill=white},
xmajorgrids,
ymajorgrids,
legend style={at={(1.4,1.1)}, anchor=south, legend cell align=left, align=left, draw=white!15!black},
legend columns=4
]
\addplot [color=mycolor1, line width=1.0pt, mark=o, mark repeat=3, mark options={solid, mycolor1}]
  table[row sep=crcr]{%
1	10.985\\
2	11.085\\
3	11.595\\
4	12.295\\
5	13.19\\
6	14.135\\
7	15.38\\
8	15.99\\
9	17.335\\
10	18.225\\
11	19.19\\
12	19.92\\
13	20.955\\
14	21.835\\
15	23.775\\
16	24.045\\
17	24.49\\
18	25.56\\
19	27.83\\
20	28.965\\
21	33.225\\
22	35.49\\
23	35.115\\
24	36.1\\
25	37.175\\
26	38.28\\
27	38.61\\
28	39.28\\
29	39.605\\
30	40.195\\
31	43.225\\
32	41.36\\
33	44.05\\
34	44.825\\
35	47.865\\
36	48.235\\
37	48.765\\
38	49.12\\
39	51.32\\
40	50.215\\
};
\addlegendentry{prop:  $\zeta/K=0.05$}

\addplot [color=mycolor4, dotted, line width=1.0pt, mark=o, mark repeat=3, mark options={solid, mycolor4}]
table[row sep=crcr]{%
	1	11.12\\
	2	11.64\\
	3	12.39\\
	4	13.515\\
	5	14.965\\
	6	16.93\\
	7	17.835\\
	8	19.625\\
	9	21.09\\
	10	23.865\\
	11	24.335\\
	12	26.585\\
	13	28.425\\
	14	29.505\\
	15	30.555\\
	16	33.49\\
	17	33.72\\
	18	36.35\\
	19	36.21\\
	20	38.905\\
	21	42.975\\
	22	42.91\\
	23	44.28\\
	24	45.89\\
	25	44.64\\
	26	45.705\\
	27	45.97\\
	28	48.09\\
	29	47.81\\
	30	50.07\\
	31	51.745\\
	32	52.835\\
	33	54.68\\
	34	56.15\\
	35	56.05\\
	36	57.26\\
	37	58.4\\
	38	59\\
	39	59.525\\
	40	60\\
};
\addlegendentry{prop:  $\zeta/K=0.1$}

\addplot [color=mycolor2, line width=1.0pt]
  table[row sep=crcr]{%
1	10.985\\
2	11.7\\
3	13.11\\
4	14.82\\
5	16.965\\
6	18.95\\
7	21.07\\
8	22.615\\
9	24.305\\
10	25.25\\
11	26.96\\
12	27.83\\
13	27.58\\
14	28.54\\
15	29.07\\
16	29.92\\
17	30.285\\
18	29.975\\
19	30.865\\
20	31.655\\
21	31.85\\
22	32.73\\
23	33.045\\
24	33.05\\
25	32.895\\
26	32.585\\
27	33.65\\
28	33.895\\
29	33.94\\
30	33.455\\
31	34.35\\
32	34.655\\
33	34.45\\
34	34.76\\
35	35.965\\
36	36.24\\
37	36.84\\
38	36.91\\
39	37.61\\
40	39.05\\
};
\addlegendentry{\cite{xu2021client}:  $\zeta/K=0.05$}

\addplot [color=mycolor5, dotted, line width=1.0pt]
table[row sep=crcr]{%
	1	11.02\\
	2	11.825\\
	3	12.845\\
	4	14.46\\
	5	16.45\\
	6	18.655\\
	7	20.685\\
	8	23.175\\
	9	25.835\\
	10	28.305\\
	11	29.955\\
	12	32.185\\
	13	33.605\\
	14	35.66\\
	15	36.855\\
	16	38.16\\
	17	39.805\\
	18	40.255\\
	19	41.885\\
	20	42.09\\
	21	43.43\\
	22	44.08\\
	23	44.2\\
	24	44.61\\
	25	44.87\\
	26	45.105\\
	27	45.245\\
	28	45.725\\
	29	46.125\\
	30	46.515\\
	31	46.8\\
	32	46.925\\
	33	46.745\\
	34	46.96\\
	35	47.15\\
	36	47.28\\
	37	48.015\\
	38	47.93\\
	39	47.915\\
	40	51.485\\
};
\addlegendentry{\cite{xu2021client}:  $\zeta/K=0.1$}

\addplot [color=mycolor3, line width=1.0pt, mark=x, mark repeat=3, mark options={solid, mycolor3}]
table[row sep=crcr]{%
	1	10.83\\
	2	11.47\\
	3	12.42\\
	4	13.62\\
	5	15.69\\
	6	17.705\\
	7	19.58\\
	8	20.715\\
	9	22.325\\
	10	23.195\\
	11	25.145\\
	12	25.85\\
	13	26.42\\
	14	28.165\\
	15	27.83\\
	16	28.965\\
	17	30.26\\
	18	30.43\\
	19	31.425\\
	20	31.74\\
	21	32.89\\
	22	34.245\\
	23	34.99\\
	24	35.575\\
	25	35.87\\
	26	35.615\\
	27	35.72\\
	28	36.805\\
	29	37.225\\
	30	37.18\\
	31	37.555\\
	32	38.515\\
	33	38.085\\
	34	38.265\\
	35	39.465\\
	36	38.175\\
	37	39.495\\
	38	39.92\\
	39	39.42\\
	40	41.23\\
};
\addlegendentry{\cite{ji2022client}:  $\zeta/K=0.05$}

\addplot [color=mycolor6, dotted, line width=1.0pt, mark=x, mark repeat=3, mark options={solid, mycolor6}]
  table[row sep=crcr]{%
1	11.115\\
2	11.865\\
3	12.76\\
4	14.19\\
5	16.24\\
6	18.2\\
7	20.135\\
8	22.25\\
9	23.945\\
10	26.46\\
11	28.13\\
12	29.925\\
13	31.74\\
14	33.36\\
15	35.355\\
16	35.845\\
17	37.4\\
18	38.58\\
19	39.585\\
20	40.08\\
21	42.995\\
22	42.965\\
23	44.12\\
24	44.78\\
25	44.835\\
26	46.255\\
27	45.935\\
28	47.055\\
29	47.225\\
30	47.89\\
31	48.385\\
32	48.98\\
33	49.8\\
34	49.635\\
35	49.66\\
36	49.97\\
37	50.82\\
38	51.095\\
39	51.845\\
40	52.965\\
};
\addlegendentry{\cite{ji2022client}:  $\zeta/K=0.1$}

\end{axis}
\end{tikzpicture}%

%% file: figure/diffMetricLoss-sglCol.tex
%
%
\definecolor{mycolor1}{rgb}{0.00000,0.44700,0.74100}%
\definecolor{mycolor2}{rgb}{0.85000,0.32500,0.09800}%
\definecolor{mycolor3}{rgb}{0.92900,0.69400,0.12500}%
\definecolor{mycolor4}{rgb}{0.49400,0.18400,0.55600}%
\definecolor{mycolor5}{rgb}{0.46600,0.67400,0.18800}%
\definecolor{mycolor6}{rgb}{0.30100,0.74500,0.93300}%
\begin{tikzpicture}

\begin{axis}[%
width=.61\linewidth,
height=.4\linewidth,
scale only axis,
xmin=0,
xmax=40,
xlabel style={font=\color{white!15!black}},
xlabel={Global iteration ($t$)},
ymin=1,
ymax=7,
ylabel style={font=\color{white!15!black}},
ylabel={Testing loss},
axis background/.style={fill=white},
xmajorgrids,
ymajorgrids,
legend style={legend cell align=left, align=left, draw=white!15!black}
]
\addplot [color=mycolor1, line width=1.0pt, mark=o, mark repeat=3, mark options={solid, mycolor1}]
  table[row sep=crcr]{%
1	4.29775\\
2	5.16415\\
3	5.4833\\
4	5.173\\
5	5.1592\\
6	4.73875\\
7	4.7019\\
8	4.3741\\
9	4.4171\\
10	4.23825\\
11	4.0511\\
12	3.96315\\
13	3.9743\\
14	3.8096\\
15	3.5833\\
16	3.60025\\
17	3.51405\\
18	3.3348\\
19	3.13185\\
20	3.0958\\
21	2.5117\\
22	2.3743\\
23	2.4836\\
24	2.34585\\
25	2.35315\\
26	2.23405\\
27	2.21625\\
28	2.15775\\
29	2.17265\\
30	2.16235\\
31	1.9646\\
32	2.11725\\
33	1.9321\\
34	1.92115\\
35	1.78335\\
36	1.7519\\
37	1.73885\\
38	1.73735\\
39	1.6357\\
40	1.7014\\
};

\addplot [color=mycolor2, line width=1.0pt]
  table[row sep=crcr]{%
1	4.38355\\
2	5.95175\\
3	6.0913\\
4	5.89935\\
5	5.62655\\
6	5.42715\\
7	5.48285\\
8	5.61505\\
9	5.64075\\
10	5.6999\\
11	5.7306\\
12	5.856\\
13	5.96375\\
14	5.9713\\
15	5.965\\
16	5.97375\\
17	6.02735\\
18	6.08045\\
19	5.88825\\
20	5.7497\\
21	5.59385\\
22	5.36835\\
23	5.27805\\
24	5.05625\\
25	4.93125\\
26	4.9097\\
27	4.60805\\
28	4.56675\\
29	4.39505\\
30	4.25275\\
31	4.0066\\
32	3.9038\\
33	3.7645\\
34	3.58585\\
35	3.4128\\
36	3.31215\\
37	3.23315\\
38	3.10125\\
39	2.9266\\
40	2.62925\\
};

\addplot [color=mycolor3, line width=1.0pt, mark=x, mark repeat=3, mark options={solid, mycolor3}]
  table[row sep=crcr]{%
1	4.38695\\
2	5.8395\\
3	5.95825\\
4	6.00315\\
5	5.652\\
6	5.35025\\
7	5.2174\\
8	5.22515\\
9	5.1723\\
10	5.0724\\
11	4.9367\\
12	4.9996\\
13	5.0353\\
14	4.97455\\
15	4.96335\\
16	4.89955\\
17	4.8104\\
18	4.7584\\
19	4.66675\\
20	4.67645\\
21	4.38095\\
22	4.2848\\
23	4.18725\\
24	4.0976\\
25	4.0366\\
26	4.0773\\
27	4.0703\\
28	3.92085\\
29	3.85875\\
30	3.85385\\
31	3.74765\\
32	3.6147\\
33	3.5774\\
34	3.4517\\
35	3.2573\\
36	3.28235\\
37	3.14625\\
38	3.0259\\
39	2.92975\\
40	2.5854\\
};

\addplot [color=mycolor4, dotted, line width=1.0pt, mark=o, mark repeat=3, mark options={solid, mycolor4}]
  table[row sep=crcr]{%
1	3.16985\\
2	4.2809\\
3	4.4639\\
4	4.3454\\
5	4.10255\\
6	3.91265\\
7	3.8253\\
8	3.66495\\
9	3.5569\\
10	3.2718\\
11	3.18845\\
12	3.15445\\
13	2.9606\\
14	2.8939\\
15	2.82\\
16	2.60845\\
17	2.6131\\
18	2.40515\\
19	2.46555\\
20	2.2761\\
21	2.03865\\
22	2.00765\\
23	1.91425\\
24	1.8579\\
25	1.91165\\
26	1.83165\\
27	1.808\\
28	1.6982\\
29	1.71695\\
30	1.6152\\
31	1.52675\\
32	1.4849\\
33	1.4335\\
34	1.3923\\
35	1.38105\\
36	1.3351\\
37	1.29275\\
38	1.2766\\
39	1.27635\\
40	1.26\\
};

\addplot [color=mycolor5, dotted, line width=1.0pt]
  table[row sep=crcr]{%
1	3.1442\\
2	4.25\\
3	4.5724\\
4	4.3976\\
5	4.198\\
6	3.9966\\
7	3.86645\\
8	3.78465\\
9	3.72095\\
10	3.67245\\
11	3.6295\\
12	3.59905\\
13	3.56245\\
14	3.5812\\
15	3.53625\\
16	3.5476\\
17	3.5167\\
18	3.5288\\
19	3.47945\\
20	3.495\\
21	3.46825\\
22	3.48385\\
23	3.4614\\
24	3.40545\\
25	3.3554\\
26	3.32985\\
27	3.2516\\
28	3.15205\\
29	3.11235\\
30	3.0383\\
31	2.96005\\
32	2.9039\\
33	2.8223\\
34	2.78085\\
35	2.7145\\
36	2.63365\\
37	2.5687\\
38	2.43495\\
39	2.27015\\
40	1.75\\
};

\addplot [color=mycolor6, dotted, line width=1.0pt, mark=x, mark repeat=3, mark options={solid, mycolor6}]
  table[row sep=crcr]{%
1	3.18605\\
2	4.33695\\
3	4.5605\\
4	4.4376\\
5	4.3107\\
6	4.1657\\
7	3.9888\\
8	3.8545\\
9	3.8001\\
10	3.71665\\
11	3.585\\
12	3.52735\\
13	3.37075\\
14	3.37035\\
15	3.2522\\
16	3.22585\\
17	3.15015\\
18	3.05755\\
19	2.98535\\
20	2.94285\\
21	2.78735\\
22	2.8305\\
23	2.73575\\
24	2.6617\\
25	2.6489\\
26	2.55835\\
27	2.5622\\
28	2.4508\\
29	2.4403\\
30	2.383\\
31	2.3206\\
32	2.29125\\
33	2.25175\\
34	2.23705\\
35	2.20315\\
36	2.17065\\
37	2.1344\\
38	2.1112\\
39	2.01035\\
40	1.73775\\
};

\end{axis}
\end{tikzpicture}%

%% file: figure/diffGammaAccuIid-sglCol.tex
%
%
\definecolor{mycolor1}{rgb}{0.00000,0.44700,0.74100}%
\definecolor{mycolor2}{rgb}{0.85000,0.32500,0.09800}%
\definecolor{mycolor3}{rgb}{0.92900,0.69400,0.12500}%
\definecolor{mycolor4}{rgb}{0.49400,0.18400,0.55600}%
\definecolor{mycolor5}{rgb}{0.46600,0.67400,0.18800}%
\definecolor{mycolor6}{rgb}{0.30100,0.74500,0.93300}%
\begin{tikzpicture}

\begin{axis}[%
width=.64\linewidth,
height=.42\linewidth,
scale only axis,
xmin=0,
xmax=40,
xlabel style={font=\color{white!15!black}},
xlabel={Global iteration ($t$)},
ymin=0,
ymax=100,
ylabel style={font=\color{white!15!black}},
ylabel={Test accuracy ($\%$)},
axis background/.style={fill=white},
xmajorgrids,
ymajorgrids,
legend style={at={(1.35,1.1)}, anchor=south, legend cell align=left, align=left, draw=white!15!black},
legend columns=6
]
\addplot [color=mycolor1, line width=1.0pt, mark=o, mark repeat=5, mark options={solid, mycolor1}]
  table[row sep=crcr]{%
1	10.6391666666667\\
2	11.4525\\
3	12.1383333333333\\
4	12.8708333333333\\
5	14.165\\
6	14.6908333333333\\
7	16.75\\
8	18.3\\
9	20.9391666666667\\
10	22.9325\\
11	24.8383333333333\\
12	27.0033333333333\\
13	29.315\\
14	32.4891666666667\\
15	35.8375\\
16	38.9758333333333\\
17	39.9891666666667\\
18	44.9266666666667\\
19	48.0558333333333\\
20	49.595\\
21	55.3375\\
22	58.9775\\
23	61.4741666666667\\
24	63.9075\\
25	65.9008333333333\\
26	67.1758333333333\\
27	69.2016666666667\\
28	71.9258333333333\\
29	73.29\\
30	75.2091666666667\\
31	77.4058333333333\\
32	79.9191666666667\\
33	80.1575\\
34	82.335\\
35	83.1925\\
36	84.0966666666667\\
37	84.7508333333333\\
38	85.6425\\
39	86.3816666666667\\
40	86.5441666666667\\
};
\addlegendentry{rdm}

\addplot [color=mycolor2, line width=1.0pt, mark=x, mark repeat=3, mark options={solid, mycolor2}]
  table[row sep=crcr]{%
1	10.29\\
2	12.69\\
3	15.0033333333333\\
4	17.64\\
5	21.2941666666667\\
6	23.8233333333333\\
7	27.7291666666667\\
8	32.0933333333333\\
9	36.8758333333333\\
10	41.11\\
11	44.2075\\
12	48.9741666666667\\
13	52.29\\
14	55.7491666666667\\
15	59.1891666666667\\
16	61.66\\
17	65.2616666666667\\
18	65.8833333333333\\
19	68.1333333333333\\
20	69.9341666666667\\
21	73.3383333333333\\
22	74.9958333333333\\
23	75.4516666666667\\
24	76.2633333333333\\
25	77.4675\\
26	77.6166666666667\\
27	78.4158333333333\\
28	79.4033333333333\\
29	79.6533333333333\\
30	80.0491666666667\\
31	80.7866666666667\\
32	81.0858333333333\\
33	81.5091666666667\\
34	82.5891666666667\\
35	82.6875\\
36	83.8483333333333\\
37	85.32\\
38	86.6325\\
39	87.7625\\
40	88.3108333333333\\
};
\addlegendentry{prop, $\gamma=1$}

\addplot [color=mycolor3, dashdotted, line width=1.0pt]
  table[row sep=crcr]{%
1	10.775\\
2	13.0208333333333\\
3	15.7983333333333\\
4	19.1216666666667\\
5	22.5841666666667\\
6	27.0125\\
7	31.7891666666667\\
8	36.4916666666667\\
9	39.3091666666667\\
10	45.3791666666667\\
11	47.7558333333333\\
12	51.9933333333333\\
13	57.1883333333333\\
14	57.8425\\
15	61.3791666666667\\
16	62.665\\
17	66.0858333333333\\
18	65.44\\
19	68.6558333333333\\
20	69.6766666666667\\
21	74.68\\
22	75.6925\\
23	77.015\\
24	77.0433333333333\\
25	77.0416666666667\\
26	78.6\\
27	78.9008333333333\\
28	80.0183333333333\\
29	79.855\\
30	80.7375\\
31	81.4625\\
32	81.2933333333333\\
33	82.7525\\
34	84.2508333333333\\
35	85.0033333333333\\
36	86.51\\
37	87.745\\
38	88.67\\
39	89.22\\
40	89.5033333333333\\
};
\addlegendentry{prop, $\gamma=0.65$}

\addplot [color=mycolor4, dashdotted, line width=1.0pt, mark=triangle, mark repeat=3, mark options={solid, rotate=270, mycolor4}]
  table[row sep=crcr]{%
1	10.8625\\
2	13.3391666666667\\
3	16.25\\
4	19.6866666666667\\
5	22.9708333333333\\
6	27.1716666666667\\
7	30.0775\\
8	34.6925\\
9	37.5008333333333\\
10	42.4958333333333\\
11	45.1683333333333\\
12	46.4991666666667\\
13	47.8675\\
14	51.5741666666667\\
15	50.9391666666667\\
16	55.1808333333333\\
17	56.8258333333333\\
18	57.9975\\
19	60.4308333333333\\
20	61.7091666666667\\
21	69.0083333333333\\
22	70.1525\\
23	72.705\\
24	72.1225\\
25	75.64\\
26	76.7283333333333\\
27	78.5091666666667\\
28	80.3175\\
29	80.9908333333333\\
30	81.9716666666667\\
31	82.53\\
32	84.1791666666667\\
33	85.0116666666667\\
34	85.6575\\
35	86.4725\\
36	87.3383333333333\\
37	87.9483333333333\\
38	88.7416666666667\\
39	89.1375\\
40	89.3025\\
};
\addlegendentry{prop, $\gamma=0.31$}

\addplot [color=mycolor5, line width=1.0pt]
  table[row sep=crcr]{%
1	10.8316666666667\\
2	12.9758333333333\\
3	14.9408333333333\\
4	19.2058333333333\\
5	20.4625\\
6	23.5416666666667\\
7	25.5108333333333\\
8	28.325\\
9	30.5875\\
10	32.0175\\
11	33.9383333333333\\
12	36.525\\
13	37.9458333333333\\
14	39.8591666666667\\
15	42.5908333333333\\
16	45.0808333333333\\
17	45.9891666666667\\
18	50.0933333333333\\
19	51.7883333333333\\
20	57.1333333333333\\
21	63.7491666666667\\
22	65.4075\\
23	70.045\\
24	71.0766666666667\\
25	73.9175\\
26	74.8191666666667\\
27	77.2033333333333\\
28	79.0383333333333\\
29	80.3241666666667\\
30	81.4533333333333\\
31	82.205\\
32	83.0983333333333\\
33	84.4475\\
34	85.1675\\
35	85.8991666666667\\
36	86.6475\\
37	87.3666666666667\\
38	87.9433333333333\\
39	88.3933333333333\\
40	88.645\\
};
\addlegendentry{prop, $\gamma=0.18$}

\addplot [color=mycolor6, dotted, line width=1.0pt, mark=square, mark repeat=3, mark options={solid, mycolor6}]
  table[row sep=crcr]{%
1	10.7758333333333\\
2	12.3108333333333\\
3	14.2658333333333\\
4	15.9058333333333\\
5	18.4883333333333\\
6	19.9691666666667\\
7	22.1808333333333\\
8	23.5375\\
9	26.7591666666667\\
10	28.0558333333333\\
11	31.0475\\
12	32.6208333333333\\
13	36.6933333333333\\
14	37.1175\\
15	40.1283333333333\\
16	45.095\\
17	46.8916666666667\\
18	50.245\\
19	53.2666666666667\\
20	55.8108333333333\\
21	60.34\\
22	62.4316666666667\\
23	64.6433333333333\\
24	66.23\\
25	68.42\\
26	70.8466666666667\\
27	72.6558333333333\\
28	74.9033333333333\\
29	76.8883333333333\\
30	77.0266666666667\\
31	79.1783333333333\\
32	80.1\\
33	81.6641666666667\\
34	82.8816666666667\\
35	83.7216666666667\\
36	84.8441666666667\\
37	85.6483333333333\\
38	86.5433333333333\\
39	87.1425\\
40	87.5166666666667\\
};
\addlegendentry{prop, $\gamma=0.1$}

\end{axis}
\end{tikzpicture}%

%% file: figure/diffGammaAccuNiid-sglCol.tex
%
%
\definecolor{mycolor1}{rgb}{0.00000,0.44700,0.74100}%
\definecolor{mycolor2}{rgb}{0.85000,0.32500,0.09800}%
\definecolor{mycolor3}{rgb}{0.92900,0.69400,0.12500}%
\definecolor{mycolor4}{rgb}{0.49400,0.18400,0.55600}%
\definecolor{mycolor5}{rgb}{0.46600,0.67400,0.18800}%
\definecolor{mycolor6}{rgb}{0.30100,0.74500,0.93300}%
\begin{tikzpicture}

\begin{axis}[%
width=.64\linewidth,
height=.42\linewidth,
scale only axis,
xmin=0,
xmax=40,
xlabel style={font=\color{white!15!black}},
xlabel={Global iteration ($t$)},
ymin=10,
ymax=70,
ylabel style={font=\color{white!15!black}},
ylabel={Test accuracy ($\%$)},
axis background/.style={fill=white},
xmajorgrids,
ymajorgrids,
legend style={at={(0.03,0.97)}, anchor=north west, legend cell align=left, align=left, draw=white!15!black}
]
\addplot [color=mycolor1, line width=1.0pt, mark=o, mark repeat=5, mark options={solid, mycolor1}]
  table[row sep=crcr]{%
1	10.7641666666667\\
2	10.7533333333333\\
3	10.8666666666667\\
4	11.0783333333333\\
5	10.9316666666667\\
6	11.66\\
7	12.2191666666667\\
8	12.5716666666667\\
9	13.1158333333333\\
10	13.8083333333333\\
11	13.7325\\
12	14.2691666666667\\
13	15.5891666666667\\
14	16.0358333333333\\
15	16.9641666666667\\
16	17.7558333333333\\
17	19.4991666666667\\
18	20.095\\
19	22.4391666666667\\
20	23.0341666666667\\
21	26.485\\
22	27.6258333333333\\
23	27.5616666666667\\
24	29.2716666666667\\
25	31.6191666666667\\
26	32.9333333333333\\
27	35.24\\
28	36.41\\
29	36.5508333333333\\
30	38.9425\\
31	40.8066666666667\\
32	42.0725\\
33	41.4883333333333\\
34	44.4591666666667\\
35	44.9575\\
36	46.2666666666667\\
37	46.89\\
38	48.245\\
39	49.9558333333333\\
40	51.105\\
};

\addplot [color=mycolor2, line width=1.0pt, mark=x, mark repeat=3, mark options={solid, mycolor2}]
  table[row sep=crcr]{%
1	10.4783333333333\\
2	10.8925\\
3	11.1666666666667\\
4	12.0925\\
5	12.6991666666667\\
6	13.2383333333333\\
7	14.4558333333333\\
8	15.495\\
9	16.7841666666667\\
10	17.8516666666667\\
11	19.0608333333333\\
12	19.9958333333333\\
13	21.9233333333333\\
14	22.8716666666667\\
15	23.19\\
16	24.1316666666667\\
17	24.9175\\
18	26.2391666666667\\
19	26.9758333333333\\
20	28.2691666666667\\
21	33.9725\\
22	35.6225\\
23	36.0266666666667\\
24	37.3483333333333\\
25	37.4925\\
26	37.7983333333333\\
27	38.2908333333333\\
28	39.7541666666667\\
29	40.5925\\
30	42.4175\\
31	43.8541666666667\\
32	44.25\\
33	43.8533333333333\\
34	45.4991666666667\\
35	46.5408333333333\\
36	47.9008333333333\\
37	49.7875\\
38	51.5558333333333\\
39	52.0966666666667\\
40	52.8733333333333\\
};

\addplot [color=mycolor3, dashdotted, line width=1.0pt]
  table[row sep=crcr]{%
1	10.5558333333333\\
2	10.9291666666667\\
3	11.3508333333333\\
4	12.4283333333333\\
5	13.3358333333333\\
6	13.9766666666667\\
7	15.2233333333333\\
8	17.0258333333333\\
9	18.0158333333333\\
10	18.665\\
11	20.2316666666667\\
12	21.6741666666667\\
13	22.8483333333333\\
14	24.3566666666667\\
15	25.8166666666667\\
16	27.205\\
17	28.425\\
18	28.3933333333333\\
19	30.0541666666667\\
20	32.8608333333333\\
21	36.9808333333333\\
22	38.7916666666667\\
23	39.4725\\
24	40.4741666666667\\
25	41.2441666666667\\
26	41.9283333333333\\
27	42.745\\
28	44.0558333333333\\
29	44.3183333333333\\
30	46.7916666666667\\
31	47.3216666666667\\
32	48.315\\
33	50.475\\
34	51.6033333333333\\
35	51.9916666666667\\
36	54.0591666666667\\
37	55.77\\
38	56.2266666666667\\
39	57.5216666666667\\
40	60.1\\
};

\addplot [color=mycolor4, dashdotted, line width=1.0pt, mark=triangle, mark repeat=3, mark options={solid, rotate=270, mycolor4}]
  table[row sep=crcr]{%
1	10.6833333333333\\
2	11.05\\
3	11.7483333333333\\
4	12.3575\\
5	13.2133333333333\\
6	14.4416666666667\\
7	15.3325\\
8	16.3958333333333\\
9	17.6433333333333\\
10	18.0008333333333\\
11	18.9575\\
12	18.965\\
13	20.2666666666667\\
14	21.8741666666667\\
15	22.2691666666667\\
16	23.5775\\
17	24.165\\
18	25.2841666666667\\
19	27.5591666666667\\
20	28.5833333333333\\
21	32.8125\\
22	36.0908333333333\\
23	37.375\\
24	37.64\\
25	38.9825\\
26	40.56\\
27	43.4525\\
28	43.875\\
29	46.18\\
30	48.1875\\
31	49.6533333333333\\
32	50.73\\
33	52.8666666666667\\
34	54.1925\\
35	55.6575\\
36	56.3708333333333\\
37	57.9741666666667\\
38	58.5133333333333\\
39	59.3733333333333\\
40	59.745\\
};

\addplot [color=mycolor5, line width=1.0pt]
  table[row sep=crcr]{%
1	10.6025\\
2	10.9566666666667\\
3	11.6958333333333\\
4	12.1025\\
5	12.6833333333333\\
6	13.5566666666667\\
7	14.3258333333333\\
8	14.8491666666667\\
9	15.6175\\
10	16.1383333333333\\
11	16.54\\
12	17.1866666666667\\
13	17.8941666666667\\
14	18.3158333333333\\
15	19.6875\\
16	20.9158333333333\\
17	21.2608333333333\\
18	22.8241666666667\\
19	24.1391666666667\\
20	25.8216666666667\\
21	31.1558333333333\\
22	33.3908333333333\\
23	35.08\\
24	36.5916666666667\\
25	38.2216666666667\\
26	40.0425\\
27	42.1833333333333\\
28	43.0433333333333\\
29	44.7725\\
30	45.8941666666667\\
31	47.7158333333333\\
32	49.1716666666667\\
33	50.1941666666667\\
34	51.7816666666667\\
35	53.2483333333333\\
36	54.16\\
37	54.6783333333333\\
38	55.5841666666667\\
39	56.5525\\
40	56.665\\
};

\addplot [color=mycolor6, dotted, line width=1.0pt, mark=square, mark repeat=3, mark options={solid, mycolor6}]
  table[row sep=crcr]{%
1	10.6766666666667\\
2	11.065\\
3	11.3658333333333\\
4	11.9633333333333\\
5	12.2241666666667\\
6	12.8041666666667\\
7	13.1975\\
8	13.865\\
9	14.9675\\
10	14.7641666666667\\
11	16.01\\
12	16.3958333333333\\
13	17.5308333333333\\
14	18.1766666666667\\
15	18.6508333333333\\
16	19.22\\
17	20.1433333333333\\
18	21.6266666666667\\
19	23.2241666666667\\
20	24.2183333333333\\
21	28.4408333333333\\
22	29.5433333333333\\
23	30.9475\\
24	32.2333333333333\\
25	32.0308333333333\\
26	34.6658333333333\\
27	35.0533333333333\\
28	37.9225\\
29	40.0008333333333\\
30	41.8591666666667\\
31	43.1066666666667\\
32	43.5033333333333\\
33	44.4341666666667\\
34	46.2\\
35	47.8158333333333\\
36	49.0583333333333\\
37	50.1866666666667\\
38	50.5325\\
39	51.415\\
40	52.5866666666667\\
};

\end{axis}
\end{tikzpicture}%